\documentclass{article}
\usepackage[utf8]{inputenc}
\usepackage{indentfirst} 

\usepackage{xcolor}
\usepackage{hyperref}
\usepackage{booktabs}
\usepackage{longtable}
\usepackage{amsfonts}
\usepackage{stfloats}
\usepackage{graphicx}
\usepackage[numbers, sort]{natbib}
\usepackage{authblk}
\usepackage{amsmath}
\usepackage{amssymb}
\usepackage{amsthm}
\usepackage[english]{babel}
\usepackage{charter}
\usepackage{fullpage}
\usepackage{pifont}
\usepackage{color,soul}
\usepackage{listings}
\usepackage{lipsum}
\usepackage{empheq}
\usepackage{subfigure}
\usepackage[most]{tcolorbox}
\usepackage{multirow}
\usepackage{tabularx}
\usepackage{makecell}
\usepackage{xspace} 
\usepackage{wrapfig}

\usepackage{amsmath}

\title{Deep Model Fusion: A Survey}

\author{
   Weishi Li$^{1 \dagger}$,\   
   Yong Peng$^{1 \dagger}$,\     
   Miao Zhang$^{1}$,\  
   Liang Ding$^2$,\    
   Han Hu$^3$,\   
   Li Shen$^{2}$\thanks{Corresponding author \\ \indent \  $\dagger$ Equal Contribution}\\ %
     \vspace{0.5em}
$^1$ National University of Defense Technology, China\\  
$^2$JD Explore Academy, China\\
$^3$Beijing Institute of Technology, China\\
   \vspace{0.5em}
    \texttt{liweishi.wh@foxmail.com};
   \texttt{\{yongpeng,zhangmiao15\}@nudt.edu.cn};
   \texttt{\{liangding.liam,mathshenli\}@gmail.com};
   \texttt{hhu@bit.edu.cn}
}

\date{}

\begin{document}

\maketitle
\begin{abstract}
   Deep model fusion/merging is an emerging technique that merges the parameters or predictions of multiple deep learning models into a single one. It combines the abilities of different models to make up for the biases and errors of a single model to achieve better performance. However, deep model fusion on large-scale deep learning models (e.g., LLMs and foundation models)  faces several challenges, including high computational cost, high-dimensional parameter space, interference between different heterogeneous models, etc. Although model fusion has attracted widespread attention due to its potential to solve complex real-world tasks, there is still a lack of complete and detailed survey research on this technique. 
Accordingly, in order to understand the model fusion method better and promote its development, we present a comprehensive survey to summarize the recent progress. 
Specifically, we categorize existing deep model fusion methods as four-fold:
(1) ``Mode connectivity", which connects the solutions in weight space via a path of non-increasing loss, in order to obtain better initialization for model fusion;
(2) ``Alignment" matches units between neural networks to create better conditions for fusion;
(3) ``Weight average", a classical model fusion method, averages the weights of multiple models to obtain more accurate results closer to the optimal solution.
(4) ``Ensemble learning" combines the outputs of diverse models, which is a foundational technique for improving the accuracy and robustness of the final model.
In addition, we analyze the challenges faced by deep model fusion and propose possible research directions for model fusion in the future.
Our review is helpful in deeply understanding the correlation between different model fusion methods and practical application methods, which can enlighten the research in the field of deep model fusion.

\end{abstract}

\section{Introduction}

In recent years, deep neural networks (DNNs) \cite{lecun2015deep} have made remarkable development, which is widely used in computer vision (CV) \cite{o2020deep}, natural language processing (NLP) \cite{chowdhary2020natural} and other fields.
Generally speaking, a single deep learning model often has certain limitations and cannot fully capture all underlying information behind complex networks \cite{sagi2018ensemble}.
Therefore, the classic ensemble learning \cite{schapire1999brief,breiman1996bagging,rokach2010ensemble} combines the outputs of multiple models to improve the final performance of model in deep learning (DL).
But it suffers from the high cost of storing and running multiple models at test time \cite{singh2020model,gao2022revisiting}, especially as the complexity and size of models increase.
Especially, for example, GPT-3 \cite{openai2023chatgpt4} has billions of parameters, and PaLM \cite{chowdhery2022palm} even reaches 540 billion parameters and 780 billion tokens.
In addition, from the perspective of loss landscape of DNNs \cite{li2018visualizing,sagun2017empirical}, gradient-optimized solutions usually converge to points near the boundary of the wide flat region instead of the central point \cite{izmailov2018averaging}.
It means that a trained network is not exactly close to the optimal solution with minimum test error.
The solutions near the relative optimal point need to be fused for a better result.
It inspires researchers not only to limit the the fusion scope to predictions (e.g., logits, etc.), but also to include the fusion of model parameters without accessing the training data or maintaining all individual models \cite{jolicoeur2023population}.
Accordingly, deep model fusion \cite{matena2022merging,jordan2022repair} aims at fusing several DNNs into a single network, which preserves their original capabilities and even outperforms multi-task training \cite{li2022branch,ainsworth2022git}. 
In addition, deep model fusion can reduce the tendency of a single model to overfit particular samples or noise so as to improve the accuracy, diversity and robustness of predictions \cite{utans1996weight,smith2017investigation}.

Deep model fusion has attracted increasing interest due to the data privacy and practical resource-saving issues.
Although the development of deep model fusion has brought many technical breakthroughs, it also produces a series of challenges, such as high computational load, model heterogeneity, and slow speed of alignment via combinatorial optimization \cite{li2019fedmd,singh2020model}, etc.
Some approaches are limited to specific scenarios \cite{yurochkin2019bayesian,wang2020federated}, which inspires researchers to investigate the principles of model fusion in different cases.
Nevertheless, there is a lack of comprehensive reviews to summarize the approaches so as to indicate the internal mechanism of deep model fusion currently.
Some work only focuses on model fusion from a single perspective (e.g., feature fusion, etc.) \cite{dong2020survey,sagi2018ensemble} and a specific scene \cite{sung2023empirical}, or the fusion of information from different ways (multi-modal fusion \cite{afyouni2022multi,jangra2023survey}) rather than the fusion of parameters.
In order to give the developers insight into deep model fusion, we analyze the principles and methodologies of deep model fusion. 
In addition, we review the recent progress and representative applications, such as federated learning (FL) \cite{mcmahan2017communication} and fine-tuning \cite{choshen2022fusing}, etc.
Our survey aims to illustrate the latest trends and potential directions in deep model fusion and provide a guideline for researchers to enhance the performance and reduce costs.
Accoordingly, we group the approaches into four-fold according to the internal mechanisms and purposes as Figure \ref{fig_1}.
For the models trained independently that are not in the vicinity of each other, ``mode connectivity" and ``alignment" bring the solutions closer so as to obtain better original conditions of average. 
For the similar models with certain differences in the weights space, ``weight average (WA)" tends to average the models directly and obtain solutions closer to the optimal point in the region of the parameter space where the value of loss function is low \cite{keskar2016large}.
Furthermore, for the predictions of existing models, ``ensemble learning" integrates different forms of predictions of the models to get better results. 
Specifically, the four categories are as follows:








\begin{figure}
	\centering 
	\includegraphics[width=1\textwidth, angle=0]{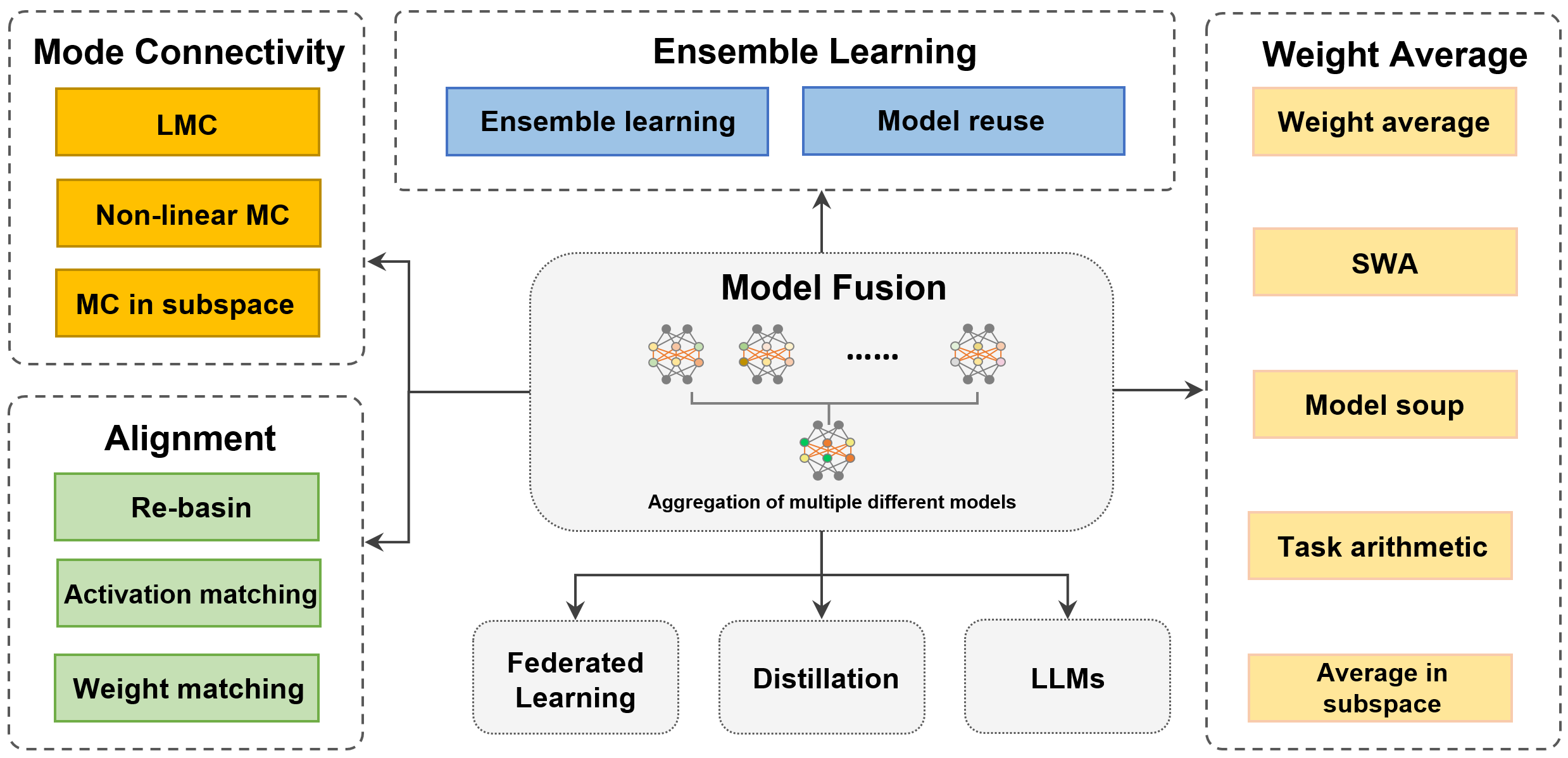}	
	\caption{Schematic diagram of the overall model fusion process, as well as classification and connection of various classification methods.
 } 
	\label{fig_1}%
\end{figure}

\begin{itemize}
\item  \textbf{Mode connectivity.}
\cite{nagarajan2019uniform,2016Topology}, 
The solutions obtained by gradient-based optimization can be connected in weight space by a path (connector) with no obstacles, which is referred to as mode connectivity \cite{entezari2021role,draxler2018essentially}.
We can obtain other models that are more suitable for model fusion along the low-loss path.
According to the mathematical form of path and the space where the connector is located, we divide this section into three parts ``linear mode connectivity (LMC) \cite{garipov2018loss}", ``non-linear mode connectivity" and ``mode connectivity in subspace".
Mode connectivity can solve local optimization problems during training.
The geometric relationships of paths of mode connectivity \cite{nagarajan2019uniform,2016Topology} could also be used to accelerate the convergence, stability and accuracy of optimization procedures like stochastic gradient descent (SGD).
In a word, mode connectivity provides a new perspective for interpreting and understanding the behaviors of model fusion \cite{garipov2018loss}.
But the difficulties of computational complexity and parameter tuning should be solved, especially when training models on large datasets.

 \item  \textbf{Alignment.} 
Alignment \cite{li2015convergent,tatro2020optimizing} matches the units of multiple models and average the models to obtain the final model.
The specific mathematical metrics (e.g., Euclidean distance \cite{tatro2020optimizing}) between different models can be closer after alignment, which can reduce the differences between models, thus enhancing the effect of deep model fusion.
Alignment can be divided into ``activation matching" and ``weight matching" depending on whether data distribution needs to be considered.
Moreover, Re-basin \cite{ainsworth2022git}  is introduced based on alignment, which explores the mechanism that solutions can be transported into a single basin (i.e., area of the flat parameter space where with relatively low loss \cite{2016Topology,2021Geometry}) by permutation invariance \cite{entezari2021role}.
However, it is often faced with the obstacles of large computation, slow speed pf combinatorial optimization and architecture difference, which makes it is not easy to be extended to other scenarios with different objectives.
For example, the memory burden that comes with graph matching \cite{liu2022deep,wang2020clustering} limits the application of deep model fusion.


\item  \textbf{Weight average.} 
WA \cite{wang2020federated} is the most direct and efficient way to fuse several parent networks into a single network \cite{matena2022merging,singh2020model}.
Compared to mode connectivity and alignment, WA does not require additional computational complexity or training to find a superior starting point, which performs well on models contain a degree of similarities.
According to the space of aggregation, WA can be classified into two parts ``weight average" and ``average in subspace" .
In addition, the typical approaches ``model soup", ``model arithmetic`` and ``stochastic weight averaging (SWA)" also provide significant improvements over the existing methods.
Furthermore, some bias may be introduced in the case of large differences in model structure or number of parameters when the parameters are normalized and merged.
Nonetheless, WA is still the mainstream method of deep model fusion because of its simplicity and efficiency.

\item  \textbf{Ensemble Learning.}
The outputs of several different models are combined to improve the prediction performance and robustness, which is regarded as ``ensemble learning" \cite{sagi2018ensemble}.
In this review, we focus on the ensemble learning in DL.
Based on ensemble learning, ``model reuse" provides specifications for each model so that useful models can be identified and merged from the pool of models when given new learning tasks \cite{zhou2016learnware, pathak2010multiparty}.
Ensemble learning has various frameworks with convenient interfaces, which is often used in practical areas such as object detection \cite{casado2020ensemble}, etc.
Although ensemble learning requires maintaining the multiple trained models and running each of them at test time \cite{singh2020model}, it is still one of the powerful techniques that have been widely adopted in DL.


\item \textbf{Applications of Model Fusion.}
As a technology to improve the accuracy and robustness of deep models, model fusion promote the improvement to many application fields.
``federated learning \cite{mcmahan2017communication}", an application of aggregating clients' models on a central server, makes it possible for various parties to contribute data to the computation of functions (e.g., various statistics, classifiers \cite{pathak2010multiparty}) without the risks of privacy disclosure.
``fine-tuning" makes small adjustments to pre-trained models, which combined with model fusion to reduce training costs and adapt to the needs of a specific task or domain.
Model fusion is also involved in ``distillation".
That is, combine soft target knowledge from multiple complex models (teachers) to train a small model for specific requirements.
``model fusion on foundation/LLMs" includes the work on large foundation models or large language models (LLMs), such as vision transformers (ViT) \cite{han2021transformer} and GPT \cite{brown2020language}, etc.
The applications of model fusion help developers adapt to the needs of various tasks and domains and promote the development of DL.
\end{itemize}

In brief, our survey reviews deep model fusion techniques.
In the first three sections ``mode connectivity", ``alignment" and ``weight average", we mainly conduct a comprehensive study from the perspective of the fusion of model parameters.
In the ``ensemble learning", we mainly investigate the issue from the perspective of model outputs aggregation.
The main contributions of this work are summarized as:

\begin{itemize}

    \item We propose a new deep model fusion classification method from the perspectives of ``mode connectivity", ``alignment", ``weight average" and "ensemble learning", which covers the theoretical synthesis approaches of model fusion, and provides guidance for the realization of high generalization and accuracy training of DNNs.

    \item We compare the advantages and disadvantages of fusion approaches, and explain the mechanism and relationship between them, which provides inspiration for designing advanced model fusion methods in the future.

    \item We summarize extensive application of deep model fusion. We also discuss current research trends so as to attract more attention and reflection in the future.

\end{itemize}



Moreover, the remainder of the paper is organized as follows:
In Section 2 to Section 5, we introduce the approaches of deep model fusion according to the four perspectives ``mode connectivity``, ``alignment``, ``weight average`` and ``ensemble learning``. 
Section 6 introduces the applications of deep model fusion ``federated learning``, ``fine-tuning``, ``distillation`` and ``model fusion on foundation/LLMs``.
Finally, in Section 7, we summarize the deep model fusion and discuss the challenges and potential directions in the future.

In addition, we illustrate the notations and their corresponding definitions in the full text.
$\boldsymbol{W}_{i}$ is the $i_{th}$ neural network with weights $W_{i}\in \mathbb{R}^{d}(i=1,2,...k)$ and bias term $\boldsymbol{b}$.
$\lambda$ denotes weighted parameters.
$\sigma$ denotes a non-linear neuron activation function.
$\mathcal{L}$ is loss function that quantify the discrepancy between the predicted and actual values.
\section{Mode Connectivity}

\begin{table}
\caption{The summary of standard training pipelines of LMC and non-linear mode connectivity.}
\label{table1}
\vspace{0.5em}
\begin{tabular}{l l l l} 
\toprule
\makecell[l]{Mode\\connectivity} & The form of path & Ref. & Eq.  \\

   \midrule
  \multirow{1}*{\makecell[l]{Linear\\path}}
&segment  & 	\cite{frankle2020revisiting,fort2020deep} &$\phi (t)=(1-t)w _{1} +t w _{2}$ \\ 

 
& polygonal chain & 	\cite{gomes2012computer,garipov2018loss} &
 $ \phi(t)=\left\{\begin{array}{ll}
2\left(t w+(0.5-t) w_{1}\right), & 0 \leq t \leq 0.5 \\
2\left((t-0.5) w_{2}+(1-t) w\right), & 0.5 \leq t \leq 1
\end{array}\right.$
    \\

\midrule
  \multirow{1}*{\makecell[l]{Non-linear\\path}}
  &quadratic Bezier curve  &\cite{lubana2023mechanistic,farouki2012bernstein} &  
 $\phi(t)=(1-t)^{2} w _{1}+2 t(1-t) w+t^{2} w_{2}, \quad 0 \leq t \leq 1$
 \\
& \makecell[l]{Fourier series \\approximate curves}  &\cite{wen2023optimizing} &  
$\hat{\phi}(t)=\frac{\beta_{0}}{2}+\sum_{i=1}^{n} \beta_{i} \cos \left(w_{i} t+ \zeta _{i}\right)$
 \\


 \bottomrule
\end{tabular}
\end{table}

In this section, we introduce the definition, principles and related methods of mode connectivity.
When training neural networks, the solutions trained by gradient-based optimization algorithms (e.g., SGD, etc.) can be merged without superior results \cite{2016Topology,draxler2018essentially}.
It is discovered that solutions can be connected via continuous paths (connectors) in the network weight space without increasing loss, which is referred to as mode connectivity \cite{entezari2021role,garipov2018loss}.
The models on the low-loss path can be fused to leverage the advantages of multiple models by mode connectivity, which is of great significance to produce a better aggregation model.



First, we explain the principles of mode connectivity.
In a representative process of DL, the minima is usually described as a point at the bottom of a convex valley, the network parameters are determined by the location of the minima \cite{keskar2016large,hochreiter1997flat,kawaguchi2016deep}.
The traditional view is that the number of local minima and saddle points is large \cite{goodfellow2014qualitatively,wang2018identifying}, and different local minima will converge to different isolated regions in the parameter space \cite{auer1995exponentially,choromanska2015loss,dauphin2014identifying}.
Recent work \cite{sagun2017empirical,kuditipudi2019explaining} demonstrates that the minima obtained by gradient-based optimizer are not walled off in isolated valleys \cite{2016Topology}.
Gotmare et al. \cite{gotmare2018using} explore the potential relationship between the minima found by different training process.
Other work \cite{draxler2018essentially,nguyen2018loss,cooper2021global,pittorino2022deep} manifest that neural network solutions form a connected manifold (i.e., solutions in the loss landscape are connected by pipelines in weight space).
Compared with mode connectivity, a direct linear path connecting two such independently trained networks usually always leaves a low-loss manifold, which creates a high loss barrier at the points on the linear path.
For example, the error at the midpoint of the line segment directly connecting two points is closed to 90$\%$ (VGG-16 on CIFAR-10 \cite{garipov2018loss}).
The above work proves the existence and effect of mode connectivity.

\begin{figure}
	\centering 
	\includegraphics[width=1\textwidth, angle=0]{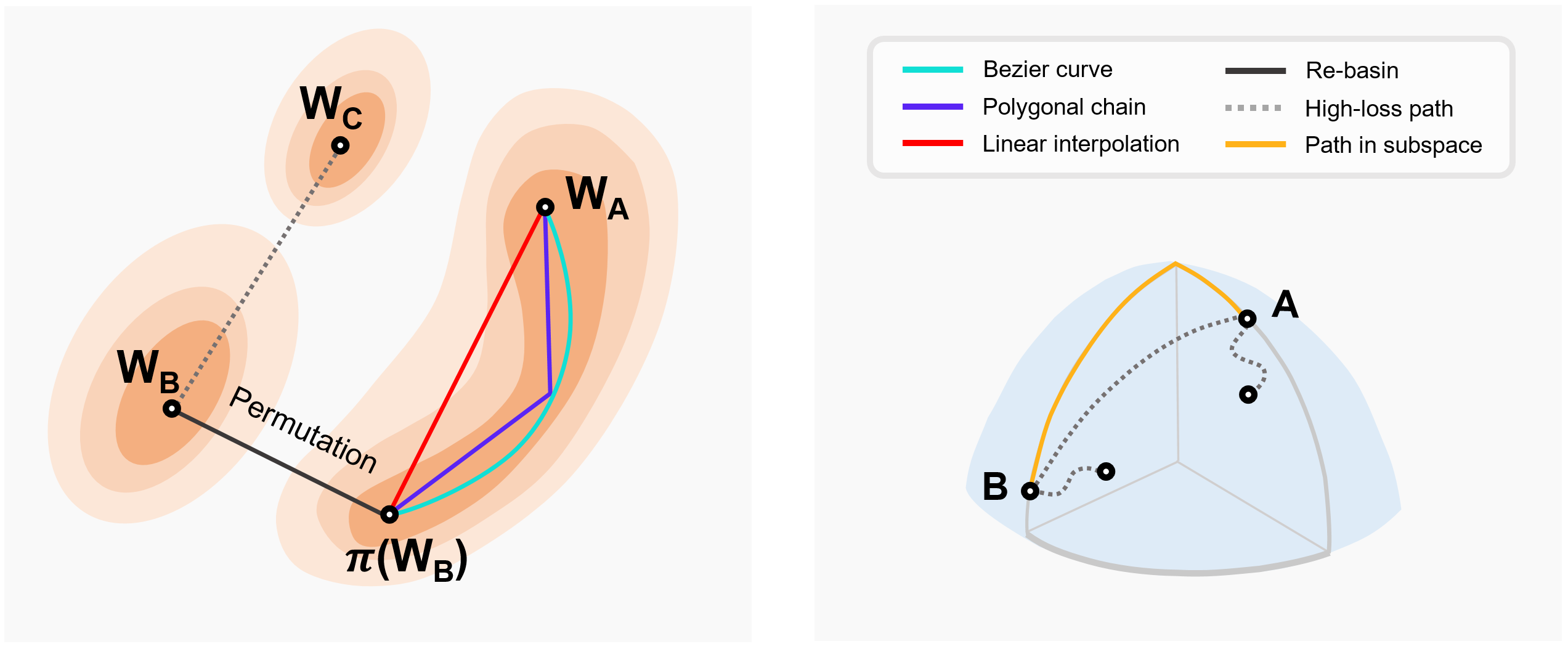}
 \caption{Mode connectivity schematic diagram in 
 two-dimensional loss landscape and other dimensional subspace. 
\textbf{Left: }Linear interpolation of the minima in the two basins results in high-loss barriers\cite{draxler2018essentially}.
The lower two optimums follow a path of near constant low loss (e.g., Bezier curve, Polygonal chain, etc.)\cite{garipov2018loss}.
$\pi(W_{2})$ is the equivalent model of $W_2$ by permutation symmetry, which is located in the same basin as $W_1$. 
Re-Basin merges models by delivering solutions to individual basins \cite{ainsworth2022git}.
\textbf{Right: }
Low loss paths connect multiple minima in subspace(e.g., a low-loss manifold composed of $d$-dim wedges \cite{fort2019large}), etc.).} 
\label{fig_2}%
\end{figure}

Second, some work \cite{frankle2018lottery,garipov2018loss,entezari2021role} quantifies the pipelines of the mode connectivity.
Let $\mathcal{L}\left(t w_{1}+(1-t) w_{2}\right)$ for $t\in (0,1)$ be the loss (train or test error) of a neural network created by linearly interpolating between $\boldsymbol{W}_{1}$ and $\boldsymbol{W}_{2}$.
The random data augmentations in each epoch can be seen as noise when using SGD with the initialization and hyperparameters fixed.
To determine whether the result of a trained network is stable to SGD noise, the loss barrier (error barrier) $B\left(w_{1}, w_{2}\right)$ \cite{frankle2020linear} is defined as the maximum difference between the linear interpolation of the loss at each point and the loss of the linear connection of two points \cite{entezari2021role}, as shown in Eq.~\eqref{Eq:lmc-1}:
\begin{equation}\label{Eq:lmc-1}
    B\left(w_{1}, w_{2}\right)=\sup _{t}\left[\mathcal{L}\left(t w_{1}+(1-t) w_{2}\right)\right]-\left[t \mathcal{L}\left(w_{1}\right)+(1-t) \mathcal{L}\left(w_{2}\right)\right].
\end{equation}
The loss barrier illustrates whether the error is constant or increased when we optimize the landscape \cite{fort2019large,2016Topology} along the path between $\boldsymbol{W}_{1}$ and $\boldsymbol{W}_{2}$.
If there is a tunnel between two networks with a barrier approximately equal to 0, which is equivalent to mode connectivity \cite{frankle2020linear,draxler2018essentially,frankle2018lottery}.
That is to say, the local minima obtained by SGD can be connected by a path $\phi$ with the lowest maximum loss as shown in Eq.~\eqref{Eq:lmc-2}:
\begin{equation}\label{Eq:lmc-2}
    \phi\left(w_{1}, w_{2}\right)=\underset{\phi \text { from } \boldsymbol{W}_{1} \text { to } \boldsymbol{W}_{2}}{\operatorname{argmin}}\left\{\max _{w \in \phi } \mathcal{L}(w)\right\},
\end{equation}
which means that the loss is low along the pathway and the network is stable to SGD noise \cite{draxler2018essentially}, as shown in Figure \ref{fig_2}. There are two steps to conduct mode connectivity: 
first determine the form of the tunnels (e.g., polygonal chain, Bezier curve \cite{garipov2018loss}, etc.) as Table \ref{table1}; then find the optimal low-loss pathway to connect different solutions, as shown in Table \ref{table2}.
According to the form of path and the space in which it is located, this section introduces ``Linear mode connectivity", ``Non-linear mode connectivity" and ``Mode connectivity in subspace".

\subsection{Linear Mode Connectivity}

%

In order to connect two points on an optimized low-loss path, we first need to determine the form of the tunnel.
If the optimal path $\phi^{*}$ is linear, then it is called LMC.
Common linear paths are linear segment, polygonal chain as Eq.\eqref{Eq:lmc-3}:
\begin{equation}
\label{Eq:lmc-3}
    \phi_{w}(t)=\left\{\begin{array}{ll}
2\left(t w+(0.5-t) w_{1}\right), & 0 \leq t \leq 0.5 \\
2\left((t-0.5) w_{2}+(1-t) w\right), & 0.5 \leq t \leq 1
\end{array}\right.
,
\end{equation}
The parametric path train using the same hyperparameters from different random initialization.
$\phi_{w}(0)=w_{1} ,\phi_{w}(1)=w_{2} $.
After deciding on the mathematical form of tunnel, the specific parameters need to be determined.
Garipov et al. \cite{garipov2018loss} suggest to minimize the expectation of loss $\ell(w)$ over a uniform distribution as Eq.\eqref{Eq:lmc-4}:
\begin{equation}
\label{Eq:lmc-4}
    \min _{w}
    \ell(w)=
    \min _{w}
    \mathbb{E}_{t \sim U(0,1)} \left[\mathcal{L}\left(\phi_{w}(t)\right)\right],
\end{equation}
In addition, the tunnel found by this way is not unique.
Nevertheless, vanilla mode connectivity are not robust enough to resolve various types of adversarial attacks.
Robust mode connectivity (RMC) \cite{wang2023exploring} uses adversarial training (AT) \cite{madry2017towards} to find tunnels between neural networks that exhibit robustness to different types of adversarial attacks as Eq.\eqref{Eq:lmc-5}:
\begin{equation}
\label{Eq:lmc-5}
\min _{w}
\ell(w)=
   \min _{w} \mathbb{E}_{t \sim U(0,1)}  \sum \max _{\operatorname{Dist}_{i}\left(\mathbf{x}^{\prime}, \mathbf{x}\right) \leq \delta_{\mathbf{i}}} \mathcal{L}\left(\phi_{w}(t) ;\left(x^{\prime}, y\right)\right),
\end{equation}
where $\delta_{i}$ are minimal values, $Dist_{i}$ denotes distance measurement function. The RMC path in the parameter space improves robustness to different types of attack.
%
Some work complements the LMC from a global connectivity perspective.
Nguyen et al. \cite{nguyen2019connected} prove that when the number of neurons in a hidden layer is larger than a certain amount of training samples, the loss function has no so-called bad local valleys, and all the global minima are connected in a large global valley.
%
Shevchenko et al. \cite{shevchenko2020landscape} demonstrate that as the number of neurons increases (over-parameterization), the landscape of the multi-layer network is connected, which is more conducive to LMC.
Although previous studies speculate that interconnected local minima in over-parameterized networks mean the mode connectivity of the loss function, which does not always hold true (e.g., over-parameterized two-layer networks \cite{kuditipudi2019explaining}).
Kuditipudi et al. \cite{kuditipudi2019explaining} explain mode connectivity by noise stability \cite{frankle2020linear,arora2018stronger}, which is somewhat equivalent to dropout stability.
In other words, all noise stabilization solutions can be connected in a sufficiently over-parameterized network.

As for the practical application of LMC, Zhao et al. \cite{zhao2020bridging} suggest to use LMC to repair backdoored or error-injected models.
Neyshabur et al. \cite{neyshabur2020being} show the application of LMC to pre-trained visual models.
Qin et al. \cite{qin2022exploring} explore the relationship between different downstream configurations and mode connectivity of language model models.

\subsection{Non-linear Mode Connectivity}
In this subsection, we focus on the non-linear pathway connected solutions in weight space, which is known as non-linear mode connectivity \cite{juneja2022linear,qin2022exploring}.
Bezier curve is one of the representative form of non-linear path as Eq.\eqref{Eq:lmc-6}:
\begin{equation}
\label{Eq:lmc-6}
     \phi_{w}(t)=(1-t)^{2} w _{1}+2 t(1-t) w+t^{2} w_{2}, \quad 0 \leq t \leq 1 .
\end{equation}
Compared with non-linear connectivity, the convex combinations (LMC) of minima within the loss basin remain in the same basin.
in contrast,  the nonlinear connectivity between minima are not located in the same basin, which means that the LMC is not available in some cases.

Recent work \cite{kuditipudi2019explaining,draxler2018essentially,lubana2023mechanistic} show that different independently trained networks can be connected by nonlinear pathways that remain in the low-loss manifold in the weight space.
Qin et al. \cite{qin2022exploring} speculate that there may be multiple loss basins connected by low loss nonlinear paths.
Yun et al. \cite{yun2023traversing} indicate that output can be obtained by connecting the Bezier curves of the two network parameters in the absence of an actual forward passing network in the Bridge network.
Gotmare et al. \cite{gotmare2018using} manifest that non-linear mode connectivity is widely applied to networks trained with different optimizers, data enhancement strategies and learning rate schedules.
Futhermore, Lubana et al. \cite{lubana2023mechanistic} explain the principle of mode connectivity by mechanistic similarity, which is defined as the fact that two models are mechanistically similar if they make predictions using the same properties (e.g., shape or background) of the input.
The mechanistic similarity of the induced models is related to LMC of two minimizers (minima).
There is no LMC between mechanistically dissimilar minimizers, but mode connections can be made via relatively non-linear paths.
The representative approach for finding nonlinear path \cite{garipov2018loss} is similar to LMC, as Eq.\eqref{Eq:lmc-7}:
\begin{equation}
\label{Eq:lmc-7}
\min _{w}
\ell(w)
=\min _{w}
\mathbb{E}_{\alpha \sim q_{w}(t)}\left[\mathcal{L}\left(\phi_{w}(t)\right)\right],
\end{equation}
where $q_{w}(t)$ is the distribution for sampling the models along the path.
Moreover, Draxler et al. \cite{draxler2018essentially} use AutoNEB \cite{kolsbjerg2016automated} and minimum spanning tree (MST) to generate the approximation of $\phi^{*}$ connecting the minima of networks on CIFAR-10 and CIFAR-100.
AutoNEB connects two solutions, which updates the pivot after each iteration until AutoNEB-tunnel approaches the optimal low-loss path $\phi^{*}$.
Nevertheless, the approximation of $\phi^{*}$ may fall into a local minima tunnel with unreasonable high saddle point losses.

To sum up, both linear and nonlinear paths can result in low test errors.
While linearly connected pathways are simple, it could have certain limitations.
As for non-linear mode connectivity, it is difficult to calculate the gradient on some non-linear path such as Bezier curve.

\subsection{Mode Connectivity in Subspace}

\begin{table}\centering

\caption{The methods of finding tunnels between different local minima.
}
\label{table2}
\vspace{0.5em}
\begin{tabular}{l l l l} 
\toprule
 Connectors &  Methods & Ref. & Introduction   \\ 
 \midrule
\multirow{1}*{2-dim path}
& line segment & \cite{goodfellow2014qualitatively} & produce big error \\

& GDSS & \cite{2016Topology}  & \makecell[l]{approximate the geodesic \\paths via GDSS}
\\


& AutoNEB & \cite{kolsbjerg2016automated,draxler2018essentially} &\makecell[l]{minimize MST to obtain \\approximation of $\phi^{*}$ } \\
  
		~ &  \makecell[l]{minimize the\\expectation} & \cite{garipov2018loss} & \makecell[l]{representative approach that \\connects solutions in a simple way} \\
  & RMC & \cite{wang2023exploring}  & \makecell[l]{ enhance the robustness of DNNs\\against different perturbations} \\

\midrule
\multirow{1}*{N-dim space} 
& MPO  &  \cite{skorokhodov2019loss,czarnecki2019deep}  &  obtain substantial memory savings \\

 &   N-dimensional connectors & \cite{fort2019large}   & connect low-dimensional wedges\\


        ~ &   train parametric subspace &\cite{wortsman2021learning} & \makecell[l]{
 learn the parameters of\\lines, curves and simplexes }\\

   & SPRO, ESPRO &  \cite{benton2021loss}  &    \makecell[l]{find simplexes and simplicial\\ complexes to seek connectors} \\

 & geodesic optimization & \cite{tan2023geodesic}  &\makecell[l]{speculate the geodesics in\\the curved distribution space} \\


\bottomrule
\end{tabular}
\end{table}

Previous work of mode connectivity \cite{garipov2018loss,fort2019large,fort2020deep} focuses on low-loss tunnels in weight space without explicitly addressing other dimensional structure.
This subsection explores the mode connectivity and model training in subspace of another dimension rather than in a native parameter space.
Subspace in machine learning typically describe linear structures generated by vectors in the initial vector space.
There are also concepts of non-linear subspace, such as nonlinear dimensionality reduction and manifold learning \cite{izenman2012introduction}.
Standard neural network training is performed on a full parameter space $\mathbb{R}^{D}$.
Limiting the optimization to a random low-dimensional affine subspace (e.g., low-dimensional hyperplanes and hyperspheres, etc.) also leads to the similar results as full-space optimization in some cases \cite{fort2019goldilocks,li2018measuring} , which lay the foundation for mode connectivity in subspace.
Definitely, mode connectivity in oriented subspace constrain the representation ability of the model and the value range of the weights, so as to overcome the over-fitting problem of model fusion.

Recent work attempts to implement mode connectivity in different subspace.
Fort et al. \cite{fort2019large} extend the concept of low-loss connectors (tunnels) between solutions to $m$-dimensional connectors ($m$ is smaller than the dimension of full parameter space).
Randomly initialized points that are not on the same wedge (i.e., a union of m-dimensional manifolds) can always pass through the intersection of their wedges, thus building a low-loss path between the different minima, as shown in Figure \ref{fig_2}.Based on the speculation, the $m$-dimensional hyperplanes are constructed on the piece-wise linear interpolation between the points, in which the low-loss connectors can be found.
Benton et al. \cite{benton2021loss} propose simplicial point-wise random optimization (SPRO) to connect models through a multi-dimensional manifold.$\mathcal{K}\left(S_{\left(w_{0}, \varepsilon _{0}\right)}, S_{\left(w_{1}, \varepsilon _{0}\right)}\right)$ denote simplicial complex composed of disjoint $0$-simplexes. 
SPRO adds the join points $\varepsilon _{i} $ to connect 0-simplexes in the complex iteratively so as to keep the loss low within the simplicial complex.
It obtains a complex $\mathcal{K}$ by sharing multiple $\varepsilon _{i} $.
When a join point $\varepsilon_{k} $ connects the two modes, the pathway of complex $\mathcal{K}\left(S_{\left(w_{0}, \varepsilon _{0}\right)},..., S_{\left(w_{n}, \varepsilon _{0}\right)}\right)$ can be found by previous method \cite{garipov2018loss}.
When some joint points connects multiple modes, the solution to $\mathcal{K}\left(S_{\left(w_{0}, \varepsilon _{0},\varepsilon _{1},\varepsilon _{2}\right)}, ...,S_{\left(w_{n}, \varepsilon _{0},\varepsilon _{1},\varepsilon _{2}\right)}\right)$ is similar to the above work \cite{fort2019large}. For narrow architectures of networks, geodesic optimization \cite{tan2023geodesic} finds a low-loss pathway connecting the solutions where general tunnels of mode connectivity can not pass through a region of high loss.
The mode connectivity pathways in weight space is associated to the geodesics $\gamma$ (i.e., shortest paths in the space of parameterized distributions, which is regarded as a Riemannian manifold with fisher information matrix $f_{ij}$).
The geodesics $\gamma$ is obtained by minimizing the loss $\mathcal{L}(\gamma)$, which is equivalent to the integral of the square root Jensen-Shannon Divergence (JSD) \cite{crooks2007measuring} as Eq.\eqref{lmc_8}:
\begin{equation}
\label{lmc_8}
    \mathcal{L}(\gamma)=\int_{t} \sqrt{\frac{d \gamma^{i}}{d t} f_{i j} \frac{d \gamma^{j}}{d t}} d t=\sqrt{8} \int_{\gamma} \sqrt{d \mathrm{JSD}}
\end{equation}



Further, the mode connectivity in subspace is affected by the properties of the subspace, such as the relationship between dimension of the plane and the inherent dimension specific to the problem, the radius in the weight space, the dimensions of the hyperplane \cite{li2018measuring}, etc.
Moreover, Fort et al. \cite{fort2019deep} explore training tracks and subspace sampling (e.g., dropout, diagonal Gaussian, low-rank Gaussian and random subspace), which further complement relevant work of mode connectivity in subspace.
In addition, recent work \cite{dimitriadis2023pareto} inspires us to explore the mode connectivity in Pareto manifold to be applied to multi-task learning.
In sum, the trained solutions can be found in both the full parameter space and the random low-dimensional hyperplane, as long as the points are distributed densely enough in most cases.

\subsection{Discussion}
In summary, mode connectivity provides a more novel and flexible perspective for deep model fusion.
The training of neural networks tends to fall into local optima, which leads to  degradation of performance.
On the basis of model connectivity, we can find other models with better performance and use that as a starting point for further optimization and fusion.
We can use the already trained model to move in the parameter space to reach the new target model, which can save time and computing overhead, and is suitable for situations where data is limited.
Nevertheless, additional complexity and flexibility may be introduced to increasing the risk of overfitting when connecting different models.
Therefore, the relevant hyperparameters and degree of variation should be carefully controlled.
Also, mode connectivity requires fine-tuning or parameter changes, which can increase training time and resource consumption.
In summary, model connectivity has many advantages in model fusion, including helping to overcome local optimal problems, providing new perspectives to explain network behavior, etc.
In the future, mode connectivity is expected to help understand the inner mechanism of neural networks and provides guidance for more efficient deep model fusion designs in the future.


\section{Alignment}

\begin{table}[!h] \small 
\centering
\caption{Comparison of representative alignment methods.} 
 \label{table_3}
\vspace{0.5em}
\begin{tabular}{l l l l}
\toprule
   &Alignment & Methods & Ref. \\
\hline

    \multirow{1}{*}{\makecell[l]{Activation \\matching}}
   
      & \multirow{1}{*}{\makecell[l]{metrics}} & coefficient of correlation & \cite{li2015convergent,tatro2020optimizing} \\

		&  & mutual information &\cite{li2015convergent} 	\\

		&  & $\ell2$ distance & \cite{tatro2020optimizing,singh2020model,ainsworth2022git} 	\\

      & \multirow{1}{*}{\makecell[l]{pre $\&$ post\\  activation} } & pre-activation & \cite{tatro2020optimizing,singh2020model}\\

		&  & post-activation & \cite{li2015convergent,tatro2020optimizing} 	\\

\midrule
    \multirow{1}{*}{\makecell[l]{Weight\\matching}}
   
      & \multirow{1}{*}{\makecell[l]{metrics}} & Wassertain distance & \cite{singh2020model,akash2022wasserstein,wei2023ntk}\\
	
		&  & Euclidean distance & \cite{ainsworth2022git,pena2022re} 	\\

 & \multirow{1}{*}{graph matching} & bipartite matching &  \cite{li2015convergent,lam2021model}\\


		&  & graph matching & \cite{liu2022deep} 	\\

 & \multirow{1}{*}{other alignment}& Bayesian & \cite{yurochkin2019bayesian,wang2020federated} \\


 & & Sinkhorn Re-basin&  \cite{pena2022re} \\

 & & SA  &  \cite{entezari2021role} \\


\bottomrule
 \end{tabular}
 \end{table}

Due to the randomness of channels and components from diverse networks, the active components of the networks interfere with each other \cite{singh2020model}.
So unaligned weighted averages could ignore correspondence between units from diverse models and damage useful information.
For example, there is a relationship between two neurons in different models that could be completely different but functionally similar.
Alignment matches the units of different models so as to obtain better initial conditions for deep model fusion.
It aims to make multiple models have smaller differences and , thus enhancing the deep model fusion effects.
Also, alignment can be regarded as a combinatorial optimization issue in essence.
In this section, we introduce a representative mechanism ``Re-basin", which delivers solutions to individual basins so as to merge models with better original conditions.
Following this, we divide the alignment into two types ``Activation matching" and ``Weight matching" depending on whether the aligned target is data-driven as Table \ref{table_3}.




\subsection{Re-basin}
Before introducing the specifics, we illustrate the permutation symmetry and Re-basin, which is the basic premise of alignment.
Generally speaking, the number of saddle points and local optima can increase exponentially with the number of parameters even for shallow neural networks \cite{auer1995exponentially,garipov2018loss}.
It is discovered that there are invariances in training that leads to the same representation of some points among these local optima \cite{hecht1990algebraic,chen1993geometry,li2015convergent}.
Specifically, the function of the network will not change if the units of hidden layer are exchanged by permutation, which is referred to as permutation symmetry \cite{dinh2017sharp,entezari2021role}.
Formally, a $\ell$-layer function of DNN $f^{(\ell)}(x,w)=\sigma(W^{(\ell -1)}f^{(\ell-1)}+b^{(\ell -1)})$ can be described as Eq.\eqref{align_9} \cite{ainsworth2022git}:

\begin{equation}
\label{align_9}
    f^{(\ell)}(x,w)=
    \boldsymbol{P}^{T}
 \sigma\left(\boldsymbol{P} W^{(\ell-1)} 
 f^{(\ell-1)}+\boldsymbol{P} b^{(\ell-1)}\right),
\end{equation}
where $\boldsymbol{P} $ denotes the permutation matrix.
We can obtain the functional equivalent model $  f(x ; w)=
             f\left(x ; \pi (w)\right)$ by rearranging the input.
On the basis of permutation symmetry, solutions from diverse area in weight space can generate equivalent solutions.
A equivalent solution is located in a same region as the original solution with low-loss barrier (basin), as shown in Figure \ref{fig_2}, which is referred to as ``Re-basin" \cite{ainsworth2022git} as Eq.\eqref{align_10}:
\begin{equation}
\label{align_10}
\text{Re-basin: }
    f^{(\ell)}(x,w)=
 \sigma\left(P^{(\ell)} W^{(\ell)} (P^{(\ell -1)})^{T}
 f^{(\ell)}+P^{(\ell)} b^{(\ell)}\right)
\end{equation}
Once the optimal permutation matrix $\boldsymbol{P}^{*}$ is obtained, it is theoretically possible to implement model fusion: $W=\lambda_{1} W_{1}^{(\ell)}+\lambda_{2} P^{(\ell)} W_{2}^{(\ell)} (P^{(\ell-1)})^{T}$.
Compared with mode connectivity, Re-basin tends to transport the points into a basin by permutation instead of low-loss tunnels.
At present, alignment is a representative approach of Re-basin\cite{ainsworth2022git,pena2022re}.
However, how to efficiently search for all possibilities of permutation symmetry so that all solutions point to the same basin is a current challenge.

\begin{figure}
	\centering 
	\includegraphics[width=1\textwidth, angle=0]{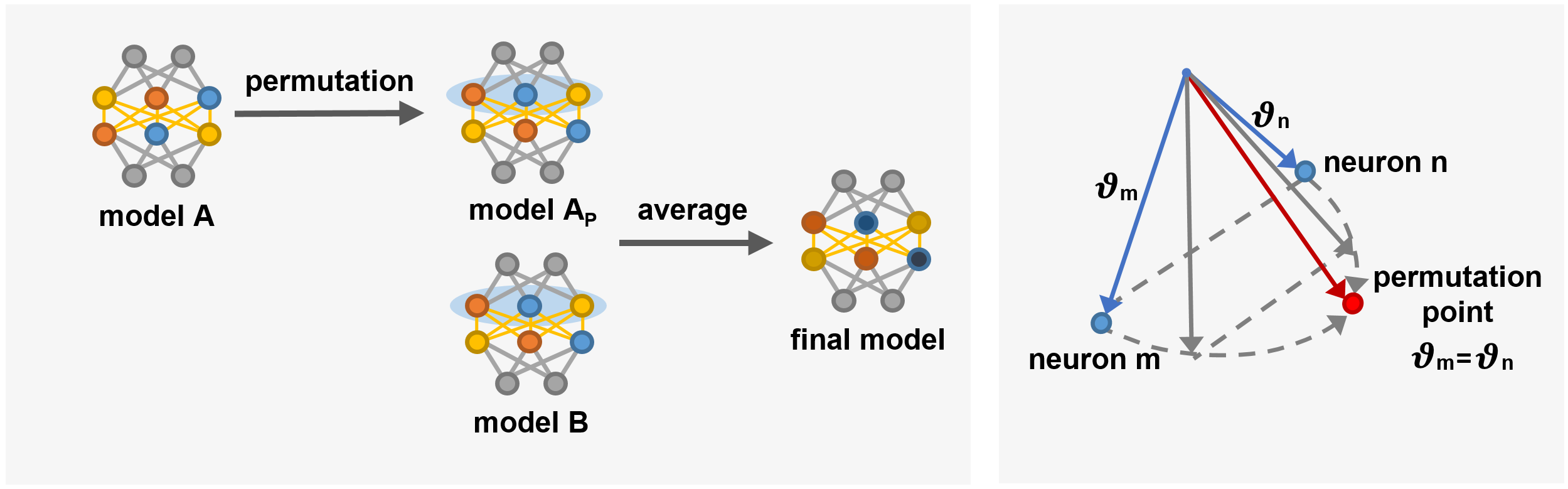}	
	\caption{
\textbf{Left: }general alignment process.
Model $A$ is transformed into model $A_{p}$ by reference to model $B$.
Then the linear combination of $A_{p}$ and $B$ produces C.
\textbf{Right:} adjust the parameter vectors of the two neurons $\vartheta_{m}$,$\vartheta _{n}$ in different hidden layers are close to the replacement point.
At the replacement point, \cite{brea2019weight}, $\vartheta_{m}^{\prime}=\vartheta _{n}^{\prime}$, and the two neurons compute the same function, which means that two neurons can be exchanged.
}
	\label{fig3}%
\end{figure}

Permutation symmetries imposed by these invariances help us understand the structure of loss landscapes better \cite{chen1993geometry,garipov2018loss}.
The invariances also can be seen as the source of saddle points in loss landscapes \cite{brea2019weight}.
Godfrey et al. \cite{godfrey2022symmetries} investigate the algebraic structure of symmetries in neural networks and how this structure manifests itself in loss landscape geometry.
Brea et al. \cite{brea2019weight} introduce permutation point in high-dimensional plateaus, at which the neurons can be exchanged without increasing losses or parameter jumps as Figure  \ref{fig3}.
Conduct gradient descent on the loss and adjust the parameter vectors $\vartheta_m$ and $\vartheta_n$ of neuron $m$ and $n$, until the vectors reach the permutation point.
At this time, the parameter configuration is called permutation point, and the parameter vectors and function of the two neurons are the same .
Furthermore, Tatro et al. \cite{tatro2020optimizing} explore the permutation symmetry of the nonlinear mode connectivity.
Benzing et al. \cite{benzing2022random} speculate that two random initialization of a network after permutation can lead to a good performance.
Furthermore, the alignment method does not always generate good low-loss connections between solutions due to variance collapse of activations.
REnormalizing Permuted Activations for Interpolation Repair (REPAIR) \cite{jordan2022repair} mitigates the variance collapse by rescaling the preactivation of networks, which eliminate the 90$\%$ barrier for ResNet-18 on CIFAR-10 after alignment.


\subsection{Activation Matching}

In this subsection, based on permutation symmetry, we focus on the matching of activation values.
The initial models for fusion can be improved by reducing the differences in activation.
Minimizing the cost functions between activations is a representative way to calculate $\boldsymbol{P}^{*}$, which can be transformed into assignment problems, such as linear assignment problem (LAP) or quadratic allocation problem (QAP), etc.
They can be solved by Hungarian algorithm or Sinkhorn algorithm.
The common cost functions $\mathcal{C}$ used in alignment are cross-correlation \cite{tatro2020optimizing} as Eq.\eqref{align_11}, mutual information (information entropy) \cite{li2015convergent} as Eq.\eqref{align_12}, $\ell2$ distance \cite{ainsworth2022git} as Eq.\eqref{align_13}, KL divergence, Wasserstein distance, etc.
\begin{equation}
\label{align_11}
    \mathcal{C}(A_{ m},A_{ n}) _{cor}=\mathbb{E}\left[\left(A_{m}-\mathbb{E}\left[A_{ m}\right]\right)\left(A_{n}-\mathbb{E}\left[A_{n}\right]\right)\right] / \xi  _{ m} \xi  _{ n},
\end{equation}

\begin{equation}
\label{align_12}
\mathcal{C}(A_{m} , A_{n})_{info}
=\sum_{a \in A_{m}^{(\boldsymbol{W}_{1})}} \sum_{b \in A_{n}^{(\boldsymbol{W}_{2})}} p(a, b) \log \left(\frac{p(a, b)}{p(a) p(b)}\right),
\end{equation}
\begin{equation}
\label{align_13}
\mathcal{C}(A_{m} , A_{n})_{\ell 2}
=
\left\|A_{m}^{(\boldsymbol{W}_{1})}-\boldsymbol{P} A_{n}^{(\boldsymbol{W}_{2})}\right\|^{2},
\end{equation}
where $A_{m}$ denotes the activation of unit $m$ with standard deviation $\xi $.
$p(a)$ denotes marginal probability distributions.
In addition, it is discovered that using post-activation is better than using pre-activation in some cases \cite{tatro2020optimizing}.
Besides the cost functions, Singh et al. \cite{singh2020model} use the optimal transport (OT) and Wasserstein barycenter to match the activations of different neural networks.
The transport map $\boldsymbol{T} \in \mathbb{R}^{(\mathrm{n} \times \mathrm{m})}$ transports neurons of $\boldsymbol{W}_{1} $ optimally to neurons of $\boldsymbol{W}_{2} $ in the same layer.
The permutation matrix and $\boldsymbol{T}$ have a similar function, which can be obtained as Eq.\eqref{align_14}:

\begin{equation}
\label{align_14}
\boldsymbol{T} \leftarrow  \mathrm{OT}\left(\mu, \nu, d_{s}\right),
\end{equation}
where $d_{s}$ denotes the support measure (reflect the $\ell2$ distance between activations here).
$\nu$ and $\mu$ are the probability measure.
This kind of methods based on OT lay the foundation for some recent work \citep{akash2022wasserstein,pena2022re,ainsworth2022git}.
Nevertheless, if the alignment problem is simply defined as linear problems, the second-order proximity of weights and the abundant edge information between channels could be ignored \citep{liu2022deep}.




\subsection{Weight Matching}

Instead of matching activation, we could alternatively align the models based on weight without data distribution.
First, the basic approaches of weight matching is also based on minimizing the cost function to obtain $\boldsymbol{P^{*}}$.
Singh et al. \cite{singh2020model} use the weights of the incoming edges to calculate support and probability measures to obtain the transport map $\boldsymbol{T}$ as Eq.\eqref{align_14}.
Ainsworth et al. \cite{ainsworth2022git} arrange the rows and columns of the modes to minimize the $\ell2$ distance between the weight vectors (restricted by ordinary least squares) as Eq.\eqref{align_15}:
\begin{equation}
\label{align_15}
\mathcal{C}(w_{1},w_{2})_{\ell2}=
\underset{}{}\left\|\operatorname{vec}\left(w_{1}\right)-\operatorname{vec}\left(\pi\left(w_{2}\right)\right)\right\|^{2}.
\end{equation}
It results in the sum of bilinear linear assignment problem (SOBLAP), which can be divided into sub-problems and solved by LAP.
Different from activation matching, weight matching is not affected by data distribution.
It means that all $\boldsymbol{P}$ need to be obtained by LAP, which is a complicated issue in essence.
And it is difficult to leverage the gradient-based optimization.
Pena et al. \cite{pena2022re} extend the scope of cost function to all differentiable objectives, such as a midpoint as Eq.\eqref{align_16} and random point between $w_{1}$ and $w_{2}$ as Eq.\eqref{align_17}:
\begin{equation}
\label{align_16}\mathcal{C}_{mid}\left(w_{1},w_{2}\right)=\mathcal{C}\left(\frac{w_{1}+\pi\left(w_{2}\right)}{2}\right),
\end{equation}
\begin{equation}
\label{align_17}
    \mathcal{C}_{random }\left(w_{1},w_{2}\right)=\mathcal{C}\left[(1-\alpha ) w_{1}+\alpha  \pi\left(w_{2}\right)\right],
\end{equation}
where $\alpha \sim  U(0,1)$.
Moreover, Sinkhorn operator $S_{\tau}$ is added to the LAP process and Sinkhorn Re-basin is shown as Eq.\eqref{align_18}:
\begin{equation}
\label{align_18}
    f^{(\ell )}(x,w)=
    \sigma\left[S_{\tau}\left(P^{(\ell)}\right) W^{(\ell)} S_{\tau}\left((P^{(\ell-1)})^{T}\right) f^{(\ell-1)}+S_{\tau}\left(P^{(\ell)}\right) b^{(\ell)}\right].
\end{equation}
It solves non-differentiable problems and can be applied to more scenarios, such as FL \cite{mcmahan2017communication}.
Based on Beta-Bernoulli Process (BBP) \cite{thibaux2007hierarchical}, Yurochkin et al. \cite{yurochkin2019bayesian} max the posterior of random variables $p_{i}$ that match neurons at any batch and the global neurons.
Hungarian algorithm can be used to solve this problem to obtained $P_{i}$.
In addition to minimizing the cost function, Wang et al. \citep{wang2020federated} regard the units of the model as a random permutation of global nodes based on the Beta-Bernoulli Process (BBP) \cite{thibaux2007hierarchical}
The permutation matrix can be obtained by BBP-MAP \citep{yurochkin2019bayesian}.
A simulated annealing (SA)-based method \cite{entezari2021role} searches for the valid permutations in the weight space Re-basin .
Due to the high cost, it unrealistic to be applied, especially for large models.
Stoica et al. \cite{stoica2023zipit} calculate merge matrix $P_{i}$ and unmerge matrix $\bar{P}_{i}$ to  fuse the models and unmerge operations, which can be applied within the model or across the models.
Instead of calculating the optimal matrix, Ainsworth et al. \cite{ainsworth2022git} optimize the approximate equivalent model $\tilde{w_{2}}$ iteratively and keep looking for the closest equivalent model until convergence, which minimizes $\mathcal{L}$ as Eq.\eqref{align_19}:

\begin{equation}
\label{align_19}
    \min _{\tilde{w}_{2}} \mathcal{L}\left(\frac{1}{2}\left(w_{1}+\operatorname{proj}\left(\tilde{w}_{2}\right)\right)\right),
\end{equation}
where projection operations can be solved by straight-through estimator (STE), which is expensive in practic.
Based on Gromov-Wasserstein barycenter (GWB) \cite{peyre2019computational}, Akash et al. \cite{akash2022wasserstein} update the coupling matrix $\Pi$ and $W $ alternately to optimize Gromov-Wasserstein barycenter distance until convergence.
Let $k$ be the number of nodes, the final aligned model can be obtained as Eq.\eqref{align_20}:
\begin{equation}
\label{align_20}
W^{\ell} \leftarrow k^{\ell} k^{\ell-1} \frac{1}{\mathbb{1}_{k^{l-1}} \mathbb{1}_{k^{l}}^{T}} \frac{1}{n} \sum_{i=1}^{n} \Pi^{\ell}_{i} W^{\ell}_{i}\left(\Pi^{\ell-1}_{i}\right)^{* T}
\end{equation}
Moreover, recent research \cite{wang2020federated,yurochkin2019bayesian} proposes to alternate for a number of iterations between finding an alignment and retraining to minimize the loss barriers between SGD minimas.

Furthermore, another significant approach of alignment is graph matching (GM) \cite{loiola2007survey}, which aims to match nodes in the graph using structural characteristics in the graph.
Since network channels and weight can be treated as nodes and edges, the alignment issues could be turned into GM \cite{yan2020learning,liu2022deep}.
General approaches could use Bipartite semi-matching or  Bipartite matching \cite{lam2021model,li2015convergent} to solve GM.
Liu et al. \cite{liu2022deep} propose graduated assignment model fusion (GAMF) \cite{wang2020clustering} uses second-order similarity of model weights to align neurons build on gradient assignment as Eq.\eqref{align_21}:

\begin{equation}
\label{align_21}
\max_{P} = \sum_{i=0}^{d_{\Sigma}-1} \sum_{j=0}^{d_{\Sigma}-1} \sum_{a=0}^{d_{\Sigma}-1} \sum_{b=0}^{d_{\Sigma}-1} \boldsymbol{P}_{[i, j]} \boldsymbol{K}_{[i, j, a, b]} \boldsymbol{P}_{[a, b]} \\,
\end{equation}
where $d_{\sum}$ denotes the sum of dimensions,$\boldsymbol{K}$ denotes affinity tensor that calculate the affinity between the edges $(i,a)$ and $(j, b)$.
The problem can be transformed into QAP by unifying the relationships of nodes and edges into a incidence matrix.
In contrast, multi-graph matching (MGM) \cite{yan2015multi,jiang2020unifying,leonardos2017distributed} ensures that the matching of two graphs is not affected by another graph, and it applies to the alignment of multiple models.
Further, Uriot et al. \cite{uriot2020safe} explore merging models that take into account more possible permutations.

\subsection{Discussion}
%
Alignment makes the models more similar by adjusting the parameters of the models, which can improve the information sharing between the models, and thus improve the generalization ability of the fused model.
In addition, alignment helps improve the performance and robustness of the model on complex tasks.
However, alignment methods face the problems of slow combinatorial optimization. Alignment requires additional computational overhead to adjust the model's parameters, which can lead to a more complex and time-consuming training process, especially in large depth models \cite{singh2020model,liu2022deep}.

In summary, alignment can improve the consistency and overall effect between different models.
With the diversification of DL application scenarios, alignment will become one of the key methods to optimize deep model fusion, improve generalization ability.
In the future, alignment could play a role in areas such as transfer learning, domain adaptive \cite{gal2022stylegan}, knowledge distillation, etc.
For example, alignment can reduce the differences between source and target domains in transfer learning, improve the learning on new domains.

\section{Weight Average}


``Weight average" combines multiple weights of networks for the final model with better performance, robustness and generalization.
It is also known as vanilla average \cite{singh2020model}, weight summation \cite{leontev2020non}, as shown in Eq.\eqref{wa_22}:
\begin{equation}
\label{wa_22}
   \sum\lambda  _{i} W _{i},
\end{equation}
where each model is assigned a weighted parameter $\lambda_{i}$ that controls how much it contributes to the fused model.
However, different from alignment or mode connectivity, the pre-conditions of WA are relatively strict.
For example, the original models must share part of the training trajectory or located in the same basin \cite{izmailov2018averaging,li2019fedmd}, etc.
It means that the final model can benefit from all models when the weights are similar enough but have certain differences \cite{jolicoeur2023population}.
In a flat basin, the solutions tend to demonstrate good performance.
Conversely, points in narrow regions are easily accessible to energy-barriers, resulting in increased losses \cite{neyshabur2020being}.
Previous sections focus on transporting solutions from different regions to the same basin through mode connectivity or alignment.
This section will focus on the fusion of convex combinations of solutions in the same basin, which makes the merged solution closer to the midpoint (optima) of the basin with better generalization performance than endpoints, such as SWA \cite{izmailov2018averaging}, model soup \cite{wortsman2022model}, etc.
The models discussed in this section includes the following cases:
\begin{itemize}
    \item      Multiple similar models with certain differences.
    \item  Multiple models after appropriate fine-tuning on foundation models (e.g., model soup, model arithmetic, etc.).
    \item     Multiple checkpoints from networks with the same architectures and sharing part of the training trajectory (e.g. SWA \cite{izmailov2018averaging}, tail average \cite{neu2018iterate}, etc.).

\end{itemize}
Accordingly, in this section, we review two-fold approaches of weight average  ``Weight average" and ``Average in subspace".
Next, we introduce representative approaches of WA ``Model soup" , ``Model arithmetic" and ``SWA".
The representative approaches are listed in Table   \ref{table4}.


\subsection{Weight Average}
Because of the high redundancy of neural network parameters, there is usually no one-to-one correspondence between weights of different neural networks.
Accordingly, there is usually no guarantee that WA will perform well by default.
For trained networks with widely varying weights, the vanilla average performs poorly \cite{singh2020model}.
From a statistical point of view, WA allows the individual model parameters in the model to be controlled, which reduces the variance of the final model, resulting in a reliable effect on regularization properties and output result \cite{guo2023stochastic,neu2018iterate}. 

\begin{table}
\caption{Summary of representative methods and formulas of weight average.
}
\label{table4}
\vspace{0.5em}
\begin{tabular}{l l l l} 
\toprule
 Method &  Method & Ref. & Introduction  \\ 
\midrule
   choose the best  & \cite{wortsman2022model}  & $\operatorname{argmax}_{i} ValAcc(W _{i}))$ & \makecell[l]{simple but without\\ the advantages of WA} \\
vanilla average &   \cite{singh2020model} & $W=\sum\lambda  _{i} W _{i}$& often have bad performance \\

Fisher & \cite{matena2022merging} & $ W=\frac{\sum \lambda_{i} f_{i} w_{i}}{\sum \lambda_{i} f_{i}}$ &\makecell[l]{maximize joint likelihood\\ of the posterior distribution} \\
RegMean & \cite{jin2022dataless} & $W=\left(\sum X_{i}^{T} X_{i}\right)^{-1} \sum\left(X_{i}^{T} X_{i} W_{i}\right)$ & \makecell[l]{minimize differences\\between merged model\\ and individual models} \\
MLP fusion & \cite{wei2023ntk}  &$W  =\sum\left[\sigma\left(\boldsymbol{X} W_{1, \cdot, i}+\boldsymbol{b}_{1, i} \mathbf{1}\right) W_{2, i, \cdot}\right]+\mathbf{1} \boldsymbol{b}_{2}^{T}$& cluster the sub MLPs via NTK\\
BTM  & \cite{li2022branch} & $W=\sum\lambda  _{i} W _{i}$&\makecell[l]{combine the expert\\ LMs on different\\ domains of corpora} \\
%

PAPA & \cite{jolicoeur2023population}&$W \leftarrow \text { Averaging }(W, N)$&\makecell[l]{average a mass of \\models trained on \\slightly different datasets}\\
 ratatouille & \cite{rame2023model-ratatouille}  & $W=\sum \lambda _{i}\left(w_{i}, \phi_{featurizer}\right)$& \makecell[l]{use the diversity of \\auxiliary tasks to enrich\\ the diversity of weights}\\

Lookahead & \cite{zhang2019lookahead}  & $W_{slow,t+1}   =ema(W_{fast})+(1-\alpha)^{t} W_{slow,0}$ & \makecell[l]{combine fast weight\\ and slow weight}\\

 SMA & \cite{NEURIPS2022_372cb780}  & $W=\frac{t-t_{0}}{t-t_{0}+1} \cdot W_{t-1}+\frac{1}{t-t_{0}+1} \cdot W_{t}$&  \makecell[l]{conduct tail average\\in later stages} \\
WiSE-FT& \cite{wortsman2022robust}&$  W=(1-\lambda) \cdot W_{0}+\lambda \cdot W_{ft}$ & \makecell[l]{ interpolation models\\ before and after fine-tuning}\\

EWC & \cite{leontev2020non} & $W=\frac{H_{1} W_{1}+H_{2} W_{2}}{H_{1,}+H_{2}}$&\makecell[l]{minimize the weight\\ variation required \\by model fusion} \\

\makecell[l]{gradient\\information }&\cite{gao2022revisiting}&$ W=\sum \lambda_{i} W_{i}- \frac{1}{i} \nabla X_{gradient}$&\makecell[l]{exploit other info-\\rmation in checkpoints}\\

PAINT & \cite{ilharco2022patching} & $W_{\text {patch }}=\left(1-\sum \lambda_{i}\right)  W_{\mathrm{0}}+\sum \lambda_{i}W_{\mathrm{ft}}$ &\makecell[l]{linear interpolation of the\\models before and after fine-\\tuning on tasks to be patched} \\
HiPro& \cite{liu2023hierarchical}& $  w_{i}=\frac{\sum w\left(\boldsymbol{p}_{i}\right) \mathbb{I}\left(\tau_{j} \in \mathcal{T}_{i}\right)}{\sum \mathbb{I}\left(\tau_{i} \in \mathcal{T}_{i}\right)} $&\makecell[l]{obtain classifier weights\\ from the individual \\prompt and the shared prompt
}\\

EWR & \cite{daheim2023elastic}  & $ W=\frac{\lambda_{0} \cdot \mathrm{f}_{W_{0}} \cdot W_{0}-\lambda_{1} \cdot \mathrm{f}_{\tau_{1}} \cdot \tau_{1}+\lambda_{2} \cdot \mathrm{f}_{\tau_{2}} \cdot \tau_{2}}{\lambda_{0} \cdot \mathrm{f}_{W_{0}}+\lambda_{1} \cdot \mathrm{f}_{\tau_{1}}+\lambda_{2} \cdot \mathrm{f}_{\tau_{2}}}$&  \makecell[l]{use Fisher to combine \\model with task vectors}\\
experts merging & \cite{jang2023exploring}  & $W=W_{\text {pre }}+\left(\sum \lambda_{i} \tau_{i}\right)$&\makecell[l]{find efficient fine-tuning\\via adapters to train experts}\\


 

\bottomrule
\end{tabular}
\end{table}

First, the weights of neural networks could be merged directly.
Generally speaking, the linear interpolation of two well-trained model in different regions does not necessarily generate a well-performing model because of the nonlinear structure of neural networks \cite{neyshabur2020being}.
However, for the solutions before and after fine-tuning are usually within a basin \cite{ilharco2022patching,wortsman2022robust}, the linear interpolation of the solutions could improve he accuracy of fused model and the robustness of the distribution shift as Eq.\eqref{wa_23}:

\begin{equation}
\label{wa_23}
    W=(1-t) \cdot W_{0}+t \cdot W_{ft}.
\end{equation}
In addition to simple linear interpolation, the fusion of weights could be transformed into another mathematical form of aggregation.
Matena et al. \cite{matena2022merging} propose Fisher merging, which regards model fusion as a approximately maximization of the joint likelihood of the posterior distribution over parameters.
It use the Fisher information $F_{i}$ of the model as the posterior precision matrix to perform a Laplacian approximation, so as to obtain the Gaussian approximation $ \log p\left(w \mid w_{i}, F_{i}\right)$ of the posterior distribution as Eq.\eqref{wa_24}:
\begin{equation}
\label{wa_24}
  \max_{w} \sum\lambda_{scale} \log p\left(w \mid w_{i}, F_{i}\right),
\end{equation}
where $\lambda_{scale}$ denotes model scalar hyperparameters.
Jin et al. \cite{jin2022dataless} tend to minimize the  $\ell2$ distance between the merged model and other multiple models trained on different datasets $\left\langle X_{i}, Y_{i}\right\rangle$, which is called Regression Mean (RegMean).
Accordingly, the optimization problem can be converted into linear regression problem as Eq.\eqref{wa_25}:
\begin{equation}
\label{wa_25}
    \min _{W}\left\|W^{T} X_{1}-W_{1}^{T} X_{1}\right\|^{2}+\left\|W^{T} X_{2}-W_{2}^{T} X_{2}\right\|^{2}.
\end{equation}
Compared with Fisher average \cite{matena2022merging}, RegMean obtain the inner product matrix of the linear layer input in the forward pass process, which improves the efficiency of the operation.
Besides, Wei et al. \cite{wei2023ntk} regard each layer of multi-layer perceptrons (MLPs) as the distribution of corresponding weights.
The sub-MLPs can be clustered by neural tangent kernel (NTK) approximating, which can be solved with GWB \cite{peyre2019computational}.
Moreover, other works choose to average the weights of multiple experts\cite{li2022branch} or leverage Bayesian algorithm \cite{yurochkin2019bayesian} to improve the generalization and efficiency.

Also, some recent work focuses on increasing the diversity of models with well-behaved and varieties of weights.
PopulAtion Parameter Averaging (PAPA) \cite{jolicoeur2023population} start at the same initialization and train each models on a slightly different data set (e.g., data orderings, augmentations, regularizations, etc.), averaging these models every few epochs.
It is equivalent to training a larger batch size, helping to improve the generalization of the model \cite{hoffer2017train}.
Further, another possible interpretation is that PAPA fuse the models under better initial conditions by improving the cosine similarity between networks (29$\%$-61$\%$ to 95$\%$-99$\%$), which is similar to some work on alignment \cite{ainsworth2022git}.
Based on the idea of maximizing the diversity of weights, Rame et al. \cite{rame2023model-ratatouille} fine-tune the base model for multiple times on different auxiliary tasks and re-fine-tune these auxiliary weights so as to obtain a variety of weights.
Gao et al. \cite{gao2022revisiting} utilize development data and softmax normalized logarithm with temperature to adjust the parameters.
The models are re-parameterized and updated iteratively to ensure normalization, which could reduce overfitting and increase robustness.
In addition, the mean of gradient information $\nabla X_{gradient}$ could be used to optimize the WA \cite{gao2022revisiting}.
Let $\eta$ be step size. The merged model is shown as Eq.\eqref{wa_26}:
\begin{equation}
\label{wa_26}
    W=\sum\lambda_{i} W_{i}- \eta \nabla X_{gradient}.
\end{equation}


Next, from the perspective of iterative averaging, we can average the weights at different times during the training process of the same or architecturally identical model \cite{gao2022revisiting,liu2018comparable,leontev2020non}.
It reduces the variance and updates the model more smoothly but need to share a portion of the training history \cite{smith2017investigation}.
Early iterative average has the problem of convergence rate \cite{polyak1990new,ruppert1988efficient} , especially for high-dimensional problems.
Then, geometric Polyak-Ruppert \cite{neu2018iterate} use the weight average instead of uniform average, and its weights decay geometrically.
It uses regularization properties (control deviation characteristics of corresponding SGD estimators) to produce stable fusion results.
Geometric Polyak-Ruppert helps to capture the overall trend of the gradient when training conditions are poor.
In contrast, tail average \cite{jain2018parallelizing} is more appropriate when data conditions are good.
Tail average average the weights of each iteration during the last period of the training, which can prevent large fluctuations of parameter in the late stage.
When the model is close to convergence, and the tail part of the gradient may contain information closer to the real gradient. 
Moreover, a great deal of factors (e.g., decaying step size \cite{polyak1990new}, constant step size \cite{nemirovski2009robust}, form of linear interpolation, etc.) in the iteration average will affect the final result.
%
Further, checkpoint average \cite{huang2017snapshot,chen2017checkpoint,wang2021boost,liu2018comparable} uses checkpoints from the same training run.

Nevertheless, simple coordinate-wise weight average may result in poor performance.
Hierarchical aggregation improves model performance by combining parameters from multiple models at different layers or structures.
The network architecture suitable for a specific aggregation approach has certain limitations \cite{yurochkin2019bayesian,matena2022merging}, so recursively processing layers with matching averages may affect the final performance.
Wang et al. \cite{wang2020federated} propose a hierarchical aggregation scheme. 
The server obtains the first layer weight of the model and broadcasts it to the client, which continues to train all the layers with the matching layers frozen.
And then repeats the procedures until the last layer before aggregation.
Hierarchical Prompt learning (HiPro) \cite{liu2023hierarchical} constructs a hierarchical task tree and average classifier weights generated from the global prompt and individual prompt $\boldsymbol{p}_{i}$.
The classifier average weights on i$_{th}$ task $\tau_{i}$ is shown as Eq.\eqref{wa_27}:
\begin{equation}
\label{wa_27}
    W_{i}=\frac{\sum W\left(\boldsymbol{p}_{i}\right) \mathbb{I}\tau_{j} }{\sum \mathbb{I}\tau_{i}},
\end{equation}
where $\mathbb{I}$ is the indicator function.
Its layer-wise structure helps to gain knowledge of diverse granularity.
Some other work \cite{shu2021zoo,qin2022exploring} propose layer-wise, module-wise and matrix-wise structure of parameter division, which reduces the cost of calculation and storage and inspires more directions of WA.


Further, WA is often used to weight scaling rules, which average the predictions of the distribution over the weights \cite{neklyudov2018variance,srivastava2014dropout}.
To ensure the efficiency of model average, Akhlaghi et al. \cite{akhlaghi2018knowledge} propose that activation functions should restrict postsynaptic activity to a limited range(e.g., sigmoid, hyperbolic tangent, etc.).
Leontev et al. \cite{leontev2020non}propose other constraints that network generates presynaptic activity in the presence of native features and the mean of the weights’ probability distribution should be zero \cite{blundell2015weight}. 
In addition, for heterogeneous issue, they can be approximated by introducing additional zero-valued weights \cite{leontev2020non}.


\subsection{SWA}

\begin{table}\centering
\caption{Comparison of characteristics of Weight Averge methods based on SWA
}
\label{Table5}
\vspace{0.5em}
\begin{tabular}{ l l l} 
\toprule
  Method & Ref. & Introduction   \\ 
\midrule

 SWA & \cite{izmailov2018averaging} & \makecell[l]{weight can be manually weighted after training}  \\
 EMA & \cite{wu2021peer,karita2021comparative} &  \makecell[l]{smoothing model weights}   \\
EWA & \cite{huang2023experts}&  \makecell[l]{improve the performance without increasing inference delay and weights} \\
 SWAG & \cite{maddox2019simple}  & \makecell[l]{approximate Bayesian model averaging in Bayesian DLand achieves the \\state-of-the-art uncertainty calibration results in various settings} \\

 SWALP & \cite{yang2019swalp}  & \makecell[l]{match the performance of SGD training with quantized parameters}\\

 SWAP &  \cite{gupta2020stochastic} & \makecell[l]{speed up the training of NN by using large batch size}\\

 SWAD & \cite{cha2021swad}  & \makecell[l]{ improve the OOD generalization performance of DNNs}\\

LAWA & \cite{kaddour2022stop} & \makecell[l]{record up-to-date checkpoints at the end of each epoch} \\

 HWA & \cite{gu2023hierarchical} & \makecell[l]{combine online WA and offline WA}\\
 PSWA & \cite{guo2023stochastic} & \makecell[l]{find high-quality local optima quickly}\\
 TWA & \cite{li2023trainable} & \makecell[l]{conduct subspace training to implicitly adjust\\ the averaging coefficients and approach better to the minima}\\

\bottomrule
\end{tabular}

\end{table}

\begin{figure}
	\centering 
\includegraphics[width=1\textwidth, angle=0]{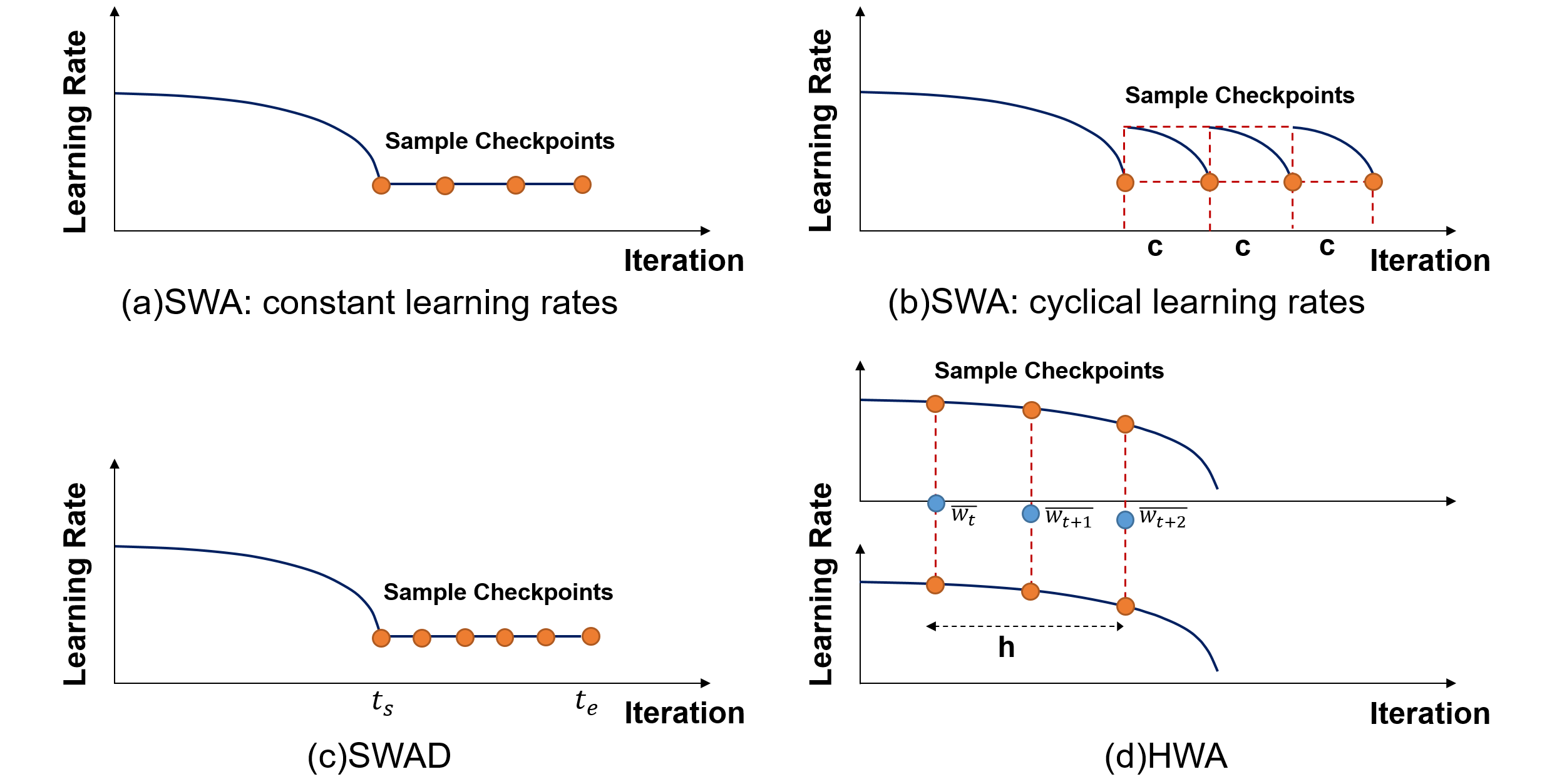}	
	\caption{\textbf{Comparison of sampling and learning rate schedule of different SWA related methods.} (a) SWA: constant learning rates. (b)SWA: cyclical learning rates $\textbf{c}$. (c)SWAD: sample densely. (d)HWA: leverages both online and offline WA, which sampled at different synchronization cycles with a slide window of length $h$, i.e. $\overline{\overline{w_{i}}}=\frac{\sum_{t=i-h+1}^{i} \overline{w_{t}}}{h}$.}
  \label{figure4}
\end{figure}

Inspired by Fast Geometric Ensembling (FGE) \cite{garipov2018loss} and checkpoint average \cite{liu2018comparable}, Izmailov et al. \cite{izmailov2018averaging} utilize a constant or periodic learning rate to average multiple points along the SGD trajectory, which is regarded as SWA.
SWA improves the training on a series of important baslines, providing better time scalability.
Instead of training a set of collected models like vanilla fusion, SWA trains a single model to find smoother solutions than SGD.
In Table \ref{Table5}, we list the approaches related to SWA .
Also, SWA can be applied to any architecture or datasets and demonstrate performance than snapshot ensemble (SSE) \cite{huang2017snapshot} and FGE.
At the end of each cycle, the SWA model $W_{SWA}$ is updated by averaging the newly obtained weights over the existing weights , as shown in Eq.\eqref{wa_28}:
\begin{equation}
\label{wa_28}
    W_{\text {SWA }} \leftarrow \frac{W_{\text{SWA}} \cdot n+W}{n+1}.
\end{equation}
Nevertheless, SWA can only average the points near the local optimal point, and finally get a relatively minimum value rather than accurately approximating the optima.
Also, the final input sample deviation could be large or insufficient due to some factors (e.g., poor convergence at early stage, large learning rate, fast weight change rate, etc.), which results in bad overall effect.
There is a good deal of work tends to change the sampling schedule of SWA.
For example, SWA-Densely (SWAD) \cite{cha2021swad} uses more dense sampling points to solve the problem of insufficient random weights.
Periodic-SWA (PSWA) \cite{guo2023stochastic} is initialized during the early stage of the operation of SGD instead of in the late convergence phase like SGD.
Latest weight averaging (LAWA) \cite{kaddour2022stop} averages only the checkpoints collected at the end of each epoch given the large weight variation during the initial training phase.
In Figure \ref{figure4}, we summarize several ways to optimize SWA with different sampling schedules.
Some work based on SWA optimizes the polymerization process to gain competitive outcome.
SWA in Low-Precision (SWALP) \cite{yang2019swalp} tends to reduce the influence of quantization noise and low learning rate so as to converge to the optima.
SWA-Gaussian (SWAG) \cite{maddox2019simple} obtains Gaussian distribution from the points of SWA, then average the Bayesian models sampled from the distribution.
Trainable Weight Averaging (TWA) \cite{li2023trainable} adjusts the fuse solution to better approximate the minimum by projecting the gradient onto the subspace as Eq.\eqref{wa_29}:
\begin{equation}
\label{wa_29}
    W_{\mathrm{TWA}} \leftarrow W_{\mathrm{TWA}}-\eta_{l} \boldsymbol{B}\left(\boldsymbol{B}^{\top} g\right),
\end{equation}
where $\boldsymbol{B}$ denotes the matrix of a set of base vectors. $\eta_{l}$ is the learning rate.
$g$ is the gradient.
TWA could eliminate errors caused by static averaging in full parameters space.
Different from the above approaches, Hierarchical Weighted Average (HWA) \cite{gu2023hierarchical} combines online and offline WA into a common training framework, 
Online WA is designed to speed up convergence, offline WA tends to increase generalization performance.
HWA tends to combines the advantages of both.
Similar to SWA, Exponential Moving Average (EMA) \cite{polyak1992acceleration,szegedy2016rethinking} is often used to smooth the model weights in order to reduce the noise and volatility of update on weights as Eq.\eqref{wa_30}:
\begin{equation}
\label{wa_30}
    W_{EMA} \leftarrow \lambda_{d} W_{EMA}+(1-\lambda_{d}) W,
\end{equation}
where $\lambda_{d}$ denotes the decay rate ($\approx 0.99$).
Some recent work \cite{caron2021emerging} combines KD with EMA, using the weights of EMA (e.g., student models \cite{tarvainen2017mean} or branches  \cite{wu2021peer}) as teacher models to transfer knowledge.
Huang et al. \cite{huang2023experts} replace the networks with Mixture-of-Experts (MoEs) \cite{shazeer2017outrageously} and perform the EMA on MOEs at the end of each iteration.
It can be used to improve generalization on a variety of 2D and 3D vision tasks on ViT architectures.
Arput et al. \cite{NEURIPS2022_372cb780} propose simple moving average (SMA), which conducts moving average in the later stages of training (after $t_0$ rounds of iteration) to improve the performance in out of domain as Eq.\eqref{wa_31}:
\begin{equation}
\label{wa_31}
    \hat{W}_{t}=\left\{\begin{array}{ll}
W_{t} &  t \leq t_{0} \\
\frac{t-t_{0}}{t-t_{0}+1} \hat{W}_{t-1}+\frac{1}{t-t_{0}+1}  W_{t}& t \leq t_{0}
\end{array}\right  .
.
\end{equation}
Lookahead algorithm \cite{zhang2019lookahead} interpolates fast and slow weights linearly from the optimized trajectory.
 as Eq.\eqref{wa_32}:
\begin{equation}
\label{wa_32}
w_{slow,t+1}   =t\left[w_{fast,t}+(1-t) w_{fast,(t-1)}+\ldots+(1-t)^{(t-1)} w_{fast,0}\right]+(1-t)^{(t)} w_{slow,0}.
\end{equation}
The trajectories of fast weights $w_{fast,t}$ are updated quickly by EMA in the direction of low curvature.
The slow weights $w_{slow,t} $ smooth the oscillations by interpolating the parameters.
Lookahead reduces variance, speeds up convergence and bring the results closer to the regions with high test accuracy.

\subsection{Model Soup}
\begin{table}[h]\small
\centering
\caption{Summary of different methods of Model Soup.
}
\label{Table6}
\vspace{0.5em}
\begin{tabular}{ l l l l} 
\toprule
  Method & Ref.  & Introduction \\ 
\midrule



 Uniform Soup  &  \cite{wortsman2022model} & \makecell[l]{average the fine-tuned models directly} \\

 Greedy Soup  &  \cite{wortsman2022model} & \makecell[l]{simple operation, good performance}\\
 
 Learned Soup  &  \cite{wortsman2022model} &  \makecell[l]{high memory cost (especially in large-scale model) }\\

 Sparse Soup  &  \cite{zimmer2023sparse} & \makecell[l]{flexible and transparent alleviates scaling issue}
 \\
   Adversarially-robust soup & \cite{Croce_2023_CVPR}  & \makecell[l]{improve adversarial robustness to multiple threat models}
   \\

   Rewarded Soup & \cite{rame2023rewarded}  &   \makecell[l]{merge networks according to user preferences}\\

 DiWA & \cite{rame2022diverse} & \makecell[l]{leverage the full potential of WA } \\
  Fed Soup &\cite{chen2023fedsoup} & \makecell[l]{alleviate overfitting and seek flat minima}\\
  Adapter Soup &\cite{chronopoulou2023adaptersoup} &\makecell[l]{maintain performance on in-domain and new domains.}\\

 \bottomrule
\end{tabular}

\end{table}

Model soup \cite{wortsman2022model} refers to the method of averaging the models fine-tuned with different hyperparameters.
It is simple but effective, achieving an accuracy of 90.94$\%$ on the ImageNet-1K, which surpasses the previous work on CoAtNet-7 (90.88$\%$) \cite{dai2021coatnet} and ViT-G (90.45$\%$) \cite{zhai2022scaling}.
In Table \ref{Table6}, we summarize the different soups.
Model soup reduces the inference time required for ensemble learning $\frac{1}{n} \sum_{i=1}^{n} f\left(x, W_{i}\right)$ \cite{sagi2018ensemble}, which includes three soups as follows:
The uniform soup average all the weights of the model directly $f\left(x, \frac{1}{n} \sum_{i=1}^{n} W_{i}\right)$.
The greedy soup adds the models to the soup in sequence, keeping the model in the soup if the accuracy of the verification set does not decrease, which performs the best of the three soups as Eq.\eqref{wa_33}:
\begin{equation}
\label{wa_33}
\text { ingredients } \leftarrow \text { ingredients } \cup\left\{W_{i}\right\}
\text{  if  }
   Acc\left(\right.  Avg  \left(\right.  ingredients  \left.\left.\cup\left\{W_{i}\right\}\right)\right) \ge Acc(Avg(ingredients))  .
\end{equation}
Greedy soups \cite{wortsman2022model} can be regarded as another form of SWA \cite{izmailov2018averaging}, which take a subset of weights as the input sample of the SWA.
The learned soup removes the order rules of greedy soup, learns the mixing coefficient $\lambda_{mix}$ and temperature scaling parameters $\lambda_{temp}$ for each component in the verification set, and optimizes the soup by gradient-based optimization as Eq.\eqref{wa_34}:
\begin{equation}
\label{wa_34}
    \underset{\lambda_{mix} \in \mathbb{R}^{k}, \lambda_{temp} \in \mathbb{R}}{\arg \min } \sum_{j=1}^{n} \ell\left(\lambda_{temp} \cdot f\left(x_{j}, \sum_{i=1}^{n} \lambda_{mix,i} W_{i}\right), y_{j}\right) . 
\end{equation}
The adversarially-robust model soup \cite{Croce_2023_CVPR} moves the convex hull of parameters of each classifier to adjust the weights of soup, in order to balance the robustness to different threat models and adapt to potential attacks.
%
Based on reinforcement learning from human feedback (RLHF),  rewarded soup \cite{rame2023rewarded} fine-tunes the models according to the diverse rewards.
It selects the proper interpolating coefficients $\left \{\lambda_{i}^{j}\right  \}_{i=1}^{N} $ form $N$-simplex that maximize the reward $\hat{R}$ as Eq.\eqref{wa_35}:
\begin{equation}
\label{wa_35}
    \operatorname{argmax}_{j=1}^{n} \hat{R}\left(\sum_{i=1}^{N} \lambda_{i}^{j} W_{i}\right).
\end{equation}


\subsection{Model Arithmetic}

Different from traditional single-task learning, MTL is a kind of joint learning.
The multiple tasks are learned in parallel so as to take advantage of data resources for different tasks \cite{zhang2018overview,donyehiya2022cold}.
In general, MTL could be regarded as a parameter sharing, or ensemble \cite{dimitriadis2023pareto}, that can include major information of multiple individual tasks.
In the process of MTL, participants fine-tune the latest model on the corresponding task in each iteration.
The multiple fine-tuned models are merged to produce the final model or base model for the next iteration \cite{choshen2022fusing,donyehiya2022cold}.
The general fusion method adopted in MTL is linear combination.
Patching with interpolation (PAINT) \cite{ilharco2022patching}combines fine-tuning and initial model so as to improve performance for specific task while also maintaining accuracy for other tasks.
PAINT reduces the time of migration and adaptation between multi-tasks.
HiPro \cite{liu2023hierarchical} explore the shared information from a plenty of tasks via hierarchical structure, which adapts pre-trained vision-language models (VLMs) to multiple downstream tasks. 
In addition, there are some other approaches group similar tasks could together, which is conducive to obtain shared model parameters conveniently \cite{standley2020tasks,fifty2021efficiently,maninis2019attentive,duong2015low}.
Moreover, recent work set up metrics to measure the performance of the shared model, such as, uncertainty to weight tasks \cite{kendall2018multi}, loss weighting strategies \cite{leang2020dynamic}, etc.
Huang et al. \cite{huang2023lorahub} introduce Low-rank adaptations Hub (LoraHub), a framework that ensembles LoRA modules trained on different given tasks, which improves flexibility and scalability in MTL.

\begin{figure}
	\centering 
	\includegraphics[width=1\textwidth, angle=0]{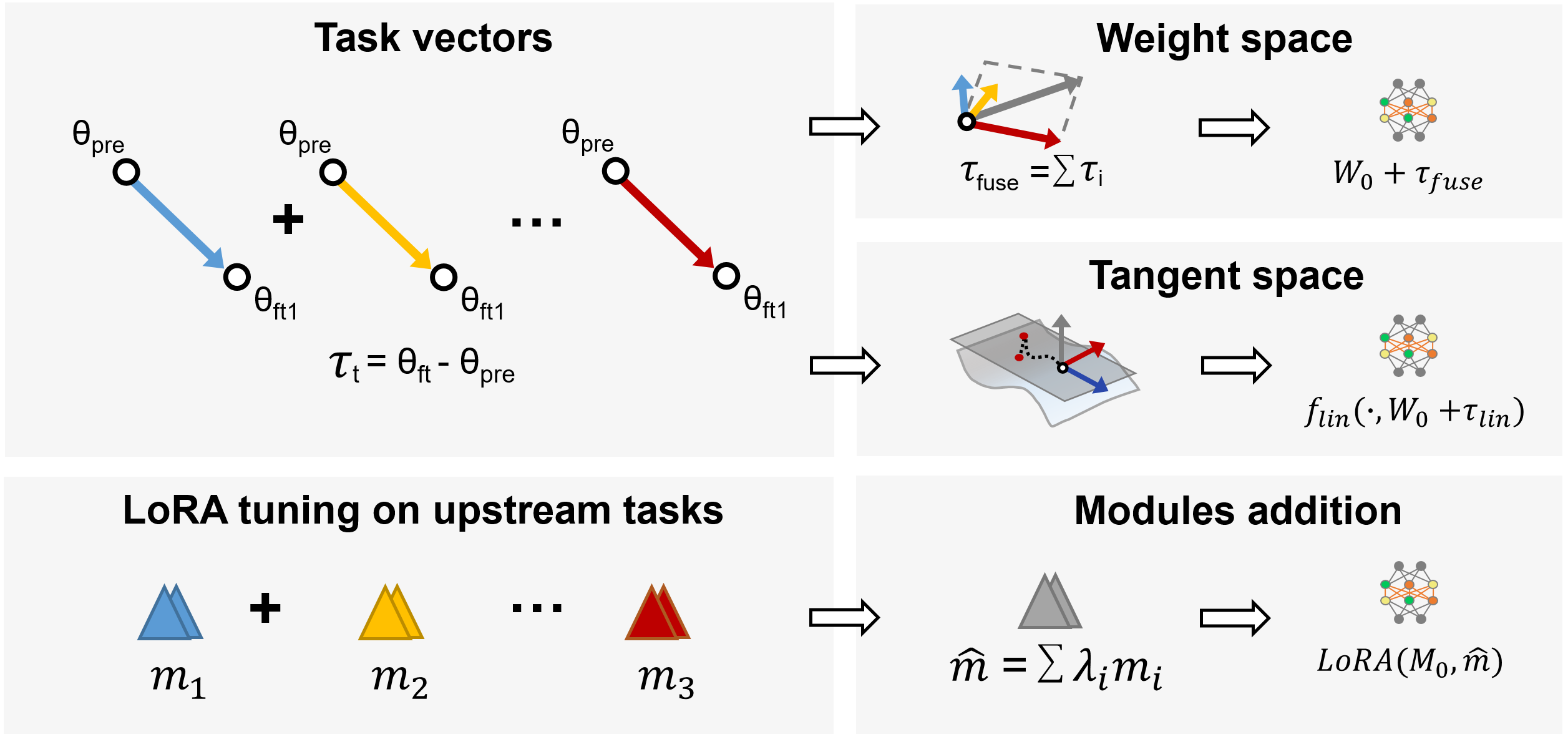}	
	\caption{The flow chart of Task Arithmetic and LoRA Hub\cite{huang2023lorahub} in multi-task scenarios.
 } 
	\label{figure_5}%
\end{figure}

In MTL, the pre-trained model and tasks vectors (i.e.,  $\tau_{i}=W_{ft}-W_{pre}$, the difference between the pre-trained model and the fine-tuned model) are combined to result in better performance on all tasks.
Based on this observation, task arithmetic \cite{ilharco2022editing} improves the performance of the model on tasks by adding and linear combination of fine-tuning task vectors, which has become a flexible and efficient method for editing pre-trained models directly as Figure \ref{figure_5}.
Ortiz et al. \cite{ortiz2023task} fine-tune the pre-trained model in the tangent space and provide a more reliable way to edit the pre-trained model by NTK linearization \cite{jacot2018neural}, improving the task algorithm significantly by reducing the accuracy gap of individual tasks \cite{sinitsin2020editable}.
Similar to the task algorithm, Daheim et al. \cite{daheim2023elastic} propose elastic weight removal (EWR), which calculates difference vectors between original models and expert models (fine-tuned on positive behaviours).   
EWR uses Fisher \cite{matena2022merging} to average the weights of the model and task vectors as Eq.\eqref{wa_37}:
\begin{equation}
\label{wa_37}
    W=\frac{\lambda_{0} \cdot \mathrm{f}_{W_{0}} \cdot W_{0}-\lambda_{1} \cdot \mathrm{f}_{\tau_{1}} \cdot \tau_{1}+\lambda_{2} \cdot \mathrm{f}_{\tau_{2}} \cdot \tau_{2}}{\lambda_{0} \cdot \mathrm{f}_{W_{0}}+\lambda_{1} \cdot \mathrm{f}_{\tau_{1}}+\lambda_{2} \cdot \mathrm{f}_{\tau_{2}}}
\end{equation}
It combine Fisher merging and task arithmetic to preserve positive behaviour in the model while removing the negative behaviours.
Jang et al. \cite{jang2023exploring} add the sum of the vectors  of a particular experts to the pre-trained language models (LMs) so as to cover the information from multiple experts trained on diverse tasks.
In sum, the essence of task arithmetic is to preserve pre-trained model behavior, thereby avoiding expensive joint fine-tuning on multiple tasks \cite{ilharco2022patching,li2022branch,wortsman2022robust}.

\subsection{Average in Subspace}
Due to the large dimension of conventional full-parameter space, from tens of millions to hundreds of millions of dimensions, model fusion in subspace will constrain the training trajectory in a low-dim subspace so as to reduce the loads and difficulties \cite{li2018measuring,li2022low,gressmann2020improving,li2022trainable}. 
In general, DNNS are over-parameterized.
The Low-dimensional Trajectory Hypothesis \cite{li2022low} speculates that the intrinsic dimension required for network training is not as large as the number of parameters given. 
The parameters trained and redundant information are reduced in a subspace, which could accelerate the convergence speed and improves robustness and generalization \cite{li2022trainable,li2022low}.
Recently, Li et al. \cite{li2023trainable} demonstrate that each point in the subspace corresponds to a base.
The linear combination of bases is equivalent to a weighted average  \cite{li2018measuring}.
Liu et al. \cite{liu2021sparse} extract submodels by sparse training to fuse multiple local models in low-dimensional subspace.
Leontev et al. \cite{leontev2020non} propose Elastic Weight Consolidation (EWC) to average the models in multi-dimensional space as Eq.\eqref{wa_37}:
\begin{equation}
\label{wa_37}
    W=\frac{H_{1} W_{1}+H_{2} W_{2}}{H_{1,}+H_{2}},
\end{equation}
where $H_{i}=\mathbb{E}_{p(x \mid w)}\left[\left(\frac{\partial L}{\partial w_{i}}\right)^{2}\right]$ represents Hessian matrix.
EWC changes the weights of individual models in the direction of the minimum change in the loss function so as to prevents catastrophic forgetting \cite{kirkpatrick2017overcoming}.

But there are difficulties in the applications of WA in subspace, such as low efficiency of random basis \cite{li2018measuring}, or expensive computation cost\cite{li2022low}, etc.
Moreover, when working with high-dimensional or large models, the projection matrix for projecting the gradient into the subspace can be too large for a single GPU to bear.
Wortsman et al. \cite{wortsman2021learning} provide a way to learn model subspace in a supervised learning.
Gaya et al. \cite{gaya2022learning} learn a convex subspace in online adaptation in reinforcement learning.
In short, how to explore the mechanism of vanilla average in subspace with numerous examples of training DNNs in subspace is a challenge for the future.




\subsection{Discussion}


WA gets the final model by averaging the weights of different deep models without additional computational complexity or training processes \cite{jin2022dataless,matena2022merging}.
In general, if random models have significant differences in presentation capabilities, structure, or training data, the results of fusion may not achieve the expected performance.
The linear interpolation of models from scratch using the same hyperparameter configuration but with different data orders is even less effective than stochastic models \cite{frankle2018lottery}.
Therefore, a large number of approaches described in this section aim to optimize the WA process in other mathematical ways.
Further, when models share part of their optimized trajectories (e.g., checkpoint averaging, tail averaginhg, SWA \cite{izmailov2018averaging,liu2018comparable}, etc.) or fine-tuned on the same pre-trained model (e.g., model soup \cite{wortsman2022model}, etc), the accuracy of interpolated models performs better \cite{neyshabur2020being}. 
Moreover, model soup \cite{wortsman2022model} averages the models with different hyperparameter configurations to get the final result.
In addition, selection of proper weights in model average can also be a challenge, which is often fraught with subjectivity.
More complex weight selection mechanisms may need plenty of complex trials and cross-validation.

WA is a promising technique in DL, which can be used as model optimization techniques in the future to reduce the weight fluctuation between different iterations, and improve the stability and convergence rate.
WA can improve the aggregation stage of FL to protect privacy better and reduce communication costs in the future.
Moreover, it is expected to reduce the storage space and computing overhead of the model on resource-constrained devices by implementing network compression on the terminal devices \cite{yao2021deep}.
In short, WA is a promising and cost-effective DL technique, which can be applied in areas such as FL to improve performance and reduce storage overhead.

\section{Ensemble Learning}

Ensemble learning, or multi-classifier system, is a technique that integrates multiple single models to generate final predictions, including voting, average \cite{sagi2018ensemble}, etc.
It improves overall performance and reduces the variance of the models, addressing issues such as overfitting, instability, and limited data volume.
In this section, we demonstrate ``Ensemble learning`` in DL and related techniques ``Model reuse``.

\subsection{Ensemble Learning}
Ensemble learning combines the outputs of networks, which surpasses than the result obtained from any model alone \cite{schapire1999brief,aniol2019ensemble,wang2016chinese}.
The general WA averages the model weights, that is, $f\left(x, \frac{1}{n} \sum_{i=1}^{n} W_{i}\right)$, which ends up with only one model.
In contrast, ensemble learning averages the output value after inference $\frac{1}{n} \sum_{i=1}^{n} f\left(x, W_{i}\right)$, resulting in multiple models \cite{wortsman2022model}.
Ensemble learning has a long history of research. There are plenty of typical algorithms, such as Adaboost \cite{freund1997decision}, Bagging \cite{breiman1996bagging}, Stacking \cite{wolpert1992stacked}, etc.

%
In order to make the network show better generalization ability, some previous work \cite{hansen1990neural,breiman2001random} applies the ensemble learning (e.g., random forest, etc.) to DNNs, which can be used to adjust the output and take full advantages in feature selection, noise filtering.
Kontschieder et al. \cite{kontschieder2015deep} propose deep neural decision forests, which uses the random decision function in the optimization algorithm of CNN to reduce the complexity of parameters.
Zhou et al. \cite{zhou2019deep} introduce a decision-tree ensemble approach to demonstrates the possibility of building models without backpropagation, which needs fewer hyperparameters than a typical deep neural network.
Moreover, Dropout \cite{srivastava2014dropout} typically needs to ensemble the output of all sub-nets to reduce prediction errors.
%
Nevertheless, if multiple models are too similar, the predictions of different networks will be too close to make sense of ensemble learning.
To find enough diverse models, snapshot ensemble \cite{huang2017snapshot} uses long learning rates, combining the predictions of multiple neural networks saved at the end of each learning rate cycle to produce one final result.
As an improvement on snapshot, FGE \cite{garipov2018loss} uses a linear piece-wise cyclic learning rate and smaller steps to find models along the low-loss path \cite{draxler2018essentially}, which inspires the relevant work of LMC.
Similarly, Laine et al. \cite{laine2016temporal} tend to ensemble the predictions over multiple previous training epochs.
Arpit et al. \cite{NEURIPS2022_372cb780} ensemble a set includes independent models and corresponding moving average models, which is referred to as ensemble of averages (EoA) as Eq.\ref{el_38}:
\begin{equation}
\label{el_38}
    \hat{y}=\arg \max _{n} \operatorname{Softmax}\left( \sum f\left(x; \hat{W}_{i}\right)\right)_{n}
\end{equation}
WAK et al. \cite{wang2022meta} present a distributed robust optimization (DRO) framework to learn from a black box model, fusing multiple models using a distributed robust optimization approach.
Hoang et al. \cite{hoang2019collective} demonstrate the ensemble of black-box experts with no access to black-box architectures.
Besides, there is a variety of work \cite{li2022branch,laine2016temporal} combines the ensemble learning with WA.
The ensemble learning in DL achieves remarkable results and is widely used in facial recognition \cite{wen2017ensemble}, speech recognition \cite{deng2014ensemble}, and other practical fields.

\subsection{Model Reuse}

\begin{table}
\centering
\caption{Summary of multiple model reuse methods based on model fusion.
}
\label{table7}
\vspace{0.5em}
\begin{tabular}{l l l} 
\toprule
  Methods & Ref. & Introduction   \\ 
\midrule


%
 FMR & \cite{yang2017deep}  &  \makecell[l]{reuse the fixed models to reduce the cost during training}
\\

PM$^{2}$R & \cite{xiang2017modal} & \makecell[l]{utilize consistency on different modalities}  \\

  MMR & \cite{lou2019towards} & \makecell[l]{reuse multiple source models  }\\

 NMMR & \cite{luo2022nonlinear} & \makecell[l]{take advantage of the nonlinear relationship}  \\

    RKME & \cite{wu2021model}   &  \makecell[l]{identify available pre-trained models by specifications in the deployment stage}\\

 HMR & \cite{wu2019heterogeneous}  &  \makecell[l]{combine and adjust the output of local models to generate a global model,}\\

   HMR for ML & \cite{tang2023improving}   & \makecell[l]{reuse of biased models trained on local datasets to construct a global model}\\

     RKHS & \cite{wu2021model}   &  \makecell[l]{
does not require calibration}\\    

     ZhiJian & \cite{zhang2023zhijian}   &  \makecell[l]{the merge module integrates the features,\\ weights, or predictions of the pre-trained models}\\

 \bottomrule
\end{tabular}
\end{table}

%
Based on existing pre-trained source models, model reuse \cite{zhou2016learnware} provides a required model applied to a new task without having to retrain the new model from scratch.
It can save time and computing resources and provide better performance in the case of limited resources \cite{yang2017deep}.
In addition, because the focus of transfer learning is to solve prediction tasks on the target domain, model reuse can be regarded as a kind of transfer learning.
But transfer learning requires labeled data for both source and target, while in model reuse, only unlabeled data can be collected and data from source domain can not be used \cite{luo2022nonlinear}.

Different from multi-classifiers ensemble learning, most current approaches reuse the existing features, labels or modalities to obtain the final prediction \cite{pan2009survey,zhou2016learnware} without storing a large amount of training data \cite{xiang2017modal}.
Fixed model reuse (FMR) \cite{yang2017deep} could be regarded as features reuse essentially.
Based on fixed models or features, FMR decreases the data required during training and provides privacy protection for fixed components.
But it can only use one type of source feature.
Jha et al. \cite{jha2018bag} present Bag of Experts (BoE) architecture
to reuse annotated data from reusable slots rather than one source domain train the target model training.
Pre-trained multi-model reuse (PM$^{2}$R) forms the predictions from pre-trained models into matrices and obtains the final predictions based on the consistency among different modalities.
But these type of methods ignore the potential information and only can be applied to limited scenarios.
Another crucial challenge of model reuse is to identify useful models from a set of pre-trained models for a given learning task.
Wu et al. \cite{wu2021model} propose reduced kernel mean embedding (RKME) specification to obtain available pre-trained models in the deployment stage.
Tang et al.\cite{tang2023improving} use optional calibration strategies and types of specifications, which combines the advantages of RKME and HMR.

Using a single model for model reuse produces too much homogenous information (e.g., a model trained in one domain may not fit data in another domain), and it is difficult to find a single pre-trained model that is perfectly suited to the target domain.
In general, we use a set of similar models to produce better performance than a single model, which is denoted as Multiple Model-Reuse (MMR) \cite{luo2022nonlinear}.
Based on MMR, Xiang et al. \cite{xiang2017modal} propose PM$^{2}$R without training data or validation instances.
Heterogeneous model reuse (HMR) \cite{wu2021model} tends to reuse the local models for global predictions at first and improve the local model by the multiparty multiclass margin (MPMC-margin).
Instead of using the output features or labels, Lou et al. \cite{lou2019towards} improve the way of representation, and use the hidden layer representation of the source model to train the target depth model, which is superior to the approach using the limited data in target domain.
nonlinear multi-model reuse (NMMR)
Nevertheless, some MMR methods will assume the linear relationship between the source model and the target model strictly, which is difficult to define in practice.
NMMR \cite{luo2022nonlinear} improves performance significantly by introducing a manifold regularization scheme to take advantage of arbitrary nonlinear relationships between the source and target models.
Specifically, we compare the characteristics of different reuse methods in Table \ref{table7}, 
Brifly, model reuse can significantly reduce the amount of data required by using pre-trained models to solve the problem of consuming a lot of bandwidth when transferring data between different ends.
Multi-model reuse also has a wide range of applications, such as speech recognition,  security and privacy interaction system, digital retina \cite{gao2018digital}, etc.

\subsection{Discussion}
Compared with related model fusion algorithms such as federated learning \cite{hsu2019measuring,hsu2020federated,mcmahan2017communication}, which have certain requirements on model parameters and sizes, ensemble methods use prediction to combine multiple heterogeneous weak classifiers without such limitations.
In addition, networks with different architectures in the ensemble approacesh will have a more obvious comparison effect than weight averge.
Ensemble methods, however, requires maintaining and running multiple trained models and running them together when tested.
Given the larger scale and complexity of deep learning models, this approach is not suitable for applications with limited computational resources and costs \cite{singh2020model}.
Due to the diversity of ensemble learning frameworks, it is possible to achieve model diversity and enhance generalization.
In the future, this will be important for dealing with changes in data and adversarial attacks.
Ensemble learning in DL is expected to provide confidence estimation and uncertainty measurement for model predictions, which is critical for safety and reliability in decision support systems, autonomous driving \cite{grigorescu2020survey}, medical diagnostics, etc.

\section{Application}

In recent years, a plenty of new research has appeared in the field of deep model fusion, which has also promoted the development of this related application field.
Based on the reviews of the development of model fusion and the current mainstream methods, this section summarizes some representative applications of the existing model fusion research ``Federated Learning``, ``Fine-tuning``, ``Distillation`` and ``Model Fusion on Foundation Models/LLMs``.
In the future, more work will try to further improve the accuracy and ease of model fusion, and gradually apply the model fusion method to real-world problems.

\subsection{Federated Learning}

\begin{figure}
	\centering 
	\includegraphics[width=1\textwidth, angle=0]{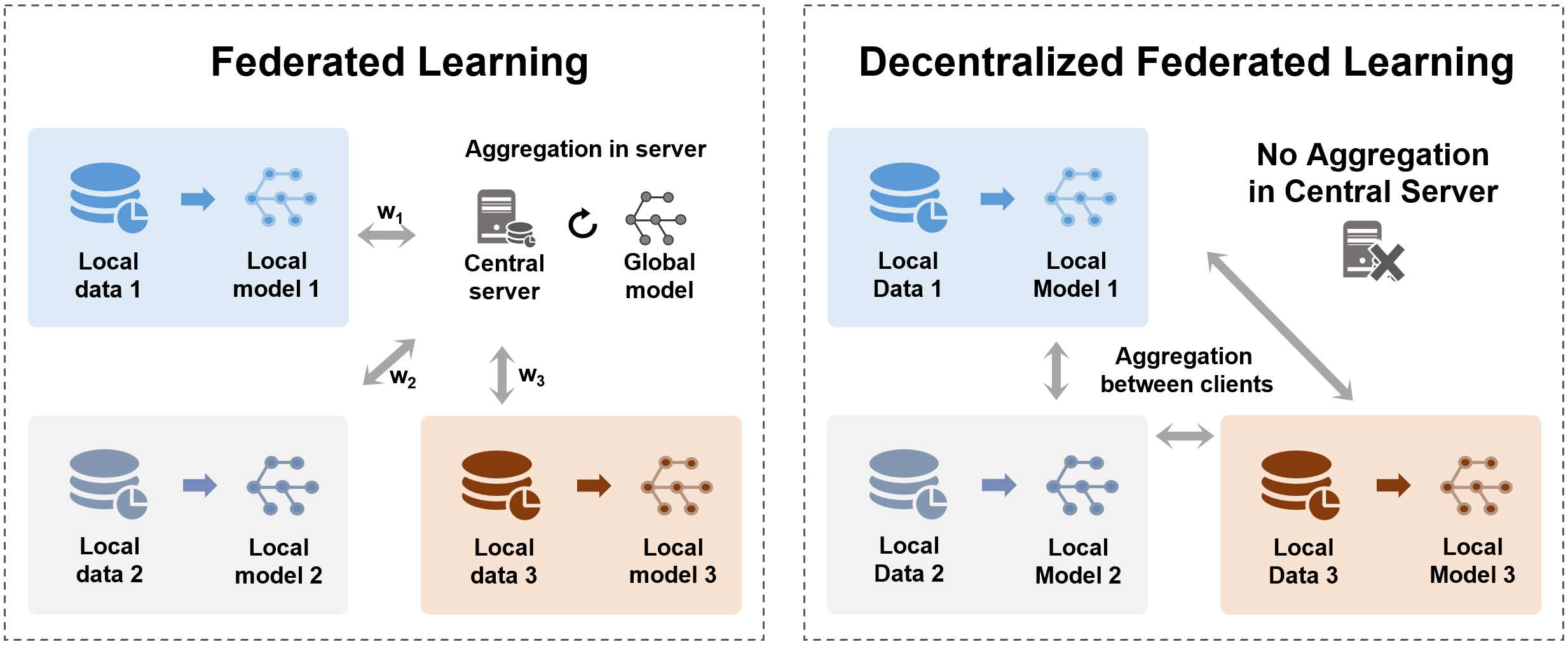}	
	\caption{\textbf{Two aggregation modes of federated learning.} \textbf{Left:} Centralized federated learning transfer models or gradients between the central server and the terminals of clients,  which are aggregated on the server finally.
 \textbf{Right:} Decentralized federated learning transfers and aggregates models between terminals of clients without a central server.
 } 
	\label{fig_6}%
\end{figure}

With the development of artificial intelligence, mobile devices, edge devices (e.g., IoT devices, sensors, etc.), and cloud computing platforms access to large amount of data.
However, due to the restrictions of practical scenarios and network bandwidth, it is is fraught with risk to collect all data from edge devices \cite{li2020federated,smith2017federated}.
To address the challenges of security and centralization of data storage, FL \cite{mcmahan2017communication,nishio2019client} allows many participants to collaborate to train shared global models while protecting data privacy, without the need to centralize datasets on a central server.
It also could be regarded as a multi-party learning problem \cite{pathak2010multiparty}.
Particularly, aggregation is a significant procedure of FL, which incorporates model or parameter updates trained by various parties (such as devices, organizations, or individuals).
In Figure \ref{fig_6}, we demonstrate two different aggregation approaches in centralized and decentralized FL.
Because of the efficient use of computing resources, low-cost nature (i.e., no need to transfer the entire datasets or maintain local parameters during training, etc.), Federated Averaging (FedAvg) \cite{mcmahan2017communication} is the most influential FL algorithms.
In the process of FedAvg, the local clients update the weights as Eq.\ref{FL_39}:
\begin{equation}
\label{FL_39}
   \mathbf{w}_{i}^{(t+1)}=\mathbf{w}_{i}^{(t)}-\eta \nabla g_{i}\left(\mathbf{w}_{i}^{(t)}, \xi_{i}^{(t)}\right) ,
\end{equation}
where $\nabla g_{i}\left(\mathbf{w}_{i}^{(t)}, \xi^{(t)}\right)$ represents stochastic gradient on the mini-batch $\xi_{i}^{(t)}$ at $t_{th}$ round \cite{mcmahan2017communication,kairouz2021advances}. The global model $\mathbf{w}^{(t)}$ is updated as Eq.\ref{FL_40}:
\begin{equation}
\label{FL_40}
    \mathbf{w}^{(t+1)}=\frac{1}{n} \sum_{i=1}^{n} \mathbf{w}_{i}^{(t)}.
\end{equation}

Due to the heterogeneity of models (e.g., data distribution, bandwidth environment, network structure, permutation invariance \cite{entezari2021role}, etc.), a simple aggregation of weights can adversely affect the performance of the final model and put the pressure on communication \cite{mohri2019agnostic}.
We list the common aggregation methods in Table \ref{table8}.
Probabilistic federated neural matching (PFNM) \cite{yurochkin2019bayesian} uses the Bayesian nonparametric mechanism to adjust the global model size to accommodate the heterogeneity of data.
But it can only be applied to simple architectures.
FedMA \cite{wang2020federated} proposes to hierarchically match neurons of a network, which is quite difficult in practice (participant models need to have the same number of layers and structure).
FedBABU \cite{oh2021fedbabu} only aggregates the body in the aggregation phase instead of the whole network, where body is related the generality of the network and head represents personalization.
It is more adaptable to adapt to the heterogeneous data of each client, and improves the representation and personalization ability of a single global model.

Moreover, centralized gradient aggregation puts pressure on communication bandwidth and computing costs.
In order to avoid the risk of failure of large-scale centralized fusion of local models, Hoang et al. \cite{hoang2019collective} compare centralized and distributed gradient aggregation that occurs only in the local experts.
Other recent work \cite{reddi2021adaptive,hsu2019measuring} regards client updates as pseudo-gradients $\Delta_{i}$, which is aggregated as Eq.\eqref{fl_42}, and the global model is updated as Eq.\eqref{fl_43}:
\begin{equation}
\label{fl_42}
     \bar{\Delta}^{(t)}=\frac{1}{n} \sum_{i=1}^{n} \Delta_{i}^{(t)}
\end{equation}
\begin{equation}
\label{fl_43}
    \mathbf{w}^{(t+1)}=\mathbf{w}^{(t)}-\eta \bar{\Delta}^{(t)}
\end{equation}
Based on it, Jhunjhunwala et al. \cite{jhunjhunwala2023fedexp} propose FedExp, a dynamically varying pseudo-gradient self-adaptive method for caculating the server step size.
FedExp accelerates convergence and reduces the overhead, which uses the extrapolation to accelerate Projection Onto Convex Sets (POCS) as Eq.\eqref{fl_44}:
\begin{equation}
\label{fl_44}
\mathbf{w}_{\mathrm{POCS}}^{(t+1)}=\mathbf{w}_{\mathrm{POCS}}^{(t)}-\lambda\left(\frac{1}{n} \sum_{i=1}^{n} P_{i}\left(\mathbf{w}_{\mathrm{POCS}}^{(t)}\right)-\mathbf{w}_{\mathrm{POCS}}^{(t)}\right)
\end{equation}
Huang et al. \cite{huang2022achieving} aggregate personalized sparse gradients and masks trained from local models to generate new global model as Eq.\eqref{fl_45}:
\begin{equation}
\label{fl_45}
    \mathbf{w}^{(t+1)}=\mathbf{w}^{(t)}-\frac{1}{\left|S_{t}\right|} \sum \left(\tilde{\mathbf{w}}_{ 0}^{(t)}-\tilde{\mathbf{w}}_{n}^{(t)}\right),
\end{equation}
where $S_{t}$ denotes the clients.
It reduces the communication overhead and solves the issues of sparse personalized FL.
In addition, the application of personalized model to FL could adapt the preferences of local users and decrease the costs \cite{fallah2020personalized,dinh2017sharp,lam2021model}.

\begin{table}[!h] \small 
\centering
 \caption{The different aggregation approaches in Federated Learning} 
 \label{table8}
\vspace{0.5em}
\begin{tabular}{l l l l}

\toprule
Model fusion in FL & Methods & Ref. & Aggregation \\

\midrule
\multirow{1}{*}{Aggregation}   
&FedAvg& \cite{mcmahan2017communication}& \makecell[l]{aggregate the parameters of the participants directly }\\
  &FedExp&\cite{jhunjhunwala2023fedexp}&\makecell[l]{determine the server step size based on  pseudo-gradients}\\

&FedMA&\cite{wang2020federated}& \makecell[l]{hierarchically match neurons of a network}\\
&PFNM&\cite{yurochkin2019bayesian}& \makecell[l]{match the neurons of the networks} \\
&FedBABU&\cite{oh2021fedbabu}& \makecell[l]{updates only the body of the models during training} \\
&FedSPA&\cite{huang2022achieving}& \makecell[l]{aggregate the sparse gradients and masks from local clients} \\

 \midrule
\multirow{1}{*}{Ensemble}   
& FedCVAE-ENS &  \cite{heinbaugh2022data} &
\makecell[l]{leverage CVAE to address statistical heterogeneity}\\\

&one-shot &\cite{guha2019one} &\makecell[l]{ensemble the predictions of clients in a single iteration}\\
    &DENSE &\cite{zhang2022dense}&\makecell[l]{ensemble local models for the global model}  \\

\midrule
    
 \multirow{1}{*}{\makecell[c]{Distillation}} 
 &  FedDF & \cite{Lintao2020FedDF} &  \makecell[l]{address the  quality loss \cite{hsieh2020non} of BN \\and heterogeneous client models  }\\

  &FedFTG & \cite{zhang2022fine}&\makecell[l]{
use a data-free KD method to fine-tune the global model}	\\
  & FedCVAE-KD&     \cite{heinbaugh2022data}&\makecell[l]{compress the ensemble of client decoders into a decoder}\\
  &FedBE &\cite{chen2020fedbe} &\makecell[l]{use Bayesian methods and ensemble distillation}\\
  &FedAUX&\cite{sattler2021fedaux} & \makecell[l]{weight the logits of local models by certainty score}\\

\bottomrule
 \end{tabular}

 \end{table}
%

Since ensemble learning does not require averaging weights, it could be a good tool for aggregation and support heterogeneous client models. 
One-shot \cite{guha2019one} utilizes ensemble learning to aggregate the local model, which achieves a relative gain of 51.5 $\%$ over the baseline on the AUC.
Similarly, there are plenty of researches that applies the ensemble learning to FL \cite{heinbaugh2022data,zhang2022dense}.
Under certain conditions ( $i_{m}< \sqrt{i_{s}} $ where $i_{m}$ denotes machines, $i_{s}$ is samples), the performance of the direct weight aggregation can be comparable to centralized algorithm that can access all samples in data distributed communication  \cite{zhang2012communication}.
Nevertheless, it is not available to apply ensemble learning techniques directly in FL due to the heavy burden of keeping all the received models on the server.
KD could solve these problems and regularize the size of global model and local learning using multi-teachers ensemble methods \cite{zhu2021data}.
Recent work \cite{Lintao2020FedDF,gong2021ensemble,jeong2018communication} present some novel FL framework based on ensemble distillation.
FedFTG \cite{zhang2022fine} does not directly broadcast the aggregate model back to each client, but uses knowledge extracted from the local model to fine-tune this preliminary global model in the server, mitigating the performance degradation after the model is aggregated.
FedDF breaks the communication barrier between heterogeneous client models \cite{hsieh2020non}.
FedCVAE-KD \cite{heinbaugh2022data} uses a lightweight knowledge distillation process to aggregate the client decoders, which generates substantially samples than FedAvg.
It address the statistical heterogeneity and pipeline security \cite{zhou2020distilled} (i.e., outside attacker who obtains transferred data cannot train a classifier) concerns.

In short, the essence of the aggregation step in FL is a model fusion technique. Selecting a reasonable model fusion method can reduce the impact of specific participants or individual data on the final model, so as to improve the generalization ability and adaptability of the model in the global scope.
In future work, a good aggregation approach is expected to be helpful in facing a series of challenges in federated learning.
In future work, a high-quality and scalable aggregation approache are expected to face a series of challenges in FL, such as client heterogeneity, non-i.i.d heterogeneous data, limited computing resources \cite{Lintao2020FedDF}, etc.
FL is expected to show its potential in many more areas, such as NLP, recommendation systems \cite{liu2021fedct}, medical image analysis \cite{liu2021feddg}, etc.

\subsection{Fine-tuning}
\begin{figure}
	\centering 
	\includegraphics[width=1\textwidth, angle=0]{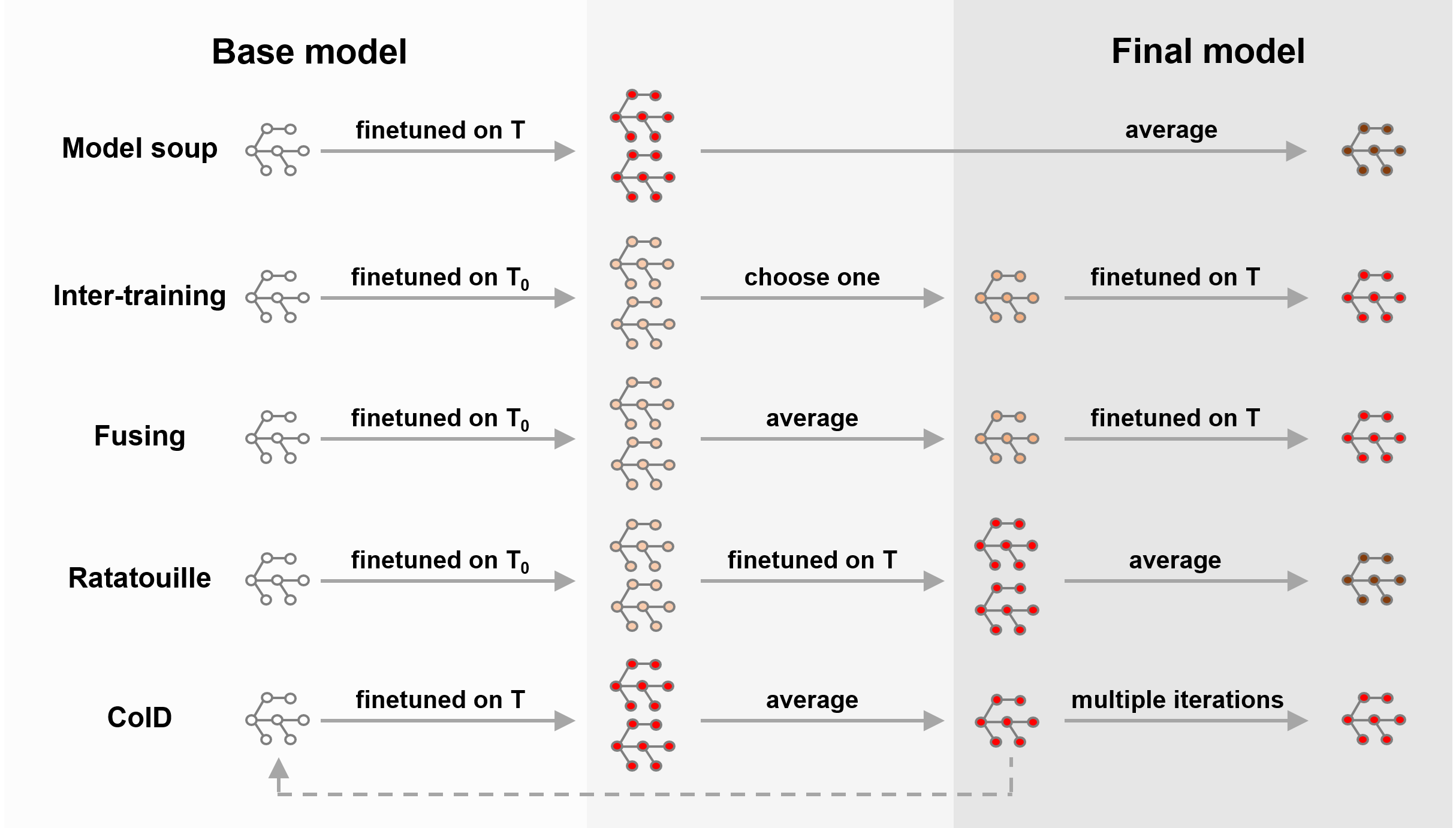}	
	\caption{\textbf{Different methods of applying Weight Average in fine-tuning scenarios.}
 (General Fine-tuning  \cite{Oquab_2014_CVPR}, WiSE\cite{wortsman2022robust}, Inter-training\cite{phang2018sentence}: selects the appropriate fine-tuning model on intermediate tasks as the base model. Fusing\cite{choshen2022fusing}:average models fine-tuned on intermediate(Source) tasks. 
Model Soup\cite{wortsman2022model}: average models that are fine-tuned on the target task.
 Source task $T_{s}$.
 Ratatouille\cite{rame2023model-ratatouille} : recycle the multiple fine-tunings on diverse auxiliary tasks, then averages all the fine-tuned weights to get the final model.
target task $T$, Auxiliary task $T_{aux}$.} 
	\label{fig_7}%
\end{figure}

Fine-tuning a base mode, such as pre-trained model, is an efficient approach for adjusting models to perform downstream tasks \cite{chen2022revisiting,devlin2018bert}, which results in better generalization and more accurate output with less labeled data.
Compared with random initialization, a pre-trained model is trained by a relatively set of task-specific data, which is always a better standard starting point for training.
Nevertheless. the average of existing fine-tuned models \cite{choshen2022fusing,choshen2022start} is even a better base model than the vanilla pre-trained model for fine-tuning on the downstream tasks.
Besides, there is a great deal of recent work combining WA with fine-tuning as shown in Figure \ref{fig_7}, such as model soup \cite{wortsman2022model}, DiWA \cite{rame2022diverse}, etc. 
Fine-tuning improves the accuracy on target distribution, but often leads to a decrease in the robustness of distribution shift.
WiSE-FT \cite{wortsman2022robust} combines the weights of the zero-shot and fine-tuned models to improve the distribution shift accuracy while retaining the high accuracy of the target distribution.
Local fine-tuning (Lo-fi) \cite{wortsman2022fi} fine-tunes each node independently without any communication, and then averages the nodes.
Lo-fi can also improve the performance of distributed shifts.
Collaborative Descent fusion (ColD) \cite{donyehiya2022cold} replaces base models with fusion models that can be recycled, which can continually improve the pre-trained models on which they are based.
ColD \cite{donyehiya2022cold} is superior to RoBERTa \cite{liu2019roberta} and even previous multitasking models.
While these strategies for averaging the fine-tuned models may be simple, they do not take full advantage of the connections between each fine-tuned model.
Therefore, training on an intermediate task before 
before training on a target task can explore the capabilities of the base models \cite{phang2020english,vu2020exploring,pruksachatkun2020intermediatetask}.
Inspired by inter-training strategies \cite{pruksachatkun2020intermediatetask}, Rame et al. \cite{rame2023model-ratatouille} fine-tune the models on auxiliary tasks, which utilize diverse auxiliary tasks and improve the out-of-distribution (OOD) generalization.



The average of fine-tuned models reduces the training time required to achieve the goal \cite{choshen2022start} and generates more accurate and better generalized models.
Essentially, different ways of fine-tuning (e.g., fine-tuning with frozen layers, top-layer fine-tuning, etc.) also have a certain impact on final accuracy and distribution shift \cite{wortsman2022robust}.
However, the combination of WA and fine-tuning is an expensive overhead, which has a certain limitation on specific application.
Also, it may face a problem of explosion of preservation checkpoints, or catastrophic forgetting \cite{kirkpatrick2017overcoming}, especially applied to transfer learning.

\subsection{Distillation}
%
\begin{figure}
	\centering 
	\includegraphics[width=1\textwidth, angle=0]{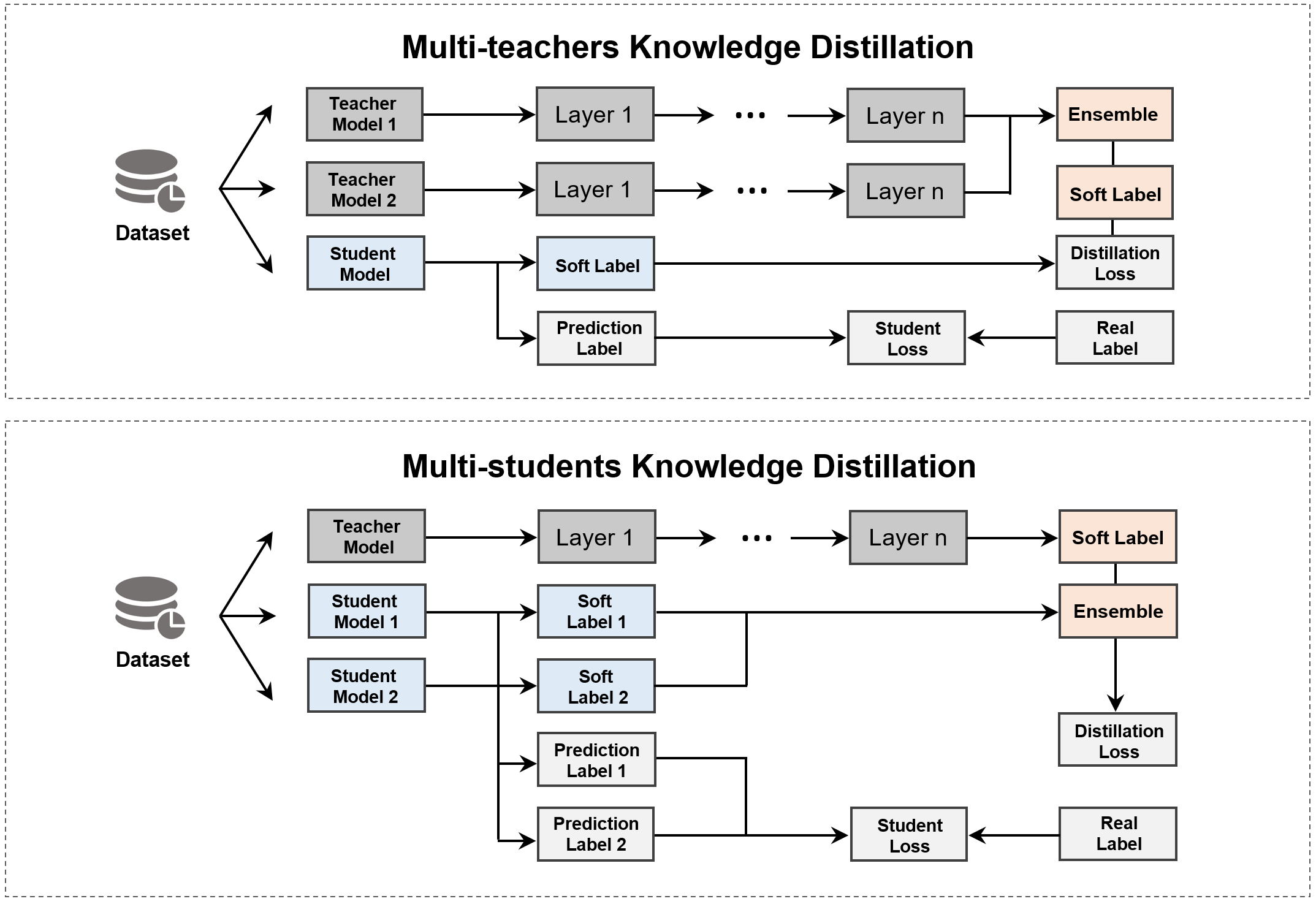}	
	\caption{\textbf{Two aggregation modes of  distillation.} \textbf{Top:} in contrast to standard KD, this framework incorporates multiple teacher models for distillation.
 \textbf{Bottom:} the framework incorporates multiple student models for distillation.
 } 
	\label{fig_8}%
\end{figure}

Knowledge distillation (KD) \cite{hinton2015distilling} is a significant method to ensemble multiple models, which involves the following two types of models.
A teacher model denotes large and powerful model trained on large-scale data and has high predictive and expressive power.
A student models is a relatively smaller model with fewer parameters and computational resource \cite{schmidhuber1992learning,buciluǎ2006model}.
Using the knowledge of the teacher (e.g., the output probability distribution, hidden layer representation, etc.) to guide the training, the student could achieve the prediction ability closed to the large model with fewer resources and faster speed \cite{ahn2019variational,koratana2019lit,kim2018paraphrasing,tung2019similarity}.
Given that multiple teachers or students are expected to have a preferable performance than a single model \cite{anil2018large}, we divide KD into two categories according to the aggregated objects as Figure \ref{fig_8}.

\textbf{The first type of approach is to merge multiple teacher models and distill the student model directly,} as shown in Table \ref{table9}. Currently, recent work mainly integrates the output of teachers (e.g., logits \cite{Dvornik_2019_ICCV,anil2018large,you2017learning} or feature-based knowledge \cite{Wu_2019_CVPR,liu2019knowledge}, etc.).
Ensemble distillation (ED) \cite{Lintao2020FedDF,malinin2019ensemble} distills the average output of multiple teachers to a student model, which can make up for the shortcomings of a single teacher model and provide more diversified and comprehensive information to the student model.
FedDF \cite{Lintao2020FedDF} distills a collection of client-teacher models $\left | S_{t} \right | $ into a server-student model.
It averages the logit output $f\left(\hat{\mathbf{x}}_{t}^{k}\right)$ of teachers as Eq.\eqref{kd_46}:

\begin{equation}
\label{kd_46}
    \mathbf{x}_{t, j}:=\mathbf{x}_{t, j-1}-\eta \frac{\partial \mathrm{KL}\left(\sigma\left(\frac{1}{\left|\mathcal{S}_{t}\right|} \sum_{k \in \mathcal{S}_{t}} f\left(\hat{\mathbf{x}}_{t}^{k}\right)\right), \sigma\left(f\left(\mathbf{x}_{t, j-1}\right)\right)\right)}{\partial \mathbf{x}_{t, j-1}},
\end{equation}
where $t$ is the communication round, KL means KL divergence. The ensemble part of FedDF does not affect the overall workflows of clients and solves the loss problem of network batch normalization (BN) \cite{hsieh2020non}, 
Wu et al. \cite{Wu_2019_CVPR} propose a multi-teacher adaptive distillation framework that can transfer knowledge from multiple teachers to student without the need for source domain data.
Although merging multiple teachers makes up for the shortcomings of a single teacher, some of the teacher's information may be overlooked when conflict exist among teachers.

\textbf{The other way is to use the teacher model to distill multiple students and then merge these student models.} Co-distillation (CD) \cite{anil2018large} regards each client device as a student model, treats the average of the logits output of the other devices as teacher`s output .
However, the same training data sample should be used to synchronize the output of the teacher and the local student model.
In order to solve the problem of CD, FD \cite{jeong2018communication} uploads these local average logit vectors to the server periodically.
Each average logits with the associated label as the current training sample will be used as the distillation regularizer for the next round of local device computation.
FD improves performance and reduces communication rounds significantly.
However, merging multi-students also has some problems, such as large demand of computing resources, poor interpretation and over-dependence on the original model.

\begin{table}[!h] \small 
\centering
 \caption{Classification of KD according to the differences of the aggregated objects}  \label{table9}
\vspace{0.5em}
\begin{tabular}{l l l l}

\toprule
\makecell[l]{Merge mode\\of distillation} & Methods & Ref. & Introduction \\

\midrule
    \multirow{1}{*}{\makecell[l]{Merge multiple\\ teachers}}
  & FedDF & \cite{Lintao2020FedDF} &  \makecell[l]{addresses the quality loss issue\cite{hsieh2020non} of BN, break the  knowledge\\ barriers among heterogeneous client models}\\

& END$^{2}$ & \cite{malinin2019ensemble} & \makecell[l]{
distilling the distribution of the predictions from an ensemble instead\\of the average prediction}	\\
& AE-KD & \cite{du2020agree} & \makecell[l]{regard the ensemble knowledge distillation\\ as a multi-objective optimization problem}	\\
 & FedFTG &   \cite{zhang2022fine} &\makecell[l]{
a data-free knowledge distillation method, \\which relieves the issue of direct model aggregation} \\

     \hline
    \multirow{1}{*}{\makecell[l]{Merge multiple\\ students}} 
 & Batch Ensemble & \cite{wen2020batchensemble}& \makecell[l]{mini-batch friendly, parallelizable within\\a device, minor memory overhead }\\
 
	
& Hydra  & \cite{tran2020hydra} &\makecell[l]{improve distillation performance while capturing\\ the uncertainty behavior of the original ensemble }	\\ 
  & LatentBE  & \cite{nam2022improving} &\makecell[l]{
average a student with multiple subnetworks, giving a single student\\network with no additional inference cost}	\\

\bottomrule
 \end{tabular}

 \end{table}

\subsection{Model Fusion on Foundation Models/LLMs}

Foundation models show strong performance and emergent abilities when dealing with complex tasks, 
Large foundation models are characterized by their sheer size, containing billions of parameters that help them learn complex patterns in the data.
Especially, with the emergence of new LLMs \cite{zhao2023survey,shanahan2022talking} recently, such as GPT-3 \cite{brown2020language,openai2023chatgpt4}, T5 \cite{raffel2020exploring}, BERT \cite{devlin2018bert}, Megatron-LM, the application of WA \cite{zhang2018graph,sun2023multitask,lv2023parameter} to LLMs attracts more attention.
You et al. \cite{you2022ranking} propose B-tuning using Bayesian learning to calculate posterior prediction distribution, which tunes top-K ranked pre-trained models by their transferabilities.
Zoo-tuning \cite{shu2021zoo} aggregates the weights of pre-trained model with aligned channels to obtain the final model adapt to downstream tasks, which improve the issue of high cost of migrating on large models.

Besides, recent work \cite{zhang2018graph,kingetsu2021neural} tends to craft better framework and modules adapted to the application LLMs.
Izacard et al. \cite{izacard2020leveraging} present fusion-in-decoder (FiD) , a novel framework to perform evidence fusion in the decoder only, which aims to efficiently aggregate multiple passages.
Based on FiD, Ravaut et al. \cite{ravaut2022towards} introduce Summa Fusion to concatenate the representations of summary candidates, which further explores the effectiveness of fusion in the context of text summarization.
However, their results improve little because they do not filter out poor quality candidates before using the algorithm.
In contrast, Jiang et al. \cite{jiang2023llm} propose an ensemble framework LLM-BLENDER, which focus on identifying subtle differences in the output of different candidates by PairRanker algorithm, and then ranking and summarizing the candidates to achieve better performance.
Huang et al. \cite{huang2023lorahub} introduce Low-rank adaptations hub (LoRAHub), a framework to combine multiple LoRA modules trained on different tasks, which is designed to increase the adaptability of the LLMs and reduce training costs.


due to the high performance and low computational resources, the application of fine-tuning to large foundation models improve obustness of distribution shifts \cite{wortsman2022robust} .
Branch-Train-Merge (BTM) \cite{li2022branch} reduce the large amount of multi-node synchronization required for parallel LLMs training.
In addition, the negative task vector of task arithmetic \cite{ortiz2023task} can reduce the number of toxic generations of LLMs.
For example,  it decreases the amount from 4.8 $\%$ to 0.8 $\%$ in GPT-2 \cite{ilharco2022editing}.

\section{Conclusion}

In this survey, we review the deep model fusion techniques which aims at improving the performance of the model.
%
We propose a new categorization that groups the tecnologies of deep model fusion into four perspective: ``mode connectivity", ``alignment," ``weight average" and ``ensemble learning".
In the first three chapters, we describe the fusion of model's weight to obtain the superior final fused model.
In the ``ensemble learning", we focus on the fusion of the output of deep models with a wealth of available methods and a large number of ensemble frameworks.
We summarize the common methods from the point of view of algorithm design and performance, and compare the differences, advantages and disadvantages of different approaches.
Finally, we discusses the applications and engineering prospects of deep model fusion technology in FL, distillation, LLMs, etc.
%

%
%

We not only summarize current technologies of deep model fusion, but also point out the bottlenecks and breakthrough.
The survey is expected to help the developers improve the performance of deep model fusion technologies, and indicate the promising and valuable directions.
In the future, it is worth designing novel deep model fusion strategies from innovative aggregation patterns, better initial conditions, diverse ensemble frameworks and other perspectives.
The abundant information in the loss landscape and the potential relationships between the components of networks remain to be further exploited.
In addition, better adaptive methods are expected to be applied in heterogeneous models and complex real scenarios, such as FL, large-scale models, transfer learning, etc.
Also, we need to pay attention to the practical effects to promote the development and application of deep model fusion technologies.




\bibliographystyle{plain}
\bibliography{reference.bib}

\begin{thebibliography}{100}

\bibitem{afyouni2022multi}
Imad Afyouni, Zaher Al~Aghbari, and Reshma~Abdul Razack.
\newblock Multi-feature, multi-modal, and multi-source social event detection:
  A comprehensive survey.
\newblock {\em Information Fusion}, 79:279--308, 2022.

\bibitem{ahn2019variational}
Sungsoo Ahn, Shell~Xu Hu, Andreas Damianou, Neil~D Lawrence, and Zhenwen Dai.
\newblock Variational information distillation for knowledge transfer.
\newblock In {\em Proceedings of the IEEE/CVF conference on computer vision and
  pattern recognition}, pages 9163--9171, 2019.

\bibitem{ainsworth2022git}
Samuel~K Ainsworth, Jonathan Hayase, and Siddhartha Srinivasa.
\newblock Git re-basin: Merging models modulo permutation symmetries.
\newblock {\em arXiv preprint arXiv:2209.04836}, 2022.

\bibitem{akash2022wasserstein}
Aditya~Kumar Akash, Sixu Li, and Nicol{\'a}s~Garc{\'\i}a Trillos.
\newblock Wasserstein barycenter-based model fusion and linear mode
  connectivity of neural networks.
\newblock {\em arXiv preprint arXiv:2210.06671}, 2022.

\bibitem{akhlaghi2018knowledge}
Milad~I Akhlaghi and Sergey~V Sukhov.
\newblock Knowledge fusion in feedforward artificial neural networks.
\newblock {\em Neural Processing Letters}, 48(1):257--272, 2018.

\bibitem{anil2018large}
Rohan Anil, Gabriel Pereyra, Alexandre Passos, Robert Ormandi, George~E Dahl,
  and Geoffrey~E Hinton.
\newblock Large scale distributed neural network training through online
  distillation.
\newblock {\em arXiv preprint arXiv:1804.03235}, 2018.

\bibitem{aniol2019ensemble}
Anna Aniol, Marcin Pietron, and Jerzy Duda.
\newblock Ensemble approach for natural language question answering problem.
\newblock In {\em 2019 Seventh International Symposium on Computing and
  Networking Workshops (CANDARW)}, pages 180--183. IEEE, 2019.

\bibitem{arora2018stronger}
Sanjeev Arora, Rong Ge, Behnam Neyshabur, and Yi~Zhang.
\newblock Stronger generalization bounds for deep nets via a compression
  approach.
\newblock In {\em International Conference on Machine Learning}, pages
  254--263. PMLR, 2018.

\bibitem{NEURIPS2022_372cb780}
Devansh Arpit, Huan Wang, Yingbo Zhou, and Caiming Xiong.
\newblock Ensemble of averages: Improving model selection and boosting
  performance in domain generalization.
\newblock In S.~Koyejo, S.~Mohamed, A.~Agarwal, D.~Belgrave, K.~Cho, and A.~Oh,
  editors, {\em Advances in Neural Information Processing Systems}, volume~35,
  2022.

\bibitem{auer1995exponentially}
Peter Auer, Mark Herbster, and Manfred~KK Warmuth.
\newblock Exponentially many local minima for single neurons.
\newblock {\em Advances in neural information processing systems}, 8, 1995.

\bibitem{benton2021loss}
Gregory Benton, Wesley Maddox, Sanae Lotfi, and Andrew Gordon~Gordon Wilson.
\newblock Loss surface simplexes for mode connecting volumes and fast
  ensembling.
\newblock In {\em International Conference on Machine Learning}, pages
  769--779. PMLR, 2021.

\bibitem{benzing2022random}
Frederik Benzing, Simon Schug, Robert Meier, Johannes Von~Oswald, Yassir Akram,
  Nicolas Zucchet, Laurence Aitchison, and Angelika Steger.
\newblock Random initialisations performing above chance and how to find them.
\newblock {\em arXiv preprint arXiv:2209.07509}, 2022.

\bibitem{blundell2015weight}
Charles Blundell, Julien Cornebise, Koray Kavukcuoglu, and Daan Wierstra.
\newblock Weight uncertainty in neural network.
\newblock In {\em International conference on machine learning}, pages
  1613--1622. PMLR, 2015.

\bibitem{brea2019weight}
Johanni Brea, Berfin Simsek, Bernd Illing, and Wulfram Gerstner.
\newblock Weight-space symmetry in deep networks gives rise to permutation
  saddles, connected by equal-loss valleys across the loss landscape.
\newblock {\em arXiv preprint arXiv:1907.02911}, 2019.

\bibitem{breiman1996bagging}
Leo Breiman.
\newblock Bagging predictors.
\newblock {\em Machine learning}, 24:123--140, 1996.

\bibitem{breiman2001random}
Leo Breiman.
\newblock Random forests.
\newblock {\em Machine learning}, 45:5--32, 2001.

\bibitem{brown2020language}
Tom Brown, Benjamin Mann, Nick Ryder, Melanie Subbiah, Jared~D Kaplan, Prafulla
  Dhariwal, Arvind Neelakantan, Pranav Shyam, Girish Sastry, Amanda Askell,
  et~al.
\newblock Language models are few-shot learners.
\newblock {\em Advances in neural information processing systems},
  33:1877--1901, 2020.

\bibitem{buciluǎ2006model}
Cristian Buciluǎ, Rich Caruana, and Alexandru Niculescu-Mizil.
\newblock Model compression.
\newblock In {\em Proceedings of the 12th ACM SIGKDD international conference
  on Knowledge discovery and data mining}, pages 535--541, 2006.

\bibitem{caron2021emerging}
Mathilde Caron, Hugo Touvron, Ishan Misra, Herv{\'e} J{\'e}gou, Julien Mairal,
  Piotr Bojanowski, and Armand Joulin.
\newblock Emerging properties in self-supervised vision transformers.
\newblock In {\em Proceedings of the IEEE/CVF international conference on
  computer vision}, pages 9650--9660, 2021.

\bibitem{casado2020ensemble}
{\'A}ngela Casado-Garc{\'\i}a and J{\'o}nathan Heras.
\newblock Ensemble methods for object detection.
\newblock In {\em ECAI 2020}, pages 2688--2695. IOS Press, 2020.

\bibitem{cha2021swad}
Junbum Cha, Sanghyuk Chun, Kyungjae Lee, Han-Cheol Cho, Seunghyun Park, Yunsung
  Lee, and Sungrae Park.
\newblock Swad: Domain generalization by seeking flat minima.
\newblock {\em Advances in Neural Information Processing Systems},
  34:22405--22418, 2021.

\bibitem{chen1993geometry}
An~Mei Chen, Haw-minn Lu, and Robert Hecht-Nielsen.
\newblock On the geometry of feedforward neural network error surfaces.
\newblock {\em Neural computation}, 5(6):910--927, 1993.

\bibitem{chen2022revisiting}
Guanzheng Chen, Fangyu Liu, Zaiqiao Meng, and Shangsong Liang.
\newblock Revisiting parameter-efficient tuning: Are we really there yet?
\newblock {\em arXiv preprint arXiv:2202.07962}, 2022.

\bibitem{chen2020fedbe}
Hong-You Chen and Wei-Lun Chao.
\newblock Fedbe: Making bayesian model ensemble applicable to federated
  learning.
\newblock {\em arXiv preprint arXiv:2009.01974}, 2020.

\bibitem{chen2017checkpoint}
Hugh Chen, Scott Lundberg, and Su-In Lee.
\newblock Checkpoint ensembles: Ensemble methods from a single training
  process.
\newblock {\em arXiv preprint arXiv:1710.03282}, 2017.

\bibitem{chen2023fedsoup}
Minghui Chen, Meirui Jiang, Qi~Dou, Zehua Wang, and Xiaoxiao Li.
\newblock Fedsoup: Improving generalization and personalization in federated
  learning via selective model interpolation, 2023.

\bibitem{choromanska2015loss}
Anna Choromanska, Mikael Henaff, Michael Mathieu, G{\'e}rard~Ben Arous, and
  Yann LeCun.
\newblock The loss surfaces of multilayer networks.
\newblock In {\em Artificial intelligence and statistics}, pages 192--204.
  PMLR, 2015.

\bibitem{choshen2022start}
Leshem Choshen, Elad Venezian, Shachar Don-Yehia, Noam Slonim, and Yoav Katz.
\newblock Where to start? analyzing the potential value of intermediate models.
\newblock {\em arXiv preprint arXiv:2211.00107}, 2022.

\bibitem{choshen2022fusing}
Leshem Choshen, Elad Venezian, Noam Slonim, and Yoav Katz.
\newblock Fusing finetuned models for better pretraining.
\newblock {\em arXiv preprint arXiv:2204.03044}, 2022.

\bibitem{chowdhary2020natural}
KR1442 Chowdhary and KR~Chowdhary.
\newblock Natural language processing.
\newblock {\em Fundamentals of artificial intelligence}, pages 603--649, 2020.

\bibitem{chowdhery2022palm}
Aakanksha Chowdhery, Sharan Narang, Jacob Devlin, Maarten Bosma, Gaurav Mishra,
  Adam Roberts, Paul Barham, Hyung~Won Chung, Charles Sutton, Sebastian
  Gehrmann, et~al.
\newblock Palm: Scaling language modeling with pathways.
\newblock {\em arXiv preprint arXiv:2204.02311}, 2022.

\bibitem{chronopoulou2023adaptersoup}
Alexandra Chronopoulou, Matthew~E Peters, Alexander Fraser, and Jesse Dodge.
\newblock Adaptersoup: Weight averaging to improve generalization of pretrained
  language models.
\newblock {\em arXiv preprint arXiv:2302.07027}, 2023.

\bibitem{cooper2021global}
Yaim Cooper.
\newblock Global minima of overparameterized neural networks.
\newblock {\em SIAM Journal on Mathematics of Data Science}, 3(2):676--691,
  2021.

\bibitem{Croce_2023_CVPR}
Francesco Croce, Sylvestre-Alvise Rebuffi, Evan Shelhamer, and Sven Gowal.
\newblock Seasoning model soups for robustness to adversarial and natural
  distribution shifts.
\newblock In {\em Proceedings of the IEEE/CVF Conference on Computer Vision and
  Pattern Recognition (CVPR)}, pages 12313--12323, June 2023.

\bibitem{crooks2007measuring}
Gavin~E Crooks.
\newblock Measuring thermodynamic length.
\newblock {\em Physical Review Letters}, 99(10):100602, 2007.

\bibitem{czarnecki2019deep}
Wojciech~Marian Czarnecki, Simon Osindero, Razvan Pascanu, and Max Jaderberg.
\newblock A deep neural network's loss surface contains every low-dimensional
  pattern.
\newblock {\em arXiv preprint arXiv:1912.07559}, 2019.

\bibitem{daheim2023elastic}
Nico Daheim, Nouha Dziri, Mrinmaya Sachan, Iryna Gurevych, and Edoardo~M Ponti.
\newblock Elastic weight removal for faithful and abstractive dialogue
  generation.
\newblock {\em arXiv preprint arXiv:2303.17574}, 2023.

\bibitem{dai2021coatnet}
Zihang Dai, Hanxiao Liu, Quoc~V Le, and Mingxing Tan.
\newblock Coatnet: Marrying convolution and attention for all data sizes.
\newblock {\em Advances in neural information processing systems},
  34:3965--3977, 2021.

\bibitem{dauphin2014identifying}
Yann~N Dauphin, Razvan Pascanu, Caglar Gulcehre, Kyunghyun Cho, Surya Ganguli,
  and Yoshua Bengio.
\newblock Identifying and attacking the saddle point problem in
  high-dimensional non-convex optimization.
\newblock {\em Advances in neural information processing systems}, 27, 2014.

\bibitem{deng2014ensemble}
Li~Deng and John Platt.
\newblock Ensemble deep learning for speech recognition.
\newblock In {\em Proc. interspeech}, 2014.

\bibitem{devlin2018bert}
Jacob Devlin, Ming-Wei Chang, Kenton Lee, and Kristina Toutanova.
\newblock Bert: Pre-training of deep bidirectional transformers for language
  understanding.
\newblock {\em arXiv preprint arXiv:1810.04805}, 2018.

\bibitem{dimitriadis2023pareto}
Nikolaos Dimitriadis, Pascal Frossard, and Fran{\c{c}}ois Fleuret.
\newblock Pareto manifold learning: Tackling multiple tasks via ensembles of
  single-task models.
\newblock In {\em International Conference on Machine Learning}, pages
  8015--8052. PMLR, 2023.

\bibitem{dinh2017sharp}
Laurent Dinh, Razvan Pascanu, Samy Bengio, and Yoshua Bengio.
\newblock Sharp minima can generalize for deep nets.
\newblock In {\em International Conference on Machine Learning}, pages
  1019--1028. PMLR, 2017.

\bibitem{donyehiya2022cold}
Shachar Don-Yehiya, Elad Venezian, Colin Raffel, Noam Slonim, Yoav Katz, and
  Leshem Choshen.
\newblock Cold fusion: Collaborative descent for distributed multitask
  finetuning, 2022.

\bibitem{dong2020survey}
Xibin Dong, Zhiwen Yu, Wenming Cao, Yifan Shi, and Qianli Ma.
\newblock A survey on ensemble learning.
\newblock {\em Frontiers of Computer Science}, 14:241--258, 2020.

\bibitem{draxler2018essentially}
Felix Draxler, Kambis Veschgini, Manfred Salmhofer, and Fred Hamprecht.
\newblock Essentially no barriers in neural network energy landscape.
\newblock In {\em International conference on machine learning}, pages
  1309--1318. PMLR, 2018.

\bibitem{du2020agree}
Shangchen Du, Shan You, Xiaojie Li, Jianlong Wu, Fei Wang, Chen Qian, and
  Changshui Zhang.
\newblock Agree to disagree: Adaptive ensemble knowledge distillation in
  gradient space.
\newblock {\em advances in neural information processing systems},
  33:12345--12355, 2020.

\bibitem{duong2015low}
Long Duong, Trevor Cohn, Steven Bird, and Paul Cook.
\newblock Low resource dependency parsing: Cross-lingual parameter sharing in a
  neural network parser.
\newblock In {\em Proceedings of the 53rd annual meeting of the Association for
  Computational Linguistics and the 7th international joint conference on
  natural language processing (volume 2: short papers)}, pages 845--850, 2015.

\bibitem{Dvornik_2019_ICCV}
Nikita Dvornik, Cordelia Schmid, and Julien Mairal.
\newblock Diversity with cooperation: Ensemble methods for few-shot
  classification.
\newblock {\em Proceedings of the IEEE/CVF International Conference on Computer
  Vision (ICCV)}, 2019.

\bibitem{entezari2021role}
Rahim Entezari, Hanie Sedghi, Olga Saukh, and Behnam Neyshabur.
\newblock The role of permutation invariance in linear mode connectivity of
  neural networks.
\newblock {\em arXiv preprint arXiv:2110.06296}, 2021.

\bibitem{fallah2020personalized}
Alireza Fallah, Aryan Mokhtari, and Asuman Ozdaglar.
\newblock Personalized federated learning with theoretical guarantees: A
  model-agnostic meta-learning approach.
\newblock {\em Advances in Neural Information Processing Systems},
  33:3557--3568, 2020.

\bibitem{farouki2012bernstein}
Rida~T Farouki.
\newblock The bernstein polynomial basis: A centennial retrospective.
\newblock {\em Computer Aided Geometric Design}, 29(6):379--419, 2012.

\bibitem{fifty2021efficiently}
Chris Fifty, Ehsan Amid, Zhe Zhao, Tianhe Yu, Rohan Anil, and Chelsea Finn.
\newblock Efficiently identifying task groupings for multi-task learning.
\newblock {\em Advances in Neural Information Processing Systems},
  34:27503--27516, 2021.

\bibitem{fort2020deep}
Stanislav Fort, Gintare~Karolina Dziugaite, Mansheej Paul, Sepideh Kharaghani,
  Daniel~M Roy, and Surya Ganguli.
\newblock Deep learning versus kernel learning: an empirical study of loss
  landscape geometry and the time evolution of the neural tangent kernel.
\newblock {\em Advances in Neural Information Processing Systems},
  33:5850--5861, 2020.

\bibitem{fort2019deep}
Stanislav Fort, Huiyi Hu, and Balaji Lakshminarayanan.
\newblock Deep ensembles: A loss landscape perspective.
\newblock {\em arXiv preprint arXiv:1912.02757}, 2019.

\bibitem{fort2019large}
Stanislav Fort and Stanislaw Jastrzebski.
\newblock Large scale structure of neural network loss landscapes.
\newblock {\em Advances in Neural Information Processing Systems}, 32, 2019.

\bibitem{fort2019goldilocks}
Stanislav Fort and Adam Scherlis.
\newblock The goldilocks zone: Towards better understanding of neural network
  loss landscapes.
\newblock In {\em Proceedings of the aaai conference on artificial
  intelligence}, volume~33, pages 3574--3581, 2019.

\bibitem{frankle2020revisiting}
Jonathan Frankle.
\newblock Revisiting" qualitatively characterizing neural network optimization
  problems".
\newblock {\em arXiv preprint arXiv:2012.06898}, 2020.

\bibitem{frankle2018lottery}
Jonathan Frankle and Michael Carbin.
\newblock The lottery ticket hypothesis: Finding sparse, trainable neural
  networks.
\newblock {\em arXiv preprint arXiv:1803.03635}, 2018.

\bibitem{frankle2020linear}
Jonathan Frankle, Gintare~Karolina Dziugaite, Daniel Roy, and Michael Carbin.
\newblock Linear mode connectivity and the lottery ticket hypothesis.
\newblock In {\em International Conference on Machine Learning}, pages
  3259--3269. PMLR, 2020.

\bibitem{2016Topology}
C.~D. Freeman and J.~Bruna.
\newblock Topology and geometry of half-rectified network optimization.
\newblock 2016.

\bibitem{freund1997decision}
Yoav Freund and Robert~E Schapire.
\newblock A decision-theoretic generalization of on-line learning and an
  application to boosting.
\newblock {\em Journal of computer and system sciences}, 55(1):119--139, 1997.

\bibitem{gal2022stylegan}
Rinon Gal, Or~Patashnik, Haggai Maron, Amit~H Bermano, Gal Chechik, and Daniel
  Cohen-Or.
\newblock Stylegan-nada: Clip-guided domain adaptation of image generators.
\newblock {\em ACM Transactions on Graphics (TOG)}, 41(4):1--13, 2022.

\bibitem{gao2018digital}
Wen Gao, Yonghong Tian, and Jian Wang.
\newblock Digital retina: revolutionizing camera systems for the smart city.
\newblock {\em Science China Information Science}, 48(8):1076--1082, 2018.

\bibitem{gao2022revisiting}
Yingbo Gao, Christian Herold, Zijian Yang, and Hermann Ney.
\newblock Revisiting checkpoint averaging for neural machine translation.
\newblock {\em arXiv preprint arXiv:2210.11803}, 2022.

\bibitem{garipov2018loss}
Timur Garipov, Pavel Izmailov, Dmitrii Podoprikhin, Dmitry~P Vetrov, and
  Andrew~G Wilson.
\newblock Loss surfaces, mode connectivity, and fast ensembling of dnns.
\newblock {\em Advances in neural information processing systems}, 31, 2018.

\bibitem{gaya2022learning}
Jean-Baptiste Gaya, Laure Soulier, and Ludovic Denoyer.
\newblock Learning a subspace of policies for online adaptation in
  reinforcement learning, 2022.

\bibitem{godfrey2022symmetries}
Charles Godfrey, Davis Brown, Tegan Emerson, and Henry Kvinge.
\newblock On the symmetries of deep learning models and their internal
  representations.
\newblock {\em arXiv preprint arXiv:2205.14258}, 2022.

\bibitem{gomes2012computer}
Jonas Gomes, Luiz Velho, and Mario~Costa Sousa.
\newblock {\em Computer graphics: theory and practice}.
\newblock CRC Press, 2012.

\bibitem{gong2021ensemble}
Xuan Gong, Abhishek Sharma, Srikrishna Karanam, Ziyan Wu, Terrence Chen, David
  Doermann, and Arun Innanje.
\newblock Ensemble attention distillation for privacy-preserving federated
  learning.
\newblock In {\em Proceedings of the IEEE/CVF International Conference on
  Computer Vision}, pages 15076--15086, 2021.

\bibitem{goodfellow2014qualitatively}
Ian~J Goodfellow, Oriol Vinyals, and Andrew~M Saxe.
\newblock Qualitatively characterizing neural network optimization problems.
\newblock {\em arXiv preprint arXiv:1412.6544}, 2014.

\bibitem{gotmare2018using}
Akhilesh Gotmare, Nitish~Shirish Keskar, Caiming Xiong, and Richard Socher.
\newblock Using mode connectivity for loss landscape analysis.
\newblock {\em arXiv preprint arXiv:1806.06977}, 2018.

\bibitem{gressmann2020improving}
Frithjof Gressmann, Zach Eaton-Rosen, and Carlo Luschi.
\newblock Improving neural network training in low dimensional random bases.
\newblock {\em Advances in Neural Information Processing Systems},
  33:12140--12150, 2020.

\bibitem{grigorescu2020survey}
Sorin Grigorescu, Bogdan Trasnea, Tiberiu Cocias, and Gigel Macesanu.
\newblock A survey of deep learning techniques for autonomous driving.
\newblock {\em Journal of Field Robotics}, 37(3):362--386, 2020.

\bibitem{gu2023hierarchical}
Xiaozhe Gu, Zixun Zhang, Yuncheng Jiang, Tao Luo, Ruimao Zhang, Shuguang Cui,
  and Zhen Li.
\newblock Hierarchical weight averaging for deep neural networks.
\newblock {\em IEEE Transactions on Neural Networks and Learning Systems},
  2023.

\bibitem{guha2019one}
Neel Guha, Ameet Talwalkar, and Virginia Smith.
\newblock One-shot federated learning.
\newblock {\em arXiv preprint arXiv:1902.11175}, 2019.

\bibitem{guo2023stochastic}
Hao Guo, Jiyong Jin, and Bin Liu.
\newblock Stochastic weight averaging revisited.
\newblock {\em Applied Sciences}, 13(5):2935, 2023.

\bibitem{gupta2020stochastic}
Vipul Gupta, Santiago~Akle Serrano, and Dennis DeCoste.
\newblock Stochastic weight averaging in parallel: Large-batch training that
  generalizes well.
\newblock {\em arXiv preprint arXiv:2001.02312}, 2020.

\bibitem{han2021transformer}
Kai Han, An~Xiao, Enhua Wu, Jianyuan Guo, Chunjing Xu, and Yunhe Wang.
\newblock Transformer in transformer.
\newblock {\em Advances in Neural Information Processing Systems},
  34:15908--15919, 2021.

\bibitem{hansen1990neural}
Lars~Kai Hansen and Peter Salamon.
\newblock Neural network ensembles.
\newblock {\em IEEE transactions on pattern analysis and machine intelligence},
  12(10):993--1001, 1990.

\bibitem{hecht1990algebraic}
Robert Hecht-Nielsen.
\newblock On the algebraic structure of feedforward network weight spaces.
\newblock In {\em Advanced Neural Computers}, pages 129--135. Elsevier, 1990.

\bibitem{heinbaugh2022data}
Clare~Elizabeth Heinbaugh, Emilio Luz-Ricca, and Huajie Shao.
\newblock Data-free one-shot federated learning under very high statistical
  heterogeneity.
\newblock In {\em The Eleventh International Conference on Learning
  Representations}, 2022.

\bibitem{hinton2015distilling}
Geoffrey Hinton, Oriol Vinyals, and Jeff Dean.
\newblock Distilling the knowledge in a neural network, 2015.

\bibitem{hoang2019collective}
Minh Hoang, Nghia Hoang, Bryan Kian~Hsiang Low, and Carleton Kingsford.
\newblock Collective model fusion for multiple black-box experts.
\newblock In {\em International Conference on Machine Learning}, pages
  2742--2750. PMLR, 2019.

\bibitem{hochreiter1997flat}
Sepp Hochreiter and J{\"u}rgen Schmidhuber.
\newblock Flat minima.
\newblock {\em Neural computation}, 9(1):1--42, 1997.

\bibitem{hoffer2017train}
Elad Hoffer, Itay Hubara, and Daniel Soudry.
\newblock Train longer, generalize better: closing the generalization gap in
  large batch training of neural networks.
\newblock {\em Advances in neural information processing systems}, 30, 2017.

\bibitem{hsieh2020non}
Kevin Hsieh, Amar Phanishayee, Onur Mutlu, and Phillip Gibbons.
\newblock The non-iid data quagmire of decentralized machine learning.
\newblock In {\em International Conference on Machine Learning}, pages
  4387--4398. PMLR, 2020.

\bibitem{hsu2019measuring}
Tzu-Ming~Harry Hsu, Hang Qi, and Matthew Brown.
\newblock Measuring the effects of non-identical data distribution for
  federated visual classification, 2019.

\bibitem{hsu2020federated}
Tzu-Ming~Harry Hsu, Hang Qi, and Matthew Brown.
\newblock Federated visual classification with real-world data distribution.
\newblock In {\em Computer Vision--ECCV 2020: 16th European Conference,
  Glasgow, UK, August 23--28, 2020, Proceedings, Part X 16}, pages 76--92.
  Springer, 2020.

\bibitem{huang2023lorahub}
Chengsong Huang, Qian Liu, Bill~Yuchen Lin, Tianyu Pang, Chao Du, and Min Lin.
\newblock Lorahub: Efficient cross-task generalization via dynamic lora
  composition, 2023.

\bibitem{huang2017snapshot}
Gao Huang, Yixuan Li, Geoff Pleiss, Zhuang Liu, John~E Hopcroft, and Kilian~Q
  Weinberger.
\newblock Snapshot ensembles: Train 1, get m for free.
\newblock {\em arXiv preprint arXiv:1704.00109}, 2017.

\bibitem{huang2022achieving}
Tiansheng Huang, Shiwei Liu, Li~Shen, Fengxiang He, Weiwei Lin, and Dacheng
  Tao.
\newblock Achieving personalized federated learning with sparse local models.
\newblock {\em arXiv preprint arXiv:2201.11380}, 2022.

\bibitem{huang2023experts}
Yongqi Huang, Peng Ye, Xiaoshui Huang, Sheng Li, Tao Chen, and Wanli Ouyang.
\newblock Experts weights averaging: A new general training scheme for vision
  transformers.
\newblock {\em arXiv preprint arXiv:2308.06093}, 2023.

\bibitem{ilharco2022editing}
Gabriel Ilharco, Marco~Tulio Ribeiro, Mitchell Wortsman, Suchin Gururangan,
  Ludwig Schmidt, Hannaneh Hajishirzi, and Ali Farhadi.
\newblock Editing models with task arithmetic.
\newblock {\em arXiv preprint arXiv:2212.04089}, 2022.

\bibitem{ilharco2022patching}
Gabriel Ilharco, Mitchell Wortsman, Samir~Yitzhak Gadre, Shuran Song, Hannaneh
  Hajishirzi, Simon Kornblith, Ali Farhadi, and Ludwig Schmidt.
\newblock Patching open-vocabulary models by interpolating weights.
\newblock {\em arXiv preprint arXiv:2208.05592}, 2022.

\bibitem{2021Geometry}
B.~Imek, F.~Ged, A.~Jacot, F.~Spadaro, C.~Hongler, W.~Gerstner, and J.~Brea.
\newblock Geometry of the loss landscape in overparameterized neural networks:
  Symmetries and invariances.
\newblock 2021.

\bibitem{izacard2020leveraging}
Gautier Izacard and Edouard Grave.
\newblock Leveraging passage retrieval with generative models for open domain
  question answering.
\newblock {\em arXiv preprint arXiv:2007.01282}, 2020.

\bibitem{izenman2012introduction}
Alan~Julian Izenman.
\newblock Introduction to manifold learning.
\newblock {\em Wiley Interdisciplinary Reviews: Computational Statistics},
  4(5):439--446, 2012.

\bibitem{izmailov2018averaging}
Pavel Izmailov, Dmitrii Podoprikhin, Timur Garipov, Dmitry Vetrov, and
  Andrew~Gordon Wilson.
\newblock Averaging weights leads to wider optima and better generalization.
\newblock {\em arXiv preprint arXiv:1803.05407}, 2018.

\bibitem{jacot2018neural}
Arthur Jacot, Franck Gabriel, and Cl{\'e}ment Hongler.
\newblock Neural tangent kernel: Convergence and generalization in neural
  networks.
\newblock {\em Advances in neural information processing systems}, 31, 2018.

\bibitem{jain2018parallelizing}
Prateek Jain, Sham Kakade, Rahul Kidambi, Praneeth Netrapalli, and Aaron
  Sidford.
\newblock Parallelizing stochastic gradient descent for least squares
  regression: mini-batching, averaging, and model misspecification.
\newblock {\em Journal of Machine Learning Research}, 18, 2018.

\bibitem{jang2023exploring}
Joel Jang, Seungone Kim, Seonghyeon Ye, Doyoung Kim, Lajanugen Logeswaran,
  Moontae Lee, Kyungjae Lee, and Minjoon Seo.
\newblock Exploring the benefits of training expert language models over
  instruction tuning, 2023.

\bibitem{jangra2023survey}
Anubhav Jangra, Sourajit Mukherjee, Adam Jatowt, Sriparna Saha, and Mohammad
  Hasanuzzaman.
\newblock A survey on multi-modal summarization.
\newblock {\em ACM Computing Surveys}, 55(13s):1--36, 2023.

\bibitem{jeong2018communication}
Eunjeong Jeong, Seungeun Oh, Hyesung Kim, Jihong Park, Mehdi Bennis, and
  Seong-Lyun Kim.
\newblock Communication-efficient on-device machine learning: Federated
  distillation and augmentation under non-iid private data.
\newblock {\em arXiv preprint arXiv:1811.11479}, 2018.

\bibitem{jha2018bag}
Rahul Jha, Alex Marin, Suvamsh Shivaprasad, and Imed Zitouni.
\newblock Bag of experts architectures for model reuse in conversational
  language understanding.
\newblock In {\em Proceedings of the 2018 Conference of the North American
  Chapter of the Association for Computational Linguistics: Human Language
  Technologies, Volume 3 (Industry Papers)}, pages 153--161, 2018.

\bibitem{jhunjhunwala2023fedexp}
Divyansh Jhunjhunwala, Shiqiang Wang, and Gauri Joshi.
\newblock Fedexp: Speeding up federated averaging via extrapolation.
\newblock {\em arXiv preprint arXiv:2301.09604}, 2023.

\bibitem{jiang2023llm}
Dongfu Jiang, Xiang Ren, and Bill~Yuchen Lin.
\newblock Llm-blender: Ensembling large language models with pairwise ranking
  and generative fusion.
\newblock {\em arXiv preprint arXiv:2306.02561}, 2023.

\bibitem{jiang2020unifying}
Zetian Jiang, Tianzhe Wang, and Junchi Yan.
\newblock Unifying offline and online multi-graph matching via finding shortest
  paths on supergraph.
\newblock {\em IEEE transactions on pattern analysis and machine intelligence},
  43(10):3648--3663, 2020.

\bibitem{jin2022dataless}
Xisen Jin, Xiang Ren, Daniel Preotiuc-Pietro, and Pengxiang Cheng.
\newblock Dataless knowledge fusion by merging weights of language models.
\newblock {\em arXiv preprint arXiv:2212.09849}, 2022.

\bibitem{jolicoeur2023population}
Alexia Jolicoeur-Martineau, Emy Gervais, Kilian Fatras, Yan Zhang, and Simon
  Lacoste-Julien.
\newblock Population parameter averaging (papa).
\newblock {\em arXiv preprint arXiv:2304.03094}, 2023.

\bibitem{jordan2022repair}
Keller Jordan, Hanie Sedghi, Olga Saukh, Rahim Entezari, and Behnam Neyshabur.
\newblock Repair: Renormalizing permuted activations for interpolation repair.
\newblock {\em arXiv preprint arXiv:2211.08403}, 2022.

\bibitem{juneja2022linear}
Jeevesh Juneja, Rachit Bansal, Kyunghyun Cho, Jo{\~a}o Sedoc, and Naomi Saphra.
\newblock Linear connectivity reveals generalization strategies.
\newblock {\em arXiv preprint arXiv:2205.12411}, 2022.

\bibitem{kaddour2022stop}
Jean Kaddour.
\newblock Stop wasting my time! saving days of imagenet and bert training with
  latest weight averaging.
\newblock {\em arXiv preprint arXiv:2209.14981}, 2022.

\bibitem{kairouz2021advances}
Peter Kairouz, H~Brendan McMahan, Brendan Avent, Aur{\'e}lien Bellet, Mehdi
  Bennis, Arjun~Nitin Bhagoji, Kallista Bonawitz, Zachary Charles, Graham
  Cormode, Rachel Cummings, et~al.
\newblock Advances and open problems in federated learning.
\newblock {\em Foundations and Trends{\textregistered} in Machine Learning},
  14(1--2):1--210, 2021.

\bibitem{karita2021comparative}
Shigeki Karita, Yotaro Kubo, Michiel Adriaan~Unico Bacchiani, and Llion Jones.
\newblock A comparative study on neural architectures and training methods for
  japanese speech recognition.
\newblock {\em arXiv preprint arXiv:2106.05111}, 2021.

\bibitem{kawaguchi2016deep}
Kenji Kawaguchi.
\newblock Deep learning without poor local minima.
\newblock {\em Advances in neural information processing systems}, 29, 2016.

\bibitem{kendall2018multi}
Alex Kendall, Yarin Gal, and Roberto Cipolla.
\newblock Multi-task learning using uncertainty to weigh losses for scene
  geometry and semantics.
\newblock In {\em Proceedings of the IEEE conference on computer vision and
  pattern recognition}, pages 7482--7491, 2018.

\bibitem{keskar2016large}
Nitish~Shirish Keskar, Dheevatsa Mudigere, Jorge Nocedal, Mikhail Smelyanskiy,
  and Ping Tak~Peter Tang.
\newblock On large-batch training for deep learning: Generalization gap and
  sharp minima.
\newblock {\em arXiv preprint arXiv:1609.04836}, 2016.

\bibitem{kim2018paraphrasing}
Jangho Kim, SeongUk Park, and Nojun Kwak.
\newblock Paraphrasing complex network: Network compression via factor
  transfer.
\newblock {\em Advances in neural information processing systems}, 31, 2018.

\bibitem{kingetsu2021neural}
Hiroaki Kingetsu, Kenichi Kobayashi, and Taiji Suzuki.
\newblock Neural network module decomposition and recomposition.
\newblock {\em arXiv preprint arXiv:2112.13208}, 2021.

\bibitem{kirkpatrick2017overcoming}
James Kirkpatrick, Razvan Pascanu, Neil Rabinowitz, Joel Veness, Guillaume
  Desjardins, Andrei~A Rusu, Kieran Milan, John Quan, Tiago Ramalho, Agnieszka
  Grabska-Barwinska, et~al.
\newblock Overcoming catastrophic forgetting in neural networks.
\newblock {\em Proceedings of the national academy of sciences},
  114(13):3521--3526, 2017.

\bibitem{kolsbjerg2016automated}
Esben~L Kolsbjerg, Michael~N Groves, and Bj{\o}rk Hammer.
\newblock An automated nudged elastic band method.
\newblock {\em The Journal of chemical physics}, 145(9), 2016.

\bibitem{kontschieder2015deep}
Peter Kontschieder, Madalina Fiterau, Antonio Criminisi, and Samuel~Rota Bulo.
\newblock Deep neural decision forests.
\newblock In {\em Proceedings of the IEEE international conference on computer
  vision}, pages 1467--1475, 2015.

\bibitem{koratana2019lit}
Animesh Koratana, Daniel Kang, Peter Bailis, and Matei Zaharia.
\newblock Lit: Learned intermediate representation training for model
  compression.
\newblock In {\em International Conference on Machine Learning}, pages
  3509--3518. PMLR, 2019.

\bibitem{kuditipudi2019explaining}
Rohith Kuditipudi, Xiang Wang, Holden Lee, Yi~Zhang, Zhiyuan Li, Wei Hu, Rong
  Ge, and Sanjeev Arora.
\newblock Explaining landscape connectivity of low-cost solutions for
  multilayer nets.
\newblock {\em Advances in neural information processing systems}, 32, 2019.

\bibitem{laine2016temporal}
Samuli Laine and Timo Aila.
\newblock Temporal ensembling for semi-supervised learning.
\newblock {\em arXiv preprint arXiv:1610.02242}, 2016.

\bibitem{lam2021model}
Thanh~Chi Lam, Nghia Hoang, Bryan Kian~Hsiang Low, and Patrick Jaillet.
\newblock Model fusion for personalized learning.
\newblock In {\em International Conference on Machine Learning}, pages
  5948--5958. PMLR, 2021.

\bibitem{leang2020dynamic}
Isabelle Leang, Ganesh Sistu, Fabian B{\"u}rger, Andrei Bursuc, and Senthil
  Yogamani.
\newblock Dynamic task weighting methods for multi-task networks in autonomous
  driving systems.
\newblock In {\em 2020 IEEE 23rd International Conference on Intelligent
  Transportation Systems (ITSC)}, pages 1--8. IEEE, 2020.

\bibitem{lecun2015deep}
Yann LeCun, Yoshua Bengio, and Geoffrey Hinton.
\newblock Deep learning.
\newblock {\em nature}, 521(7553):436--444, 2015.

\bibitem{leonardos2017distributed}
Spyridon Leonardos, Xiaowei Zhou, and Kostas Daniilidis.
\newblock Distributed consistent data association via permutation
  synchronization.
\newblock In {\em 2017 IEEE International Conference on Robotics and Automation
  (ICRA)}, pages 2645--2652. IEEE, 2017.

\bibitem{leontev2020non}
Mikhail~Iu Leontev, Viktoriia Islenteva, and Sergey~V Sukhov.
\newblock Non-iterative knowledge fusion in deep convolutional neural networks.
\newblock {\em Neural Processing Letters}, 51:1--22, 2020.

\bibitem{li2018measuring}
Chunyuan Li, Heerad Farkhoor, Rosanne Liu, and Jason Yosinski.
\newblock Measuring the intrinsic dimension of objective landscapes.
\newblock {\em arXiv preprint arXiv:1804.08838}, 2018.

\bibitem{li2019fedmd}
Daliang Li and Junpu Wang.
\newblock Fedmd: Heterogenous federated learning via model distillation.
\newblock {\em arXiv preprint arXiv:1910.03581}, 2019.

\bibitem{li2018visualizing}
Hao Li, Zheng Xu, Gavin Taylor, Christoph Studer, and Tom Goldstein.
\newblock Visualizing the loss landscape of neural nets.
\newblock {\em Advances in neural information processing systems}, 31, 2018.

\bibitem{li2022branch}
Margaret Li, Suchin Gururangan, Tim Dettmers, Mike Lewis, Tim Althoff, Noah~A
  Smith, and Luke Zettlemoyer.
\newblock Branch-train-merge: Embarrassingly parallel training of expert
  language models.
\newblock {\em arXiv preprint arXiv:2208.03306}, 2022.

\bibitem{li2022trainable}
Tao Li, Zhehao Huang, Qinghua Tao, Yingwen Wu, and Xiaolin Huang.
\newblock Trainable weight averaging: Efficient training by optimizing
  historical solutions.
\newblock In {\em The Eleventh International Conference on Learning
  Representations}, 2022.

\bibitem{li2023trainable}
Tao Li, Zhehao Huang, Qinghua Tao, Yingwen Wu, and Xiaolin Huang.
\newblock Trainable weight averaging: A general approach for subspace training,
  2023.

\bibitem{li2022low}
Tao Li, Lei Tan, Zhehao Huang, Qinghua Tao, Yipeng Liu, and Xiaolin Huang.
\newblock Low dimensional trajectory hypothesis is true: Dnns can be trained in
  tiny subspaces.
\newblock {\em IEEE Transactions on Pattern Analysis and Machine Intelligence},
  45(3):3411--3420, 2022.

\bibitem{li2020federated}
Tian Li, Anit~Kumar Sahu, Ameet Talwalkar, and Virginia Smith.
\newblock Federated learning: Challenges, methods, and future directions.
\newblock {\em IEEE signal processing magazine}, 37(3):50--60, 2020.

\bibitem{li2015convergent}
Yixuan Li, Jason Yosinski, Jeff Clune, Hod Lipson, and John Hopcroft.
\newblock Convergent learning: Do different neural networks learn the same
  representations?
\newblock {\em arXiv preprint arXiv:1511.07543}, 2015.

\bibitem{Lintao2020FedDF}
Tao Lin, Lingjing Kong, Sebastian~U Stich, and Martin Jaggi.
\newblock Ensemble distillation for robust model fusion in federated learning.
\newblock In H.~Larochelle, M.~Ranzato, R.~Hadsell, M.F. Balcan, and H.~Lin,
  editors, {\em Advances in Neural Information Processing Systems}, volume~33,
  pages 2351--2363. Curran Associates, Inc., 2020.

\bibitem{liu2022deep}
Chang Liu, Chenfei Lou, Runzhong Wang, Alan~Yuhan Xi, Li~Shen, and Junchi Yan.
\newblock Deep neural network fusion via graph matching with applications to
  model ensemble and federated learning.
\newblock In {\em International Conference on Machine Learning}, pages
  13857--13869. PMLR, 2022.

\bibitem{liu2019knowledge}
Iou-Jen Liu, Jian Peng, and Alexander~G. Schwing.
\newblock Knowledge flow: Improve upon your teachers, 2019.

\bibitem{liu2021feddg}
Quande Liu, Cheng Chen, Jing Qin, Qi~Dou, and Pheng-Ann Heng.
\newblock Feddg: Federated domain generalization on medical image segmentation
  via episodic learning in continuous frequency space.
\newblock In {\em Proceedings of the IEEE/CVF Conference on Computer Vision and
  Pattern Recognition}, pages 1013--1023, 2021.

\bibitem{liu2021sparse}
Shiwei Liu, Tianlong Chen, Xiaohan Chen, Zahra Atashgahi, Lu~Yin, Huanyu Kou,
  Li~Shen, Mykola Pechenizkiy, Zhangyang Wang, and Decebal~Constantin Mocanu.
\newblock Sparse training via boosting pruning plasticity with
  neuroregeneration.
\newblock {\em Advances in Neural Information Processing Systems},
  34:9908--9922, 2021.

\bibitem{liu2021fedct}
Shuchang Liu, Shuyuan Xu, Wenhui Yu, Zuohui Fu, Yongfeng Zhang, and Amelie
  Marian.
\newblock Fedct: Federated collaborative transfer for recommendation.
\newblock In {\em Proceedings of the 44th international ACM SIGIR conference on
  research and development in information retrieval}, pages 716--725, 2021.

\bibitem{liu2023hierarchical}
Yajing Liu, Yuning Lu, Hao Liu, Yaozu An, Zhuoran Xu, Zhuokun Yao, Baofeng
  Zhang, Zhiwei Xiong, and Chenguang Gui.
\newblock Hierarchical prompt learning for multi-task learning.
\newblock In {\em Proceedings of the IEEE/CVF Conference on Computer Vision and
  Pattern Recognition}, pages 10888--10898, 2023.

\bibitem{liu2019roberta}
Yinhan Liu, Myle Ott, Naman Goyal, Jingfei Du, Mandar Joshi, Danqi Chen, Omer
  Levy, Mike Lewis, Luke Zettlemoyer, and Veselin Stoyanov.
\newblock Roberta: A robustly optimized bert pretraining approach.
\newblock {\em arXiv preprint arXiv:1907.11692}, 2019.

\bibitem{liu2018comparable}
Yuchen Liu, Long Zhou, Yining Wang, Yang Zhao, Jiajun Zhang, and Chengqing
  Zong.
\newblock A comparable study on model averaging, ensembling and reranking in
  nmt.
\newblock In {\em Natural Language Processing and Chinese Computing: 7th CCF
  International Conference, NLPCC 2018, Hohhot, China, August 26--30, 2018,
  Proceedings, Part II 7}, pages 299--308. Springer, 2018.

\bibitem{loiola2007survey}
Eliane~Maria Loiola, Nair Maria~Maia De~Abreu, Paulo~Oswaldo Boaventura-Netto,
  Peter Hahn, and Tania Querido.
\newblock A survey for the quadratic assignment problem.
\newblock {\em European journal of operational research}, 176(2):657--690,
  2007.

\bibitem{lou2019towards}
Yihang Lou, Ling-Yu Duan, Yong Luo, Ziqian Chen, Tongliang Liu, Shiqi Wang, and
  Wen Gao.
\newblock Towards digital retina in smart cities: A model generation,
  utilization and communication paradigm.
\newblock In {\em 2019 IEEE International Conference on Multimedia and Expo
  (ICME)}, pages 19--24. IEEE, 2019.

\bibitem{lubana2023mechanistic}
Ekdeep~Singh Lubana, Eric~J Bigelow, Robert~P Dick, David Krueger, and Hidenori
  Tanaka.
\newblock Mechanistic mode connectivity.
\newblock In {\em International Conference on Machine Learning}, pages
  22965--23004. PMLR, 2023.

\bibitem{luo2022nonlinear}
Yong Luo, Ling-Yu Duan, Yan Bai, Tongliang Liu, Yihang Lou, and Yonggang Wen.
\newblock Nonlinear multi-model reuse.
\newblock In {\em 2022 IEEE 24th International Workshop on Multimedia Signal
  Processing (MMSP)}, pages 1--6. IEEE, 2022.

\bibitem{lv2023parameter}
Xingtai Lv, Ning Ding, Yujia Qin, Zhiyuan Liu, and Maosong Sun.
\newblock Parameter-efficient weight ensembling facilitates task-level
  knowledge transfer.
\newblock In {\em Proceedings of the 61st Annual Meeting of the Association for
  Computational Linguistics (Volume 2: Short Papers)}, pages 270--282, 2023.

\bibitem{maddox2019simple}
Wesley~J Maddox, Pavel Izmailov, Timur Garipov, Dmitry~P Vetrov, and
  Andrew~Gordon Wilson.
\newblock A simple baseline for bayesian uncertainty in deep learning.
\newblock {\em Advances in neural information processing systems}, 32, 2019.

\bibitem{madry2017towards}
Aleksander Madry, Aleksandar Makelov, Ludwig Schmidt, Dimitris Tsipras, and
  Adrian Vladu.
\newblock Towards deep learning models resistant to adversarial attacks.
\newblock {\em arXiv preprint arXiv:1706.06083}, 2017.

\bibitem{malinin2019ensemble}
Andrey Malinin, Bruno Mlodozeniec, and Mark Gales.
\newblock Ensemble distribution distillation.
\newblock {\em arXiv preprint arXiv:1905.00076}, 2019.

\bibitem{maninis2019attentive}
Kevis-Kokitsi Maninis, Ilija Radosavovic, and Iasonas Kokkinos.
\newblock Attentive single-tasking of multiple tasks.
\newblock In {\em Proceedings of the IEEE/CVF conference on computer vision and
  pattern recognition}, pages 1851--1860, 2019.

\bibitem{matena2022merging}
Michael~S Matena and Colin~A Raffel.
\newblock Merging models with fisher-weighted averaging.
\newblock {\em Advances in Neural Information Processing Systems},
  35:17703--17716, 2022.

\bibitem{mcmahan2017communication}
Brendan McMahan, Eider Moore, Daniel Ramage, Seth Hampson, and Blaise~Aguera
  y~Arcas.
\newblock Communication-efficient learning of deep networks from decentralized
  data.
\newblock In {\em Artificial intelligence and statistics}, pages 1273--1282.
  PMLR, 2017.

\bibitem{mohri2019agnostic}
Mehryar Mohri, Gary Sivek, and Ananda~Theertha Suresh.
\newblock Agnostic federated learning.
\newblock In {\em International Conference on Machine Learning}, pages
  4615--4625. PMLR, 2019.

\bibitem{nagarajan2019uniform}
Vaishnavh Nagarajan and J~Zico Kolter.
\newblock Uniform convergence may be unable to explain generalization in deep
  learning.
\newblock {\em Advances in Neural Information Processing Systems}, 32, 2019.

\bibitem{nam2022improving}
Giung Nam, Hyungi Lee, Byeongho Heo, and Juho Lee.
\newblock Improving ensemble distillation with weight averaging and
  diversifying perturbation.
\newblock {\em arXiv preprint arXiv:2206.15047}, 2022.

\bibitem{neklyudov2018variance}
Kirill Neklyudov, Dmitry Molchanov, Arsenii Ashukha, and Dmitry Vetrov.
\newblock Variance networks: When expectation does not meet your expectations.
\newblock {\em arXiv preprint arXiv:1803.03764}, 2018.

\bibitem{nemirovski2009robust}
Arkadi Nemirovski, Anatoli Juditsky, Guanghui Lan, and Alexander Shapiro.
\newblock Robust stochastic approximation approach to stochastic programming.
\newblock {\em SIAM Journal on optimization}, 19(4):1574--1609, 2009.

\bibitem{neu2018iterate}
Gergely Neu and Lorenzo Rosasco.
\newblock Iterate averaging as regularization for stochastic gradient descent.
\newblock In {\em Conference On Learning Theory}, pages 3222--3242. PMLR, 2018.

\bibitem{neyshabur2020being}
Behnam Neyshabur, Hanie Sedghi, and Chiyuan Zhang.
\newblock What is being transferred in transfer learning?
\newblock {\em Advances in neural information processing systems}, 33:512--523,
  2020.

\bibitem{nguyen2019connected}
Quynh Nguyen.
\newblock On connected sublevel sets in deep learning.
\newblock In {\em International conference on machine learning}, pages
  4790--4799. PMLR, 2019.

\bibitem{nguyen2018loss}
Quynh Nguyen, Mahesh~Chandra Mukkamala, and Matthias Hein.
\newblock On the loss landscape of a class of deep neural networks with no bad
  local valleys.
\newblock {\em arXiv preprint arXiv:1809.10749}, 2018.

\bibitem{nishio2019client}
Takayuki Nishio and Ryo Yonetani.
\newblock Client selection for federated learning with heterogeneous resources
  in mobile edge.
\newblock In {\em ICC 2019-2019 IEEE international conference on communications
  (ICC)}, pages 1--7. IEEE, 2019.

\bibitem{oh2021fedbabu}
Jaehoon Oh, Sangmook Kim, and Se-Young Yun.
\newblock Fedbabu: Towards enhanced representation for federated image
  classification.
\newblock {\em arXiv preprint arXiv:2106.06042}, 2021.

\bibitem{openai2023chatgpt4}
OpenAI.
\newblock Gpt-4 technical report.
\newblock {\em arXiv preprint arXiv:2303.08774}, 2023.

\bibitem{Oquab_2014_CVPR}
Maxime Oquab, Leon Bottou, Ivan Laptev, and Josef Sivic.
\newblock Learning and transferring mid-level image representations using
  convolutional neural networks.
\newblock In {\em Proceedings of the IEEE Conference on Computer Vision and
  Pattern Recognition (CVPR)}, June 2014.

\bibitem{ortiz2023task}
Guillermo Ortiz-Jimenez, Alessandro Favero, and Pascal Frossard.
\newblock Task arithmetic in the tangent space: Improved editing of pre-trained
  models.
\newblock {\em arXiv preprint arXiv:2305.12827}, 2023.

\bibitem{o2020deep}
Niall O’Mahony, Sean Campbell, Anderson Carvalho, Suman Harapanahalli,
  Gustavo~Velasco Hernandez, Lenka Krpalkova, Daniel Riordan, and Joseph Walsh.
\newblock Deep learning vs. traditional computer vision.
\newblock In {\em Advances in Computer Vision: Proceedings of the 2019 Computer
  Vision Conference (CVC), Volume 1 1}, pages 128--144. Springer, 2020.

\bibitem{pan2009survey}
Sinno~Jialin Pan and Qiang Yang.
\newblock A survey on transfer learning.
\newblock {\em IEEE Transactions on knowledge and data engineering},
  22(10):1345--1359, 2009.

\bibitem{pathak2010multiparty}
Manas Pathak, Shantanu Rane, and Bhiksha Raj.
\newblock Multiparty differential privacy via aggregation of locally trained
  classifiers.
\newblock {\em Advances in neural information processing systems}, 23, 2010.

\bibitem{pena2022re}
Fidel A~Guerrero Pe{\~n}a, Heitor~Rapela Medeiros, Thomas Dubail, Masih
  Aminbeidokhti, Eric Granger, and Marco Pedersoli.
\newblock Re-basin via implicit sinkhorn differentiation.
\newblock {\em arXiv preprint arXiv:2212.12042}, 2022.

\bibitem{peyre2019computational}
Gabriel Peyr{\'e}, Marco Cuturi, et~al.
\newblock Computational optimal transport: With applications to data science.
\newblock {\em Foundations and Trends{\textregistered} in Machine Learning},
  11(5-6):355--607, 2019.

\bibitem{phang2020english}
Jason Phang, Iacer Calixto, Phu~Mon Htut, Yada Pruksachatkun, Haokun Liu, Clara
  Vania, Katharina Kann, and Samuel~R Bowman.
\newblock English intermediate-task training improves zero-shot cross-lingual
  transfer too.
\newblock {\em arXiv preprint arXiv:2005.13013}, 2020.

\bibitem{phang2018sentence}
Jason Phang, Thibault F{\'e}vry, and Samuel~R Bowman.
\newblock Sentence encoders on stilts: Supplementary training on intermediate
  labeled-data tasks.
\newblock {\em arXiv preprint arXiv:1811.01088}, 2018.

\bibitem{pittorino2022deep}
Fabrizio Pittorino, Antonio Ferraro, Gabriele Perugini, Christoph Feinauer,
  Carlo Baldassi, and Riccardo Zecchina.
\newblock Deep networks on toroids: removing symmetries reveals the structure
  of flat regions in the landscape geometry.
\newblock In {\em International Conference on Machine Learning}, pages
  17759--17781. PMLR, 2022.

\bibitem{polyak1990new}
Boris~T Polyak.
\newblock New stochastic approximation type procedures.
\newblock {\em Automat. i Telemekh}, 7(98-107):2, 1990.

\bibitem{polyak1992acceleration}
Boris~T Polyak and Anatoli~B Juditsky.
\newblock Acceleration of stochastic approximation by averaging.
\newblock {\em SIAM journal on control and optimization}, 30(4):838--855, 1992.

\bibitem{pruksachatkun2020intermediatetask}
Yada Pruksachatkun, Jason Phang, Haokun Liu, Phu~Mon Htut, Xiaoyi Zhang,
  Richard~Yuanzhe Pang, Clara Vania, Katharina Kann, and Samuel~R. Bowman.
\newblock Intermediate-task transfer learning with pretrained models for
  natural language understanding: When and why does it work?, 2020.

\bibitem{qin2022exploring}
Yujia Qin, Cheng Qian, Jing Yi, Weize Chen, Yankai Lin, Xu~Han, Zhiyuan Liu,
  Maosong Sun, and Jie Zhou.
\newblock Exploring mode connectivity for pre-trained language models.
\newblock {\em arXiv preprint arXiv:2210.14102}, 2022.

\bibitem{raffel2020exploring}
Colin Raffel, Noam Shazeer, Adam Roberts, Katherine Lee, Sharan Narang, Michael
  Matena, Yanqi Zhou, Wei Li, and Peter~J Liu.
\newblock Exploring the limits of transfer learning with a unified text-to-text
  transformer.
\newblock {\em The Journal of Machine Learning Research}, 21(1):5485--5551,
  2020.

\bibitem{rame2023model-ratatouille}
Alexandre Rame, Kartik Ahuja, Jianyu Zhang, Matthieu Cord, Leon Bottou, and
  David Lopez-Paz.
\newblock Model ratatouille: Recycling diverse models for out-of-distribution
  generalization.
\newblock 2023.

\bibitem{rame2023rewarded}
Alexandre Rame, Guillaume Couairon, Mustafa Shukor, Corentin Dancette,
  Jean-Baptiste Gaya, Laure Soulier, and Matthieu Cord.
\newblock Rewarded soups: towards pareto-optimal alignment by interpolating
  weights fine-tuned on diverse rewards, 2023.

\bibitem{rame2022diverse}
Alexandre Rame, Matthieu Kirchmeyer, Thibaud Rahier, Alain Rakotomamonjy,
  Patrick Gallinari, and Matthieu Cord.
\newblock Diverse weight averaging for out-of-distribution generalization.
\newblock {\em Advances in Neural Information Processing Systems}, 2022.

\bibitem{ravaut2022towards}
Mathieu Ravaut, Shafiq Joty, and Nancy~F Chen.
\newblock Towards summary candidates fusion.
\newblock {\em arXiv preprint arXiv:2210.08779}, 2022.

\bibitem{reddi2021adaptive}
Sashank Reddi, Zachary Charles, Manzil Zaheer, Zachary Garrett, Keith Rush,
  Jakub Konečný, Sanjiv Kumar, and H.~Brendan McMahan.
\newblock Adaptive federated optimization, 2021.

\bibitem{rokach2010ensemble}
Lior Rokach.
\newblock Ensemble-based classifiers.
\newblock {\em Artificial intelligence review}, 33:1--39, 2010.

\bibitem{ruppert1988efficient}
David Ruppert.
\newblock Efficient estimations from a slowly convergent robbins-monro process.
\newblock Technical report, Cornell University Operations Research and
  Industrial Engineering, 1988.

\bibitem{sagi2018ensemble}
Omer Sagi and Lior Rokach.
\newblock Ensemble learning: A survey.
\newblock {\em Wiley Interdisciplinary Reviews: Data Mining and Knowledge
  Discovery}, 8(4):e1249, 2018.

\bibitem{sagun2017empirical}
Levent Sagun, Utku Evci, V~Ugur Guney, Yann Dauphin, and Leon Bottou.
\newblock Empirical analysis of the hessian of over-parametrized neural
  networks.
\newblock {\em arXiv preprint arXiv:1706.04454}, 2017.

\bibitem{sattler2021fedaux}
Felix Sattler, Tim Korjakow, Roman Rischke, and Wojciech Samek.
\newblock Fedaux: Leveraging unlabeled auxiliary data in federated learning.
\newblock {\em IEEE Transactions on Neural Networks and Learning Systems},
  2021.

\bibitem{schapire1999brief}
Robert~E Schapire et~al.
\newblock A brief introduction to boosting.
\newblock In {\em Ijcai}, volume~99, pages 1401--1406. Citeseer, 1999.

\bibitem{schmidhuber1992learning}
J{\"u}rgen Schmidhuber.
\newblock Learning complex, extended sequences using the principle of history
  compression.
\newblock {\em Neural Computation}, 4(2):234--242, 1992.

\bibitem{shanahan2022talking}
Murray Shanahan.
\newblock Talking about large language models.
\newblock {\em arXiv preprint arXiv:2212.03551}, 2022.

\bibitem{shazeer2017outrageously}
Noam Shazeer, Azalia Mirhoseini, Krzysztof Maziarz, Andy Davis, Quoc Le,
  Geoffrey Hinton, and Jeff Dean.
\newblock Outrageously large neural networks: The sparsely-gated
  mixture-of-experts layer.
\newblock {\em arXiv preprint arXiv:1701.06538}, 2017.

\bibitem{shevchenko2020landscape}
Alexander Shevchenko and Marco Mondelli.
\newblock Landscape connectivity and dropout stability of sgd solutions for
  over-parameterized neural networks.
\newblock In {\em International Conference on Machine Learning}, pages
  8773--8784. PMLR, 2020.

\bibitem{shu2021zoo}
Yang Shu, Zhi Kou, Zhangjie Cao, Jianmin Wang, and Mingsheng Long.
\newblock Zoo-tuning: Adaptive transfer from a zoo of models.
\newblock In {\em International Conference on Machine Learning}, pages
  9626--9637. PMLR, 2021.

\bibitem{singh2020model}
Sidak~Pal Singh and Martin Jaggi.
\newblock Model fusion via optimal transport.
\newblock {\em Advances in Neural Information Processing Systems},
  33:22045--22055, 2020.

\bibitem{sinitsin2020editable}
Anton Sinitsin, Vsevolod Plokhotnyuk, Dmitriy Pyrkin, Sergei Popov, and Artem
  Babenko.
\newblock Editable neural networks.
\newblock {\em arXiv preprint arXiv:2004.00345}, 2020.

\bibitem{skorokhodov2019loss}
Ivan Skorokhodov and Mikhail Burtsev.
\newblock Loss landscape sightseeing with multi-point optimization.
\newblock {\em arXiv preprint arXiv:1910.03867}, 2019.

\bibitem{smith2017investigation}
Joshua Smith and Michael Gashler.
\newblock An investigation of how neural networks learn from the experiences of
  peers through periodic weight averaging.
\newblock In {\em 2017 16th IEEE International Conference on Machine Learning
  and Applications (ICMLA)}, pages 731--736. IEEE, 2017.

\bibitem{smith2017federated}
Virginia Smith, Chao-Kai Chiang, Maziar Sanjabi, and Ameet~S Talwalkar.
\newblock Federated multi-task learning.
\newblock {\em Advances in neural information processing systems}, 30, 2017.

\bibitem{srivastava2014dropout}
Nitish Srivastava, Geoffrey Hinton, Alex Krizhevsky, Ilya Sutskever, and Ruslan
  Salakhutdinov.
\newblock Dropout: a simple way to prevent neural networks from overfitting.
\newblock {\em The journal of machine learning research}, 15(1):1929--1958,
  2014.

\bibitem{standley2020tasks}
Trevor Standley, Amir Zamir, Dawn Chen, Leonidas Guibas, Jitendra Malik, and
  Silvio Savarese.
\newblock Which tasks should be learned together in multi-task learning?
\newblock In {\em International Conference on Machine Learning}, pages
  9120--9132. PMLR, 2020.

\bibitem{stoica2023zipit}
George Stoica, Daniel Bolya, Jakob Bjorner, Taylor Hearn, and Judy Hoffman.
\newblock Zipit! merging models from different tasks without training.
\newblock {\em arXiv preprint arXiv:2305.03053}, 2023.

\bibitem{sun2023multitask}
Tianxiang Sun, Zhengfu He, Qin Zhu, Xipeng Qiu, and Xuan-Jing Huang.
\newblock Multitask pre-training of modular prompt for chinese few-shot
  learning.
\newblock In {\em Proceedings of the 61st Annual Meeting of the Association for
  Computational Linguistics (Volume 1: Long Papers)}, pages 11156--11172, 2023.

\bibitem{sung2023empirical}
Yi-Lin Sung, Linjie Li, Kevin Lin, Zhe Gan, Mohit Bansal, and Lijuan Wang.
\newblock An empirical study of multimodal model merging.
\newblock {\em arXiv preprint arXiv:2304.14933}, 2023.

\bibitem{szegedy2016rethinking}
Christian Szegedy, Vincent Vanhoucke, Sergey Ioffe, Jon Shlens, and Zbigniew
  Wojna.
\newblock Rethinking the inception architecture for computer vision.
\newblock In {\em Proceedings of the IEEE conference on computer vision and
  pattern recognition}, pages 2818--2826, 2016.

\bibitem{tan2023geodesic}
Charlie Tan, Theodore Long, Sarah Zhao, and Rudolf Laine.
\newblock Geodesic mode connectivity.
\newblock 2023.

\bibitem{tang2023improving}
Anke Tang, Yong Luo, Han Hu, Fengxiang He, Kehua Su, Bo~Du, Yixin Chen, and
  Dacheng Tao.
\newblock Improving heterogeneous model reuse by density estimation.
\newblock {\em arXiv preprint arXiv:2305.13871}, 2023.

\bibitem{tarvainen2017mean}
Antti Tarvainen and Harri Valpola.
\newblock Mean teachers are better role models: Weight-averaged consistency
  targets improve semi-supervised deep learning results.
\newblock {\em Advances in neural information processing systems}, 30, 2017.

\bibitem{tatro2020optimizing}
Norman Tatro, Pin-Yu Chen, Payel Das, Igor Melnyk, Prasanna Sattigeri, and
  Rongjie Lai.
\newblock Optimizing mode connectivity via neuron alignment.
\newblock {\em Advances in Neural Information Processing Systems},
  33:15300--15311, 2020.

\bibitem{thibaux2007hierarchical}
Romain Thibaux and Michael~I Jordan.
\newblock Hierarchical beta processes and the indian buffet process.
\newblock In {\em Artificial intelligence and statistics}, pages 564--571.
  PMLR, 2007.

\bibitem{tran2020hydra}
Linh Tran, Bastiaan~S Veeling, Kevin Roth, Jakub Swiatkowski, Joshua~V Dillon,
  Jasper Snoek, Stephan Mandt, Tim Salimans, Sebastian Nowozin, and Rodolphe
  Jenatton.
\newblock Hydra: Preserving ensemble diversity for model distillation.
\newblock {\em arXiv preprint arXiv:2001.04694}, 2020.

\bibitem{tung2019similarity}
Frederick Tung and Greg Mori.
\newblock Similarity-preserving knowledge distillation.
\newblock In {\em Proceedings of the IEEE/CVF international conference on
  computer vision}, pages 1365--1374, 2019.

\bibitem{uriot2020safe}
Thomas Uriot and Dario Izzo.
\newblock Safe crossover of neural networks through neuron alignment.
\newblock In {\em Proceedings of the 2020 Genetic and Evolutionary Computation
  Conference}, pages 435--443, 2020.

\bibitem{utans1996weight}
Joachim Utans.
\newblock Weight averaging for neural networks and local resampling schemes.
\newblock In {\em Proc. AAAI-96 Workshop on Integrating Multiple Learned
  Models. AAAI Press}, pages 133--138. Citeseer, 1996.

\bibitem{vu2020exploring}
Tu~Vu, Tong Wang, Tsendsuren Munkhdalai, Alessandro Sordoni, Adam Trischler,
  Andrew Mattarella-Micke, Subhransu Maji, and Mohit Iyyer.
\newblock Exploring and predicting transferability across nlp tasks.
\newblock {\em arXiv preprint arXiv:2005.00770}, 2020.

\bibitem{wang2016chinese}
Benyou Wang, Jiabin Niu, Liqun Ma, Yuhua Zhang, Lipeng Zhang, Jingfei Li, Peng
  Zhang, and Dawei Song.
\newblock A chinese question answering approach integrating count-based and
  embedding-based features.
\newblock pages 934--941. Springer, 2016.

\bibitem{wang2021boost}
Feng Wang, Guoyizhe Wei, Qiao Liu, Jinxiang Ou, Hairong Lv, et~al.
\newblock Boost neural networks by checkpoints.
\newblock {\em Advances in Neural Information Processing Systems},
  34:19719--19729, 2021.

\bibitem{wang2020federated}
Hongyi Wang, Mikhail Yurochkin, Yuekai Sun, Dimitris Papailiopoulos, and
  Yasaman Khazaeni.
\newblock Federated learning with matched averaging.
\newblock {\em arXiv preprint arXiv:2002.06440}, 2020.

\bibitem{wang2018identifying}
Huan Wang, Nitish~Shirish Keskar, Caiming Xiong, and Richard Socher.
\newblock Identifying generalization properties in neural networks.
\newblock {\em arXiv preprint arXiv:1809.07402}, 2018.

\bibitem{wang2023exploring}
Ren Wang, Yuxuan Li, and Sijia Liu.
\newblock Exploring diversified adversarial robustness in neural networks via
  robust mode connectivity.
\newblock In {\em Proceedings of the IEEE/CVF Conference on Computer Vision and
  Pattern Recognition}, pages 2345--2351, 2023.

\bibitem{wang2020clustering}
Tianzhe Wang, Zetian Jiang, and Junchi Yan.
\newblock Clustering-aware multiple graph matching via decayed pairwise
  matching composition.
\newblock In {\em Proceedings of the The Thirty-Fourth AAAI Conference on
  Artificial Intelligence (AAAI-20), New York, NY, USA}, pages 7--12, 2020.

\bibitem{wang2022meta}
Zhenyi Wang, Xiaoyang Wang, Li~Shen, Qiuling Suo, Kaiqiang Song, Dong Yu, Yan
  Shen, and Mingchen Gao.
\newblock Meta-learning without data via wasserstein distributionally-robust
  model fusion.
\newblock In {\em Uncertainty in Artificial Intelligence}, pages 2045--2055.
  PMLR, 2022.

\bibitem{wei2023ntk}
Tianxin Wei, Zeming Guo, Yifan Chen, and Jingrui He.
\newblock Ntk-approximating mlp fusion for efficient language model
  fine-tuning.
\newblock 2023.

\bibitem{wen2017ensemble}
Guihua Wen, Zhi Hou, Huihui Li, Danyang Li, Lijun Jiang, and Eryang Xun.
\newblock Ensemble of deep neural networks with probability-based fusion for
  facial expression recognition.
\newblock {\em Cognitive Computation}, 9(5):597--610, 2017.

\bibitem{wen2023optimizing}
Haitao Wen, Haoyang Cheng, Heqian Qiu, Lanxiao Wang, Lili Pan, and Hongliang
  Li.
\newblock Optimizing mode connectivity for class incremental learning.
\newblock 2023.

\bibitem{wen2020batchensemble}
Yeming Wen, Dustin Tran, and Jimmy Ba.
\newblock Batchensemble: an alternative approach to efficient ensemble and
  lifelong learning.
\newblock {\em arXiv preprint arXiv:2002.06715}, 2020.

\bibitem{wolpert1992stacked}
David~H Wolpert.
\newblock Stacked generalization.
\newblock {\em Neural networks}, 5(2):241--259, 1992.

\bibitem{wortsman2022fi}
Mitchell Wortsman, Suchin Gururangan, Shen Li, Ali Farhadi, Ludwig Schmidt,
  Michael Rabbat, and Ari~S Morcos.
\newblock lo-fi: distributed fine-tuning without communication.
\newblock {\em arXiv preprint arXiv:2210.11948}, 2022.

\bibitem{wortsman2021learning}
Mitchell Wortsman, Maxwell~C Horton, Carlos Guestrin, Ali Farhadi, and Mohammad
  Rastegari.
\newblock Learning neural network subspaces.
\newblock In {\em International Conference on Machine Learning}, pages
  11217--11227. PMLR, 2021.

\bibitem{wortsman2022model}
Mitchell Wortsman, Gabriel Ilharco, Samir~Ya Gadre, Rebecca Roelofs, Raphael
  Gontijo-Lopes, Ari~S Morcos, Hongseok Namkoong, Ali Farhadi, Yair Carmon,
  Simon Kornblith, et~al.
\newblock Model soups: averaging weights of multiple fine-tuned models improves
  accuracy without increasing inference time.
\newblock In {\em International Conference on Machine Learning}, pages
  23965--23998. PMLR, 2022.

\bibitem{wortsman2022robust}
Mitchell Wortsman, Gabriel Ilharco, Jong~Wook Kim, Mike Li, Simon Kornblith,
  Rebecca Roelofs, Raphael~Gontijo Lopes, Hannaneh Hajishirzi, Ali Farhadi,
  Hongseok Namkoong, et~al.
\newblock Robust fine-tuning of zero-shot models.
\newblock In {\em Proceedings of the IEEE/CVF Conference on Computer Vision and
  Pattern Recognition}, pages 7959--7971, 2022.

\bibitem{Wu_2019_CVPR}
Ancong Wu, Wei-Shi Zheng, Xiaowei Guo, and Jian-Huang Lai.
\newblock Distilled person re-identification: Towards a more scalable system.
\newblock In {\em Proceedings of the IEEE/CVF Conference on Computer Vision and
  Pattern Recognition (CVPR)}, June 2019.

\bibitem{wu2021peer}
Guile Wu and Shaogang Gong.
\newblock Peer collaborative learning for online knowledge distillation.
\newblock In {\em Proceedings of the AAAI Conference on artificial
  intelligence}, volume~35, pages 10302--10310, 2021.

\bibitem{wu2019heterogeneous}
Xi-Zhu Wu, Song Liu, and Zhi-Hua Zhou.
\newblock Heterogeneous model reuse via optimizing multiparty multiclass
  margin.
\newblock In {\em International Conference on Machine Learning}, pages
  6840--6849. PMLR, 2019.

\bibitem{wu2021model}
Xi-Zhu Wu, Wenkai Xu, Song Liu, and Zhi-Hua Zhou.
\newblock Model reuse with reduced kernel mean embedding specification.
\newblock {\em IEEE Transactions on Knowledge and Data Engineering},
  35(1):699--710, 2021.

\bibitem{xiang2017modal}
Yang Yang De-Chuan~Zhan Xiang and Yu~Guo~Yuan Jiang.
\newblock Modal consistency based pre-trained multi-model reuse.
\newblock In {\em Proc. IJCAI}, 2017.

\bibitem{yan2015multi}
Junchi Yan, Minsu Cho, Hongyuan Zha, Xiaokang Yang, and Stephen~M Chu.
\newblock Multi-graph matching via affinity optimization with graduated
  consistency regularization.
\newblock {\em IEEE transactions on pattern analysis and machine intelligence},
  38(6):1228--1242, 2015.

\bibitem{yan2020learning}
Junchi Yan, Shuang Yang, and Edwin~R Hancock.
\newblock Learning for graph matching and related combinatorial optimization
  problems.
\newblock In {\em Proceedings of the Twenty-Ninth International Joint
  Conference on Artificial Intelligence, IJCAI-20}, pages 4988--4996.
  International Joint Conferences on Artificial Intelligence Organization,
  2020.

\bibitem{yang2019swalp}
Guandao Yang, Tianyi Zhang, Polina Kirichenko, Junwen Bai, Andrew~Gordon
  Wilson, and Chris De~Sa.
\newblock Swalp: Stochastic weight averaging in low precision training.
\newblock In {\em International Conference on Machine Learning}, pages
  7015--7024. PMLR, 2019.

\bibitem{yang2017deep}
Yang Yang, De-Chuan Zhan, Ying Fan, Yuan Jiang, and Zhi-Hua Zhou.
\newblock Deep learning for fixed model reuse.
\newblock In {\em Proceedings of the AAAI Conference on Artificial
  Intelligence}, volume~31, 2017.

\bibitem{yao2021deep}
Kaixuan Yao, Feilong Cao, Yee Leung, and Jiye Liang.
\newblock Deep neural network compression through interpretability-based filter
  pruning.
\newblock {\em Pattern Recognition}, 119:108056, 2021.

\bibitem{you2022ranking}
Kaichao You, Yong Liu, Ziyang Zhang, Jianmin Wang, Michael~I Jordan, and
  Mingsheng Long.
\newblock Ranking and tuning pre-trained models: a new paradigm for exploiting
  model hubs.
\newblock {\em The Journal of Machine Learning Research}, 23(1):9400--9446,
  2022.

\bibitem{you2017learning}
Shan You, Chang Xu, Chao Xu, and Dacheng Tao.
\newblock Learning from multiple teacher networks.
\newblock In {\em Proceedings of the 23rd ACM SIGKDD International Conference
  on Knowledge Discovery and Data Mining}, pages 1285--1294, 2017.

\bibitem{yun2023traversing}
EungGu Yun, Hyungi Lee, Giung Nam, and Juho Lee.
\newblock Traversing between modes in function space for fast ensembling.
\newblock {\em arXiv preprint arXiv:2306.11304}, 2023.

\bibitem{yurochkin2019bayesian}
Mikhail Yurochkin, Mayank Agarwal, Soumya Ghosh, Kristjan Greenewald, Nghia
  Hoang, and Yasaman Khazaeni.
\newblock Bayesian nonparametric federated learning of neural networks.
\newblock In {\em International conference on machine learning}, pages
  7252--7261. PMLR, 2019.

\bibitem{zhai2022scaling}
Xiaohua Zhai, Alexander Kolesnikov, Neil Houlsby, and Lucas Beyer.
\newblock Scaling vision transformers.
\newblock In {\em Proceedings of the IEEE/CVF Conference on Computer Vision and
  Pattern Recognition}, pages 12104--12113, 2022.

\bibitem{zhang2018graph}
Chris Zhang, Mengye Ren, and Raquel Urtasun.
\newblock Graph hypernetworks for neural architecture search.
\newblock {\em arXiv preprint arXiv:1810.05749}, 2018.

\bibitem{zhang2022dense}
Jie Zhang, Chen Chen, Bo~Li, Lingjuan Lyu, Shuang Wu, Shouhong Ding, Chunhua
  Shen, and Chao Wu.
\newblock Dense: Data-free one-shot federated learning.
\newblock {\em Advances in Neural Information Processing Systems},
  35:21414--21428, 2022.

\bibitem{zhang2022fine}
Lin Zhang, Li~Shen, Liang Ding, Dacheng Tao, and Ling-Yu Duan.
\newblock Fine-tuning global model via data-free knowledge distillation for
  non-iid federated learning.
\newblock In {\em Proceedings of the IEEE/CVF conference on computer vision and
  pattern recognition}, pages 10174--10183, 2022.

\bibitem{zhang2019lookahead}
Michael Zhang, James Lucas, Jimmy Ba, and Geoffrey~E Hinton.
\newblock Lookahead optimizer: k steps forward, 1 step back.
\newblock {\em Advances in neural information processing systems}, 32, 2019.

\bibitem{zhang2023zhijian}
Yi-Kai Zhang, Lu~Ren, Chao Yi, Qi-Wei Wang, De-Chuan Zhan, and Han-Jia Ye.
\newblock Zhijian: A unifying and rapidly deployable toolbox for pre-trained
  model reuse.
\newblock {\em arXiv preprint arXiv:2308.09158}, 2023.

\bibitem{zhang2018overview}
Yu~Zhang and Qiang Yang.
\newblock An overview of multi-task learning.
\newblock {\em National Science Review}, 5(1):30--43, 2018.

\bibitem{zhang2012communication}
Yuchen Zhang, Martin~J Wainwright, and John~C Duchi.
\newblock Communication-efficient algorithms for statistical optimization.
\newblock {\em Advances in neural information processing systems}, 25, 2012.

\bibitem{zhao2020bridging}
Pu~Zhao, Pin-Yu Chen, Payel Das, Karthikeyan~Natesan Ramamurthy, and Xue Lin.
\newblock Bridging mode connectivity in loss landscapes and adversarial
  robustness.
\newblock {\em arXiv preprint arXiv:2005.00060}, 2020.

\bibitem{zhao2023survey}
Wayne~Xin Zhao, Kun Zhou, Junyi Li, Tianyi Tang, and Xiaolei Wang.
\newblock A survey of large language models, 2023.

\bibitem{zhou2020distilled}
Yanlin Zhou, George Pu, Xiyao Ma, Xiaolin Li, and Dapeng Wu.
\newblock Distilled one-shot federated learning.
\newblock {\em arXiv preprint arXiv:2009.07999}, 2020.

\bibitem{zhou2016learnware}
Zhi-Hua Zhou.
\newblock Learnware: on the future of machine learning.
\newblock {\em Frontiers Comput. Sci.}, 10(4):589--590, 2016.

\bibitem{zhou2019deep}
Zhi-Hua Zhou and Ji~Feng.
\newblock Deep forest.
\newblock {\em National science review}, 6(1):74--86, 2019.

\bibitem{zhu2021data}
Zhuangdi Zhu, Junyuan Hong, and Jiayu Zhou.
\newblock Data-free knowledge distillation for heterogeneous federated
  learning.
\newblock In {\em International conference on machine learning}, pages
  12878--12889. PMLR, 2021.

\bibitem{zimmer2023sparse}
Max Zimmer, Christoph Spiegel, and Sebastian Pokutta.
\newblock Sparse model soups: A recipe for improved pruning via model
  averaging.
\newblock {\em arXiv preprint arXiv:2306.16788}, 2023.

\end{thebibliography}

\end{document}